\documentclass[sigconf, screen, nonacm=true]{acmart}

\pdfoutput=1

\usepackage[utf8]{inputenc} 
\usepackage[T1]{fontenc}    
\usepackage{enumitem}
\usepackage{url}            
\usepackage{xspace}
\usepackage{booktabs}       
\usepackage{amsfonts}       
\usepackage{nicefrac}       
\usepackage{microtype}      
\usepackage{xcolor}         
\usepackage[noend]{algpseudocode}
\usepackage{algorithm}
\usepackage{multirow}
\usepackage{multicol}
\usepackage{amsmath}
\usepackage{amsthm}
\usepackage{graphicx}
\usepackage{lipsum}
\usepackage{subcaption}
\usepackage{wrapfig}
\usepackage{mathtools}
\usepackage{array}
\usepackage[capitalise, noabbrev]{cleveref}
\usepackage{tabularx}

\usepackage{xcolor}
\usepackage[suppress]{color-edits}  
\addauthor{lw}{red}   

\hypersetup{
    colorlinks = true,
    citecolor = [RGB]{0,130,130},
    urlcolor = [RGB]{200,0,100}
}

\title[Off-Policy Evaluation for Large Action Spaces via Policy Convolution]{Off-Policy Evaluation for Large Action Spaces\\via Policy Convolution}

\author{Noveen Sachdeva}
\authornote{Work done during internship at Netflix.}
\email{nosachde@ucsd.edu}
\affiliation{%
  \institution{University of California, San Diego}
  \city{La Jolla}
  \state{CA}
  \country{USA}
}

\author{Lequn Wang}
\email{lequnw@netflix.com}
\affiliation{%
  \institution{Netflix}
  \city{Los Gatos}
  \state{CA}
  \country{USA}
}

\author{Dawen Liang}
\email{dliang@netflix.com}
\affiliation{%
  \institution{Netflix}
  \city{Los Gatos}
  \state{CA}
  \country{USA}
}

\author{Nathan Kallus}
\email{kallus@cornell.edu}
\affiliation{%
  \institution{Netflix \& Cornell University}
  \city{Los Gatos}
  \state{CA}
  \country{USA}
}

\author{Julian McAuley}
\email{jmcauley@ucsd.edu}
\affiliation{%
  \institution{University of California, San Diego}
  \city{La Jolla}
  \state{CA}
  \country{USA}
}

\begin{document}

\makeatletter \renewcommand\paragraph{\@startsection{paragraph}{4}{\z@} {2mm \@plus1ex \@minus.2ex} {-0.7em} {\normalfont\normalsize\bfseries}} \makeatother

\newcommand{\listheader}[1]{\item \textbf{#1} }

\newcommand\overstar[1]{\ThisStyle{\ensurestackMath{%
  \setbox0=\hbox{$\SavedStyle#1$}%
  \stackengine{0pt}{\copy0}{\kern.2\ht0\smash{\SavedStyle*}}{O}{c}{F}{T}{S}}}}

\newtheorem{assumption}[theorem]{Assumption}

\newcommand{\STAB}[1]{\begin{tabular}{@{}c@{}}#1\end{tabular}}

\newcommand{\bs}[1]{\ensuremath{\bm{\mathit{#1}}}}
\newcommand{\ie}[0]{\emph{i.e.}\xspace}
\newcommand{\etc}[0]{\emph{etc.}\xspace}
\newcommand{\eg}[0]{\emph{e.g.}\xspace}
\newcommand{\vs}[0]{\emph{vs.}\xspace}
\newcommand{\wrt}[0]{\emph{w.r.t.}\xspace}

\newcommand{\argmin}[1]{\underset{#1}{\operatorname{arg}\,\operatorname{min}}\;}
\newcommand{\expectation}[2]{\underset{#1}{\mathbb{E}} \left[#2\right]}
\newcommand{\inlineexpectation}[2]{{\mathbb{E}_{#1}} \left[#2\right]}
\newcommand{\independent}{\perp\!\!\!\!\perp}

\newcommand{\estimatorfull}{Policy Convolution\xspace}
\newcommand{\estimator}{\texttt{PC}\xspace}
\newcommand{\oracle}{\texttt{PC}\xspace}
\newcommand{\is}{\texttt{IS}\xspace}
\newcommand{\ips}{\texttt{IPS}\xspace}
\newcommand{\snips}{\texttt{SNIPS}\xspace}
\newcommand{\dr}{\texttt{DR}\xspace}
\newcommand{\dm}{\texttt{DM}\xspace}
\newcommand{\sndr}{\texttt{SNDR}\xspace}

\begin{abstract}
    Developing accurate off-policy estimators is crucial for both evaluating and optimizing for new policies. The main challenge in off-policy estimation is the distribution shift between the logging policy that generates data and the target policy that we aim to evaluate. Typically, techniques for correcting distribution shift involve some form of importance sampling. This approach results in unbiased value estimation but often comes with the trade-off of high variance, even in the simpler case of one-step contextual bandits. Furthermore, importance sampling relies on the common support assumption, which becomes impractical when the action space is large.
    To address these challenges, we introduce the \estimatorfull (\estimator) family of estimators. These methods leverage latent structure within actions---made available through action embeddings---to strategically \emph{convolve} the logging and target policies. This convolution introduces a unique bias-variance trade-off, which can be controlled by adjusting the \emph{amount of convolution}. 
    Our experiments on synthetic and 
    benchmark datasets demonstrate remarkable mean squared error (MSE) improvements when using \estimator, especially when either the action space or policy mismatch becomes large, with gains of up to $5-6$ orders of magnitude over existing estimators. \looseness=-1
    
\end{abstract}

\maketitle

\section{Introduction}
Off-policy estimation (OPE) is a fundamental problem in reinforcement learning and decision making under uncertainty. It involves estimating the expected value of a target policy, given access to only an offline dataset logged by deploying a different policy, often referred as the logging policy (see \citep{ope_survey} for a comprehensive survey). This decoupling between data collection and policy evaluation is crucial in many real-world applications, as it allows for the assessment of new policies using historical data without having to deploy them in the environment, which can be costly and/or risky. 
In this paper, we focus on OPE for the one-step contextual bandit setting, \ie, we perform decision making with only an observed context that is assumed to be independently sampled (\eg, a user coming to a website), and do not consider any recurrent dependencies in the context transitions as is the case in the general formulation of reinforcement learning.
A variety of practical applications naturally fall into the off-policy contextual bandit framework, \eg, recommender systems \cite{rec_as_treatments, ope_recsys}, healthcare \cite{ope_healthcare_1, ope_healthcare_2}, robotics \cite{ope_robotics}, \etc 

OPE, in its most general setting, can be a very challenging problem due to its inherently counterfactual nature, as we observe the reward for only those actions taken by the logging policy, while we aim to evaluate 
\emph{any} target policy.
For example, consider a scenario where the logging policy in a movie recommendation platform, 
for a given segment of users, rarely recommends romantic movies. This can often happen when we think a user will not like certain type of movies. On the other hand, a target policy---whose value we aim to estimate---due to numerous potential reasons, now chooses to recommend romantic movies for the same user segment.
This distribution-shift can lead to irrecoverable bias in our estimates \cite{ips_support}, making it difficult to accurately evaluate a target policy or learn a better one, which typically involves optimizing over the value estimates~\cite{crm, banditnet}. 


Typical off-policy estimators utilize Importance Sampling (\is) to correct for the policy mismatch between the target and logging policies \cite{ips, snips, dr, sndr1, cab, switch, dr_shrinkage, kernel_ips, mips, subgaussian_ope}, leading to unbiased value estimation, at the cost of high variance. The variance problem caused by \is is exacerbated if the target and logging policies exhibit significant divergence, and even more so if the action space is large. Notably, large action spaces frequently occur in practical OPE scenarios, \eg, recommender systems which can have millions of items (actions) \cite{pinsage, nips_22}, extreme classification \cite{eclare, extreme_opl, extreme_bandits}, discretized continuous action-spaces \cite{discrete_continuous_action_space}, \etc 

To address the aforementioned limitations of \is, we propose the \estimatorfull (\estimator) family of estimators. \estimator strategically convolves the logging and target policies by exploiting the inherent latent-structure amongst actions---available through action-embeddings---to make importance sampling operate in a more favorable bias-variance trade-off region. Such structure can occur naturally in different forms like action meta-data (text, images, \etc), action hierarchies, categories, \etc Or they can be estimated using domain-specific representation learning techniques \cite{simclr}. Notably, the utilization of additional action-structure has also been studied in the online multi-armed bandit literature (Lipschitz bandits) \cite{lipshchitz_bandits_1, zooming_lipshchitz_bandits, tree_zooming_lipshchitz_bandits}, albeit in the context of regret minimization.

To be more specific, the \estimator framework for OPE consists of two components: (1) conventional \is-based value estimation; and (2) convolving \emph{both} the target and logging policies using action-action similarity. 
\estimator allows full freedom over the choice of the \emph{backbone \is estimator}, the convolution function, as well as the \emph{amount of convolution} to conduct on the target and logging policies respectively.

To better understand the practical effectiveness of \estimator, we compare its performance with various off-policy estimators on synthetic and real-world benchmark datasets, simulating a variety of off-policy scenarios. Our results demonstrate that \estimator can effectively balance the bias-variance trade-off posited by policy convolutions, leading to up to $5-6$ orders of magnitude better off-policy evaluation in terms of mean squared error (MSE), particularly when the action space is large or the policy mismatch is high. To summarize, \emph{in this paper}, we make the following contributions:
\begin{itemize}[leftmargin=.3in,topsep=.1in]
    \item Introduce the \estimatorfull (\estimator) family of off-policy estimators that posit a novel bias-variance trade-off controlled by the \emph{amount of convolution} on the logging and target policies.
    \item Propose four different convolution functions for the \estimator framework, where each convolution function is accompanied with its unique set of inductive biases, thereby leading to distinct performance comparisons.
    \item Conduct empirical analyses on synthetic and real-world benchmark datasets that demonstrate the superiority of \estimator over a variety of off-policy estimators, especially when the action space or policy mismatch becomes large.
\end{itemize}

\section{Preliminaries}
\subsection{OPE in Contextual Bandits}
\label{sec:setup}
We study OPE in the standard stochastic contextual bandits setting with a context space $\mathcal{X}$ and a finite action space $\mathcal{A}$. In each round $i$, the agent observes a context $x_i \in \mathcal{X}$, takes an action $a_i \in \mathcal{A}$, and observes a reward $r_i \in [0, 1]$. The context $x_i$ is drawn from some unknown distribution $p(x)$. The action $a_i$ follows some policy $\pi(\cdot | x_i)$, and the reward $r_i$ is draw from an unknown distribution $p(r | x_i, a_i)$ with expected value $\delta(a,x) \triangleq \inlineexpectation{r \sim p(r | a, x)}{r}$. 
The value of a policy is its expected reward
\begin{align*}
V(\pi) \triangleq \expectation{x \sim p(x)}{\expectation{a \sim \pi(\cdot|x)}{\delta(a,x)}}. 
\end{align*}


In OPE, given a target policy $\pi$, we aim to estimate its value $V(\pi)$ using some bandit feedback data $\mathcal{D} \triangleq \{ (x_i, a_i, r_i) \}_{i=1}^n$ collected by deploying a different policy $\mu$. We call $\mu$ the logging policy and assume it is known.


\begin{figure*}[!t] 
    \centering
    \includegraphics[width=0.9\linewidth]{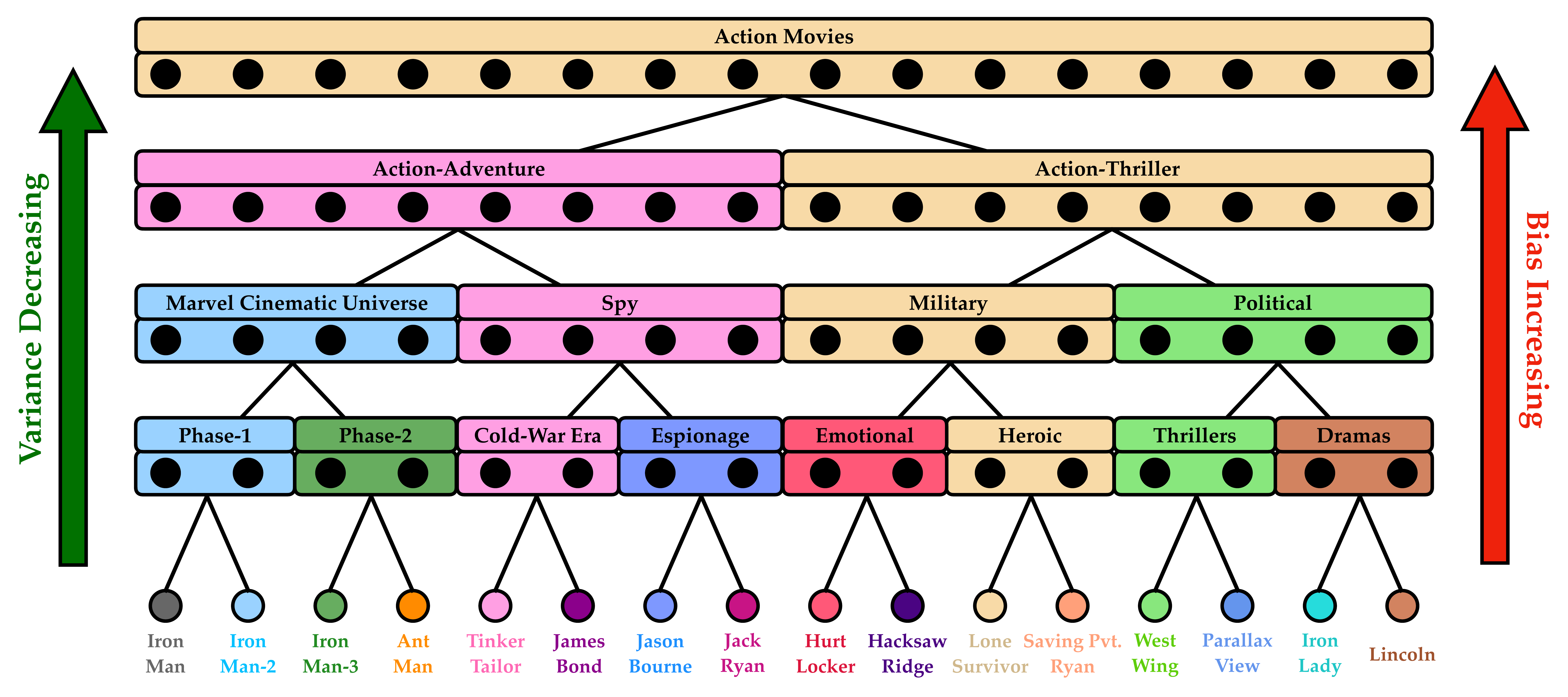}
    \caption{Intuition for the \estimator framework demonstrated using a hierarchical action tree, where similar actions (movies in this example) are recursively agglomerated together. The last level (leaf nodes) represents the complete action space, and the higher levels consist of ``meta-actions'' that represent a group of individual actions. As we go higher, \estimator defines the convolved policy for a given action, as the mean probability of all actions inside its corresponding \emph{meta-action}. Hence, we obtain the uniform policy at the topmost-level, and recover the original policy at the last-level.}
    \label{fig:intuition}
\end{figure*}

\subsection{Conventional OPE Estimators} \label{sec:preliminaries}

We now briefly discuss a few prominent OPE estimators, which will also be used to instantiate our proposed \estimatorfull (\estimator) estimator discussed in \cref{sec:pc}.

\subsubsection{Direct Method (\dm)} \label{sec:dm}
Taking a model-based approach, \dm leverages a reward-model to estimate the value of the target policy. Formally, given a suitable $\hat{\delta} : \mathcal{A} \times \mathcal{X} \mapsto \mathbb{R}$, the estimator is defined as follows:
\begin{align*}
    \hat{V}_{\dm}(\pi) \triangleq  \expectation{(x,\cdot,\cdot) \sim \mathcal{D}}{\sum_{a \in \mathcal{A}} \pi(a|x) \cdot \hat{\delta}(a, x)},
\end{align*}
where the outer expectation is over the finite set of logged bandit feedback data $\mathcal{D}$. Notably, the variance of $\hat{V}_{\dm}(\cdot)$ is often quite low, since $\hat{\delta}$ is typically bounded. However, it can suffer from a large bias problem due to model misspecification \cite{mrdr}.


\subsubsection{Inverse Propensity Scoring (\ips)} \label{sec:ips}
\ips \cite{ips} estimator uses Monte-Carlo approximation and importance sampling to account for the policy-mismatch between $\pi$ and $\mu$ as follows:
\begin{align*}
    \hat{V}_{\ips}(\pi) \triangleq  \expectation{(x,a,r) \sim \mathcal{D}}{\frac{\pi(a|x)}{\mu(a|x)} \cdot r}.
\end{align*}
The \ips estimator is unbiased under the following two assumptions which we assume throughout the paper unless otherwise specified:

\begin{assumption} \label{assumption:unconfoundness}
    {\normalfont \textbf{(Unconfoundedness)}} The action selection procedure is independent of all potential outcomes given the context, i.e., $\forall x \in \mathcal{X}, \pi \in \Pi, a \sim \pi(\cdot|x): ~~ \{ \delta(a', x) \}_{a' \in \mathcal{A}} \independent a ~|~ x$. 
\end{assumption}
\begin{assumption} \label{assumption:common_support}
    {\normalfont \textbf{(Common Support)}} The target policy $\pi$ shares common support with the logging policy $\mu$, $\forall x \in \mathcal{X}$, $a \in \mathcal{A}$: $~~ \pi(a|x) > 0 \implies \mu(a|x) > 0$. 
\end{assumption}

However, IPS estimator can suffer from a large variance problem, since the importance weights $\pi(a|x) / \mu(a|x)$ can be unbounded and huge. Several estimators are proposed to reduce the variance of the IPS estimator. 

\subsubsection{Self-normalized Inverse Propensity Scoring (\snips)} \label{sec:snips}
Built on the observation that the expected propensity weight in \ips equals $1$ , \snips \cite{snips} uses the empirical average of the propensity weights as a control variate for \ips as follows:
\begin{gather*}
    \hat{V}_{\snips}(\pi) \triangleq \expectation{(x,a,r) \sim \mathcal{D}}{\frac{\pi(a|x)}{\rho \cdot \mu(a|x)} \cdot r} 
    \text{s.t.} ~~ \rho \triangleq \expectation{(x,a,\cdot) \sim \mathcal{D}}{\frac{\pi(a|x)}{\mu(a|x)}}
    .
\end{gather*}
\snips typically enjoys smaller variance at the cost of a slight added bias in comparison to \ips, especially when the variance of the propensity weight is large\cite{snips_var_reduction}. Further, $\hat{V}_{\snips}(\pi)$ is a strongly consistent estimator of $V(\pi)$ by the law of large numbers.

\subsubsection{Doubly Robust (\dr)} \label{sec:dr}
DR combines the benefits of unbiased estimation in \ips and the low-variance, model-based estimation in \dm: 
\begin{gather*}
    \hat{V}_{\dr}(\pi) \triangleq  \expectation{(x,a,r) \sim \mathcal{D}}{\frac{\pi(a|x)}{\mu(a|x)} \cdot (r - \hat{\delta}(a, x)) + \Delta(\pi, x)} \\
    \text{s.t.} ~~~~ \Delta(\pi, x) \triangleq \sum_{a' \in \mathcal{A}} \pi(a'|x) \cdot \hat{\delta}(a', x)
    ,
\end{gather*}
where $\hat{\delta}$ is the same reward-model as used in \dm (\cref{sec:dm}). Intuitively, \dr uses the reward-model as a baseline, and performs importance sampling only on the error of the given reward-model. \dr is unbiased and can be of smaller variance than \ips when the reward-model $\hat{\delta}$ is close to the true reward $\delta$ \cite{dr}.

\subsubsection{Self-normalized Doubly Robust (\sndr)} \label{sec:sndr}
Similar to the idea behind \snips (\cref{sec:snips}); \sndr \cite{sndr1, sndr2} performs the same control variate technique on the \dr estimator (\cref{sec:dr}) as follows:
\begin{gather*}
    \hat{V}_{\sndr}(\pi) \triangleq \expectation{(x,a,r) \sim \mathcal{D}}{\frac{\pi(a|x)}{\rho \cdot \mu(a|x)} \cdot (r - \hat{\delta}(a, x)) + \Delta(\pi, x)} \\
    \text{s.t.} ~~~~ \rho \triangleq \expectation{(x,a,\cdot) \sim \mathcal{D}}{\frac{\pi(a|x)}{\mu(a|x)}} ~~~~;~~~~ \Delta(\pi, x) \triangleq \sum_{a \in \mathcal{A}} \pi(a|x) \cdot \hat{\delta}(a, x)
    .
\end{gather*}
Hence, \sndr encapsulates the ideas behind all the aforementioned estimators to conduct strongly consistent, low-variance policy value estimation that might perform well (in terms of MSE) in practice.


While effective to some extent, the importance sampling based estimators mentioned above can still suffer from large variance due to large importance sampling weights, especially when the action space is large. In particular, the variance of these importance sampling based estimators grows roughly linearly \wrt the maximum propensity weight in $\mathcal{D}$. And the maximum propensity weight can grow linearly \wrt the size of action space $\Omega(|\mathcal{A}|)$ \cite{ope_slate}, making these estimators undesirable for OPE for large action-space problems. 
Further, when Assumption \ref{assumption:common_support} is violated, the variance of such importance sampling based estimators becomes unbounded, in addition to incurring a bias of $\inlineexpectation{x}{\sum_{a \in \mathcal{U}(x)}\pi(a|x)\delta(a|x)}$, where $\mathcal{U}(x)$ is the set of actions where $\mu(\cdot|x)$ doesn't put any probability mass on (blind spots) \cite{ips_support}.

To address the aforementioned problems of
 importance sampling based estimators, we introduce policy convolution that makes use of the latent structure within actions in the next section. 
\section{OPE via \estimatorfull} 
\label{sec:pc}

In addition to the offline dataset $\mathcal{D}$, we further posit access to some embeddings $\mathcal{E} : \mathcal{A} \mapsto \mathbb{R}^d$ of the actions, which maps an action $a$ to a $d$-dimensional embedding space $\mathcal{E}(a) \in \mathbb{R}^d$. Let $\mathcal{E}_{\mathcal{A}} \subset \mathbb{R}^d$ be the subspace spanned by $\mathcal{E}$. Ideally, the embedding should capture action-similarity information, \ie, smaller distance in the embedding space should imply smaller difference in terms of expected reward for each context $x \in \mathcal{X}$. Notably, such action-embeddings are typically readily available in many industrial recommender systems, \eg, via matrix factorization \cite{mf}.

We are now ready to define our \estimatorfull framework for OPE. Taking \ips (\cref{sec:ips}) as a representative ``backbone'' estimator for \estimator, we define \estimator-\ips as follows: 
\begin{align*}
    \hat{V}_{\estimator-\ips}(\pi) &\triangleq \expectation{(x, a, r) \sim \mathcal{D}}{
        \frac{(\pi(\cdot|x) \ast f_{\tau_1})(a)}{(\mu(\cdot|x) \ast f_{\tau_2})(a)} 
        \cdot r
    } \\
    &=
    \expectation{(x, a, r) \sim \mathcal{D}}{
        \frac{\sum_{a'} \pi(a'|x) \cdot f_{\tau_1}(\mathcal{E}(a) , \mathcal{E}(a'))}{\sum_{a'} \mu(a'|x) \cdot f_{\tau_2}(\mathcal{E}(a) , \mathcal{E}(a'))} 
        \cdot r
    } 
    ,
\end{align*}
where `$\ast$' represents the convolution operator specified in the action-embedding domain, and $f_\tau : \mathbb{R}^d \times \mathbb{R}^d \mapsto \mathbb{R}$ is an action-action similarity (or convolution) function which has a parameter `$\tau$' to control the amount of convolution. Notably, \estimator is not limited to the \ips backbone estimator discussed hitherto, and we analogously define \estimator for other backbone estimators, namely, Self-Normalized IPS (\snips), Doubly Robust (\dr), Self-Normalized DR (\sndr) discussed in \cref{sec:snips,sec:dr,sec:sndr}, and call such estimators \estimator-\snips, \estimator-\dr, and \estimator-\sndr for convenience. We provide their exact specifications in \cref{appendix:pc_estimators}.

\begin{table*}[!t]
    \begin{center}
        \begin{tabular}{c c | c c c c | c | c c c}
            \toprule
            
            & & \multirow{2}{*}{$a_1$} & \multirow{2}{*}{$a_2$} & \multirow{2}{*}{$a_3$} & \multirow{2}{*}{$a_4$} & \multirow{2}{*}{$V^*(\cdot)$} & \multicolumn{3}{c}{\ips with $|\mathcal{D}| = 10$} \\[5pt] 

            & & & & & & & $\operatorname{MSE}$ & $\operatorname{Bias}^2$ & $\operatorname{Var}$ \\

            \midrule
            
            \multicolumn{2}{c|}{$\delta(\cdot, x_1)$} & 5 & 10 & 15 & 20 & $-$ & $-$ & $-$ & $-$ \\
            
            \midrule

            \multirow{2}{*}{$\tau=1$} & $\pi(\cdot|x_1)$ & 0.0 & 0.2 & 0.2 & 0.6 & 17 & \multirow{2}{*}{48.0} & \multirow{2}{*}{$\approx0$} & \multirow{2}{*}{48.0} \\
            & $\mu(\cdot|x_1)$ & 0.2 & 0.2 & 0.4 & 0.2 & 13 \\
            
            \midrule

            \multirow{2}{*}{$\tau=2$} & $(\pi \ast f_{\tau})(\cdot|x_1)$ & 0.1 & 0.1 & 0.4 & 0.4 & 15.5 & \multirow{2}{*}{13.4} & \multirow{2}{*}{4.6} & \multirow{2}{*}{8.8} \\
            & $(\mu \ast f_{\tau})(\cdot|x_1)$ & 0.2 & 0.2 & 0.3 & 0.3 & 13.5 \\
            
            \midrule

            \multirow{2}{*}{$\tau=3$} & $(\pi \ast f_{\tau})(\cdot|x_1)$ & 0.25 & 0.25 & 0.25 & 0.25 & 12.5 & \multirow{2}{*}{18.6} & \multirow{2}{*}{16} & \multirow{2}{*}{2.6} \\
            & $(\mu \ast f_{\tau})(\cdot|x_1)$ & 0.25 & 0.25 & 0.25 & 0.25 & 12.5 \\
            
            \bottomrule
        \end{tabular}
    \end{center}
    \vspace{3mm} 
    \caption{A toy example to intuit \estimator using the Tree similarity function, and constrained to have $\tau_1=\tau_2$. Similar to \cref{fig:intuition}, in this toy example, the action tree is a $3$-level complete binary tree, where the action partitioning is defined as $\{ (a_1, a_2, a_3, a_4) \} \rightarrow \{ (a_1, a_2), (a_3, a_4) \} \rightarrow \{ (a_1), (a_2), (a_3), (a_4) \}$ from top to bottom.}
    \label{tab:toy}
\end{table*}

We also illustrate the intuition behind policy convolutions in \cref{fig:intuition}, where \estimator \emph{strategically} biases the logging and target policies toward the uniform policy, by leveraging the underlying action structure, in this case, specified via a hierarchical grouping of actions. As we will observe in \cref{sec:results}, such convolutions in-turn lead to a new bias-variance trade-off, controlled by the amount of convolution ($\tau$). We propose four suitable instantiations of the action-action convolution (or pooling) function, $f_\tau(\cdot, \cdot)$ for \estimator:
\begin{itemize}[leftmargin=0.1in]
    \listheader{Kernel Smoothing.} Perhaps the most intuitive, we use the idea of multi-variate kernel smoothing \cite{kernel_smoothing_book} in the action-embedding space to derive our similarity function as:
    \begin{align*}
        f_{\tau}\left(\mathcal{E}(a) , \mathcal{E}(a')\right) ~~&\triangleq~~ \mathcal{K}\left(\mathcal{E}(a) , \mathcal{E}(a')\right) \\
        &=~~ \frac{1}{\tau^d} \prod_{i=1}^d \mathbf{K}\left(\frac{\mathcal{E}(a)_i , \mathcal{E}(a')_i}{\tau}\right)
        ,
    \end{align*}
    where, $\mathbf{K}$ is a suitable kernel function (\eg, Gaussian), and $\tau \in \mathbb{R}$ now corresponds to the bandwidth. It is worth noting that such a formulation can also be derived by viewing actions as continuous treatments, as defined by their embeddings \cite{kernel_ips, continuous_action_propensity}. However, since an inverse mapping from $\mathbb{R}^d \mapsto \mathcal{A}$ does not exist in our discrete action problem, having treatments outside $\mathcal{E}_{\mathcal{A}}$ is meaningless.
    
    \listheader{Tree Smoothing.} In this setting, we use $\mathcal{E}$ to recursively partition the action-space (see \cref{fig:intuition} for a depiction) into a tree-like structure of depth $D$, where each depth can be specified by $\mathcal{T}_d \triangleq \{ \mathbf{a}_{d, 1}, \mathbf{a}_{d, 2}, \ldots, \mathbf{a}_{d, k} \}$, where $\mathbf{a}_{d, i}$ is a \emph{meta-action} (set) comprising of singular actions, such that $\mathcal{A} \equiv \bigcup_{i=1}^k \mathbf{a}_{d, i}$, and $\mathbf{a}_{d, i} \cap \mathbf{a}_{d, j} = \Phi$ for all $i \neq j$ pairs. Notably, $\mathcal{T}_D$ (root node) consists of a single meta-action with all actions, and $\mathcal{T}_1 \equiv \mathcal{A}$ (last level) consists of each singular action. The similarity function is then defined as:
    \begin{align*}
        f_{\tau}\left(\mathcal{E}(a) , \mathcal{E}(a')\right) \triangleq
        \frac{\mathbb{I}(a' \in \mathbf{a}_{\tau}(a))}{|\mathbf{a}_{\tau}(a)|}
        ,
    \end{align*}
    where, $\mathbb{I}(\cdot)$ represents the indicator function, $\tau$ signifies the depth of the action-tree to use, and $\mathbf{a}_{\tau}(a)$ represents the meta-action at depth $\tau$ corresponding to the action $a$.

    \listheader{Ball Smoothing.} In this setting, we define a binary similarity function based on a fixed-radius ball around the given action, as defined by $\mathcal{E}$ as follows:
    \begin{align*}
        f_{\tau}\left(\mathcal{E}(a) , \mathcal{E}(a')\right) \triangleq
        \frac{\mathbb{I}\left(\lVert \mathcal{E}(a) - \mathcal{E}(a') \rVert_2^2 < \tau\right)}{\left|\left\{ ~~ \lVert \mathcal{E}(a) - \mathcal{E}(a'') \rVert_2^2 < \tau ~~|~~ \forall a'' \in \mathcal{A} ~~ \right\}\right|}
        ,
    \end{align*}
    where, $\tau$ signifies the radius of the ball around $\mathcal{E}(a)$. 

    \listheader{kNN Smoothing.} In this setting, we define a binary similarity function as the k-nearest neighbors decision function:
    \begin{align*}
        f_{\tau}\left(\mathcal{E}(a) , \mathcal{E}(a')\right) \triangleq
        \frac{\mathbb{I}\left(a' \in \operatorname{kNN}(a, \tau)\right)}{\tau}
        ,
    \end{align*}
    where, $\tau$ signifies the number of nearest neighbors to use, and $\operatorname{kNN}(a, \tau)$ represents the set of $\tau$ nearest neighbors of $\mathcal{E}(a)$ in $\mathcal{E}_{\mathcal{A}}$.
\end{itemize}


We note that our general \estimatorfull framework encompasses the existing OPE estimators designed for large action spaces \cite{saito2023off, peng2023offline}. 
As we will later show in Section~\ref{sec:results}, using the convolution functions proposed in \cref{sec:pc}, \estimator is able to achieve significantly better performance than existing estimators.
More specifically, groupIPS~\cite{peng2023offline} can be generalized as \estimator-\ips and offCEM~\cite{saito2023off} as \estimator-\dr, both using a two-depth tree (\ie, flat clustering) in the tree convolution function, both with an additional constraint of $\tau_1 = \tau_2 = 1$. As we will further note in our experiments (\cref{sec:results}): (1) using the kernel and kNN convolution functions tend to perform better than others; and (2) convolving the logging and target policies \emph{differently} (\ie, $\tau_1 \neq \tau_2$) adds a lot of flexibility to \estimator, leading to much better estimation than either convolving the two policies equally, or convolving only one out of the two policies.

\paragraph{Motivating example.} To gain a better intuition of \estimator, we refer to \cref{fig:intuition} and construct a four-action, single-context toy example described in \cref{tab:toy}. We conduct OPE using the \ips estimator at various levels of the action-tree, with a sample size of $|\mathcal{D}| = 10$ and repeat the experiment $50$k times. The results demonstrate that as we progress to higher levels of the tree (increased pooling), variance decreases, but bias increases. At the leaf level, \ips is unbiased but exhibits high variance. At the top-most level, while variance is the lowest, bias is significantly increased. When $\tau = 2$, we observe the best bias-variance trade-off, leading to the lowest MSE. 


\section{Experiments} \label{sec:experiments}

\begin{figure*}[!t]    
    \includegraphics[width=0.9\linewidth,trim={0 8.5cm 0 0},clip]{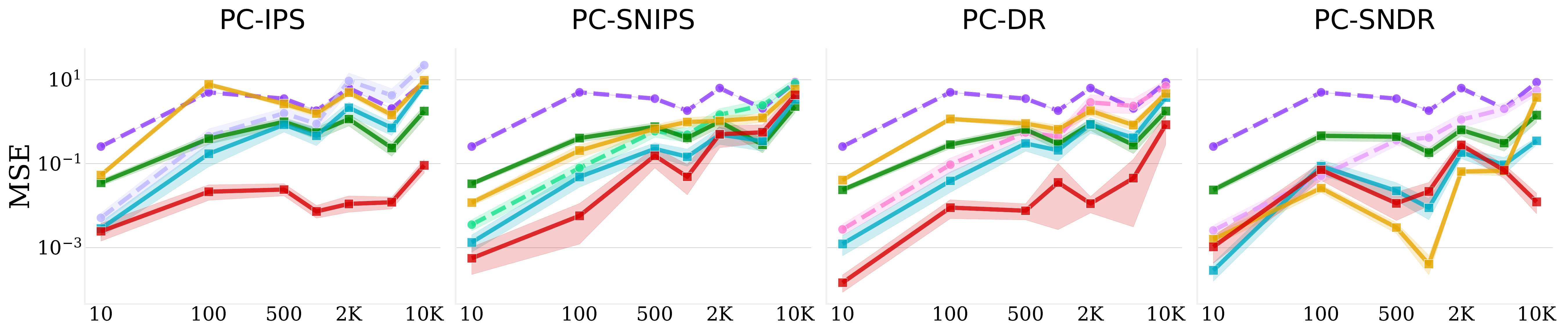} 

    \vspace{0.15cm}
    
    \includegraphics[width=0.9\linewidth,trim={0 0 0 3.2cm},clip]{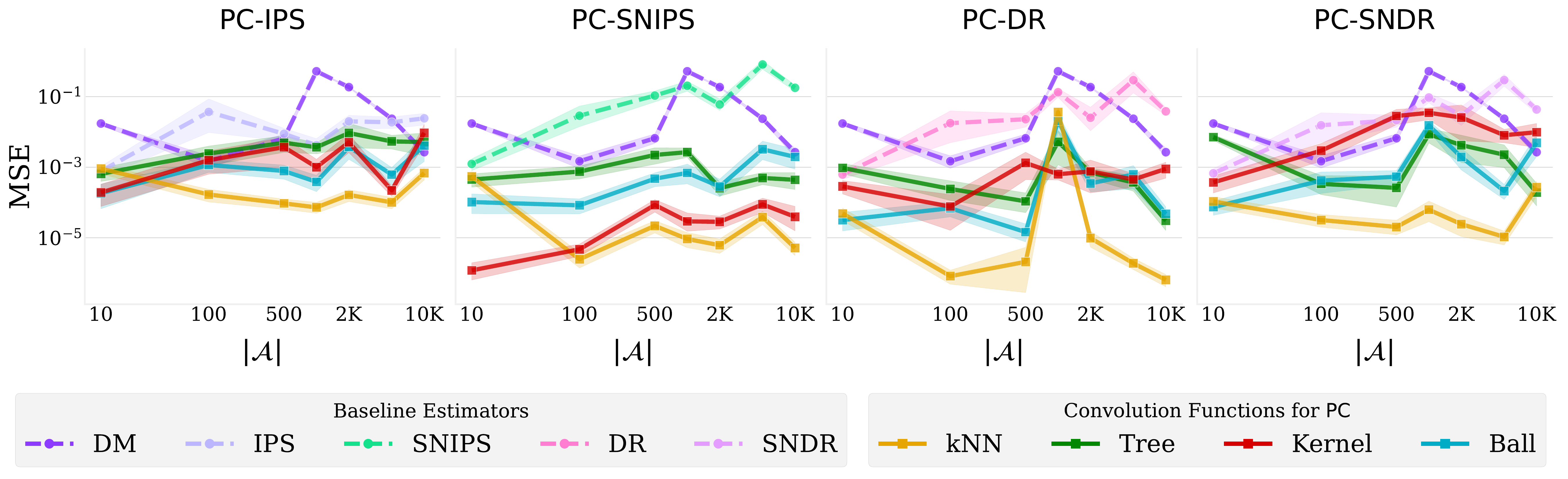} 
    \caption{Change in MSE while estimating $V(\pi_{\mathsf{good}})$ with \textbf{varying number of actions} ($\log$-$\log$ scale) for the synthetic dataset, using data logged by (top) $\mu_{\mathsf{uniform}}$, or (bottom) $\mu_{\mathsf{good}}$. Results for $V(\pi_{\mathsf{bad}})$ can be found in \cref{appendix:experiments}, \cref{fig:v_actions_pi_bad_mu_unif,fig:v_actions_pi_bad_mu_good}.}
    \label{fig:n_actions}
\end{figure*}

\subsection{Setup}
We measure \estimator's empirical effectiveness on two datasets. Firstly, we simulate a synthetic contextual bandit setup, beginning by sampling contexts $x \sim \mathcal{N}(0, I)$. Subsequently, to realize our assumption that there indeed exists a latent structure amongst individual actions, we randomly assign each action to one of $32$ latent \emph{topics}, denoted by $\Tilde{a}$. We also assign each latent topic a corresponding mean $\{ \mu_i \sim \mathcal{N}(0, I) \}_{i=1}^{32}$ and covariance $\{ \sigma_i \sim \mathcal{N}(0, I) \}_{i=1}^{32}$. To realize the assigned structure in the action-space, we sample each action's embedding from its correspondingly assigned topic, \ie, $\mathcal{E}(a) \sim \mathcal{N}(\mu_{\Tilde{a}}, \sigma_{\Tilde{a}})$. We then model the reward function $\delta(a, x)$ as a noisy and non-linear function of the underlying context- and action-embedding: $\delta(a, x) \triangleq \Phi(x \lVert \mathcal{E}(a) \lVert \epsilon)$, where ``$\lVert$'' represents concatenation, $\epsilon \sim \mathcal{N}(0, I)$ is white-noise, and $\Phi$ is a randomly initialized, two-layer neural network. Such a formulation realizes two crucial assumptions: (a) semantically closer actions are nearby according to $\mathcal{E}$, and (b) $\mathcal{E}$ shares a causal connection with the downstream reward function. Finally, we define the logging policy $\mu(\cdot|x)$ as a temperature-activated softmax distribution on the ground-truth reward distribution $\delta(\cdot, x)$, and the target policy as the $\epsilon-$greedy policy: 
\begin{gather} \raisetag{\baselineskip}\begin{aligned}
\label{eqn:synthetic_policies_defn}
    \mu(a|x) &\triangleq \frac{\operatorname{exp}(\beta \cdot \delta(a, x))}{\sum_{a'}\operatorname{exp}(\beta \cdot \delta(a', x))}
    \\
    \pi(a|x) &\triangleq (1 - \epsilon) \cdot \mathbb{I}\left(\delta(a, x) = \underset{a' \in \mathcal{A}}{\sup} \{ \delta(a', x) \}\right) + \frac{\epsilon}{|\mathcal{A}|}
\end{aligned} \end{gather}
Furthermore, to test the practicality of \estimator on a real-world, large-scale data, we also synthesize a bandit-variation of the Movielens-100k dataset \cite{movielens} which consists of numerous $(\texttt{user,} \texttt{item,} \texttt{rating})$ tuples. Taking inspiration from previous recommender system $\rightarrow$ bandit feedback conversion setups \cite{recsys_bandit}, we define a positive reward if the provided rating $\geq 4$, or else zero. We then define contexts and action-embeddings as the user- and item-factors attained by performing SVD on the binary user-item rating matrix, respectively. Furthermore, to simulate continuous instead of binary reward, for missing entries, we define the reward as the dot product of the corresponding user- and item-factors, estimated using SVD before. We define the target policy similarly as in \cref{eqn:synthetic_policies_defn}, and aiming to follow a realistic two-stage recommender system setup \cite{two_stage_ope}, we define the logging policy $\mu(\cdot|x)$ as follows: (1) shortlist a set of $100$ best actions defined by $\delta(\cdot, x)$, and $400$ actions at random; (2) sample a logit from $U(0, 1)$ for each positive action, and from $U(0, 0.8)$ for the random actions; (3) take a temperature softmax as in \cref{eqn:synthetic_policies_defn} only on the sampled logits; and (4) perform $\epsilon-$greedy on the obtained action probabilities to satisfy Assumption \ref{assumption:common_support}. 

As per our setup, $V^*(\mu) \propto \beta$ and $V^*(\pi) \propto \epsilon^{-1}$ for both the datasets. To avoid clutter, we define $\mu_{\mathsf{uniform}}$ when $\beta = 0$, and $\mu_{\mathsf{good}}$ when $\beta=3$. Similarly, we define $\pi_{\mathsf{bad}}$ when $\epsilon = 0.8$, and $\pi_{\mathsf{good}}$ when $\epsilon = 0.05$. Unless specifically mentioned, we use the default values of the remaining hyper-parameters in the bandit data generation procedure, as listed in \cref{appendix:details}. 

For evaluating the performance of various estimators, we compute the Mean Squared Error (MSE) between the true and predicted value of the target policy. We reserve a large test-set just to compute the true value of the target policy. We also estimate the squared bias and variance of our predicted estimates by repeating each experiment for 50 random seeds, and also compute the 95\% confidence interval for visualization purposes. Note that the bias, variance, and MSE of any estimator are naturally linked to each other by the following decomposition: $\operatorname{MSE}(\cdot) = \operatorname{Bias}(\cdot)^2 + \operatorname{Var}(\cdot)$.

For estimators in the \estimator framework, we chose the optimal convolution values (\ie, $\tau_1$ and $\tau_2$) using the MSE obtained on a validation set. Notably, while \estimator for any given backbone estimator strictly contains the na\"ive backbone (\ie, when $\tau_1 = \tau_2 = 0$); to de-confound the effect of policy convolution and the backbone estimator, we only report results for \estimator with a non-zero amount of pooling.

\subsection{Results} \label{sec:results}

\begin{figure*}[!t] 
    \includegraphics[width=0.9\linewidth,trim={0 8.5cm 0 0},clip]{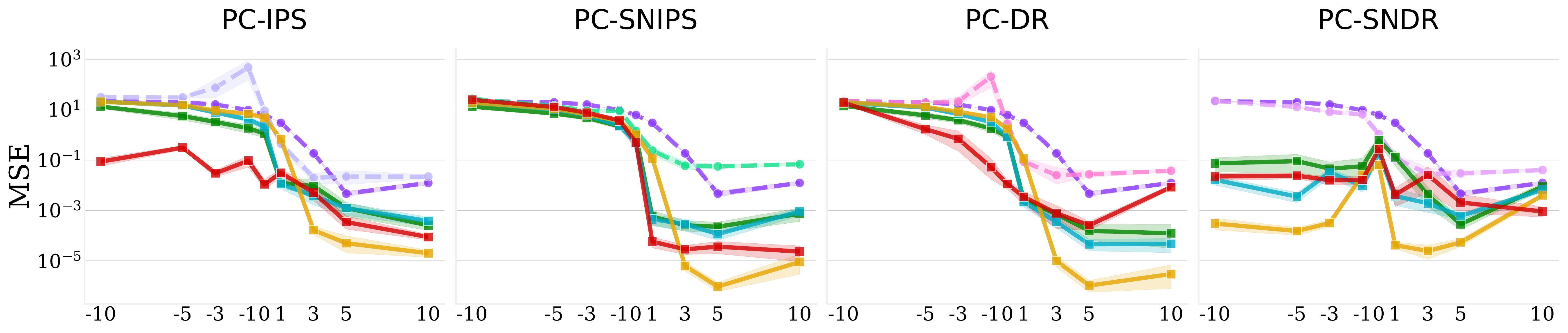} 
    
    \vspace{0.15cm}
    \includegraphics[width=0.9\linewidth,trim={0 0 0 2.5cm},clip]{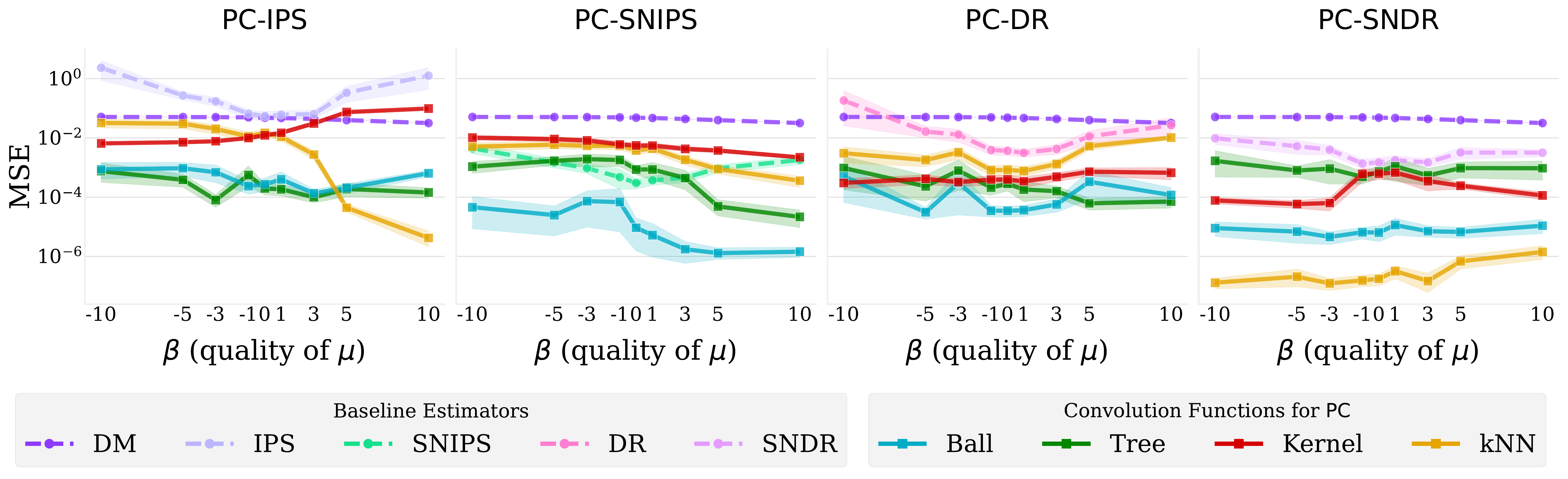}
    \caption{Change in MSE while estimating $V(\pi_{\mathsf{good}})$ with \textbf{varying policy-mismatch} ($\log$ scale) for (top) synthetic, and (bottom) movielens dataset. The policy-mismatch is higher when $\beta$ is lower. Results for estimating $V(\pi_{\mathsf{bad}})$, and the observed bias-variance trade-off can be found in \cref{appendix:experiments}, \cref{fig:v_beta_synthetic_pi_good,fig:v_beta_synthetic_pi_bad,fig:v_beta_ml_pi_good,fig:v_beta_ml_pi_bad}.}
    \label{fig:policy_mismatch}
\end{figure*}

\begin{figure*}[!t]
    \centering
    \vspace{-0.3cm}
    \begin{subfigure}{0.45\textwidth}
        \caption{\oracle-\ips}
        \includegraphics[width=\linewidth,trim={19.5cm 5.2cm 24cm 0},clip]{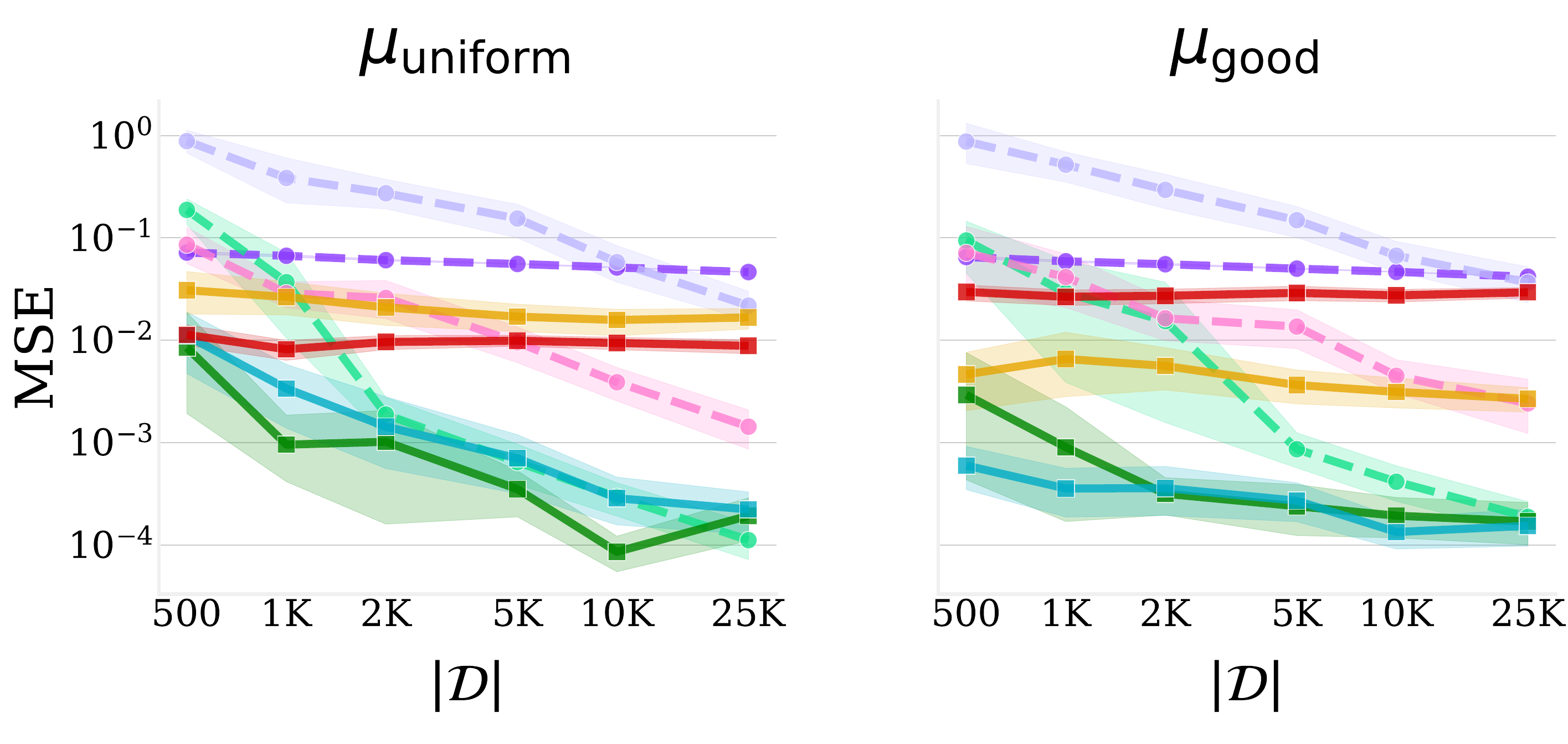}
    \end{subfigure} \hfill %
    \begin{subfigure}{0.45\textwidth}
        \caption{\oracle-\snips}
        \includegraphics[width=\linewidth,trim={19.5cm 5.2cm 24cm 0},clip]{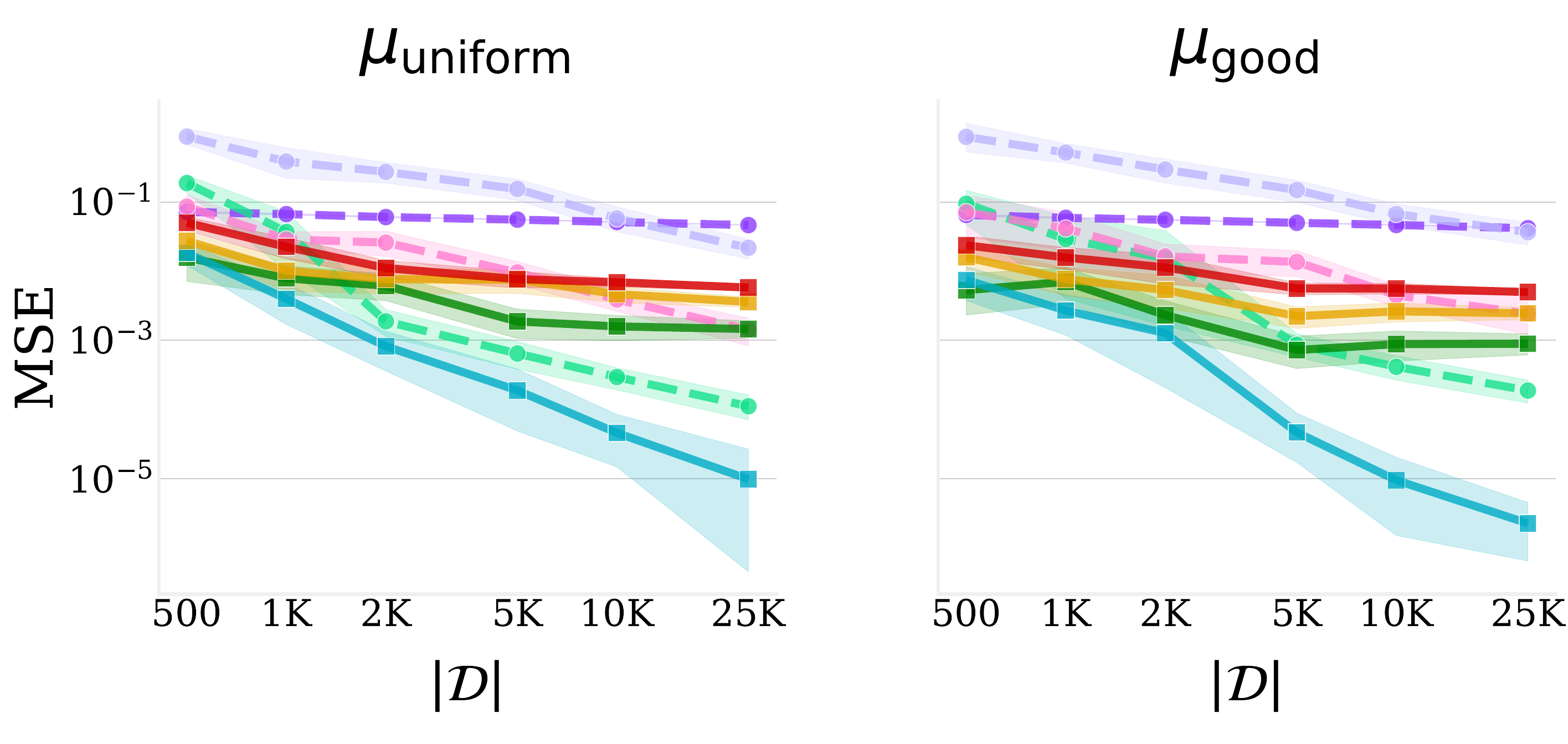}
    \end{subfigure}
    \includegraphics[width=\linewidth,trim={0 0 0 22cm},clip]{figures/custom_n_data/ml-100k/eps_0.02/IPS.pdf}
    \caption{Change in MSE while estimating $V(\pi_{\mathsf{good}})$ with \textbf{varying amounts of bandit feedback} ($\log$-$\log$ scale) for the movielens dataset. Results for \oracle-\dr, \oracle-\sndr, estimating $V(\pi_{\mathsf{bad}})$, and the synthetic dataset can be found in \cref{appendix:experiments}, \cref{fig:v_data_synthetic_pi_good,fig:v_data_synthetic_pi_bad,fig:v_data_ml_pi_good,fig:v_data_ml_pi_bad}.}
    \label{fig:n_data}
\end{figure*}

\begin{figure*}
    \centering
    \includegraphics[width=0.9\linewidth,trim={0 4.4cm 0 0},clip]{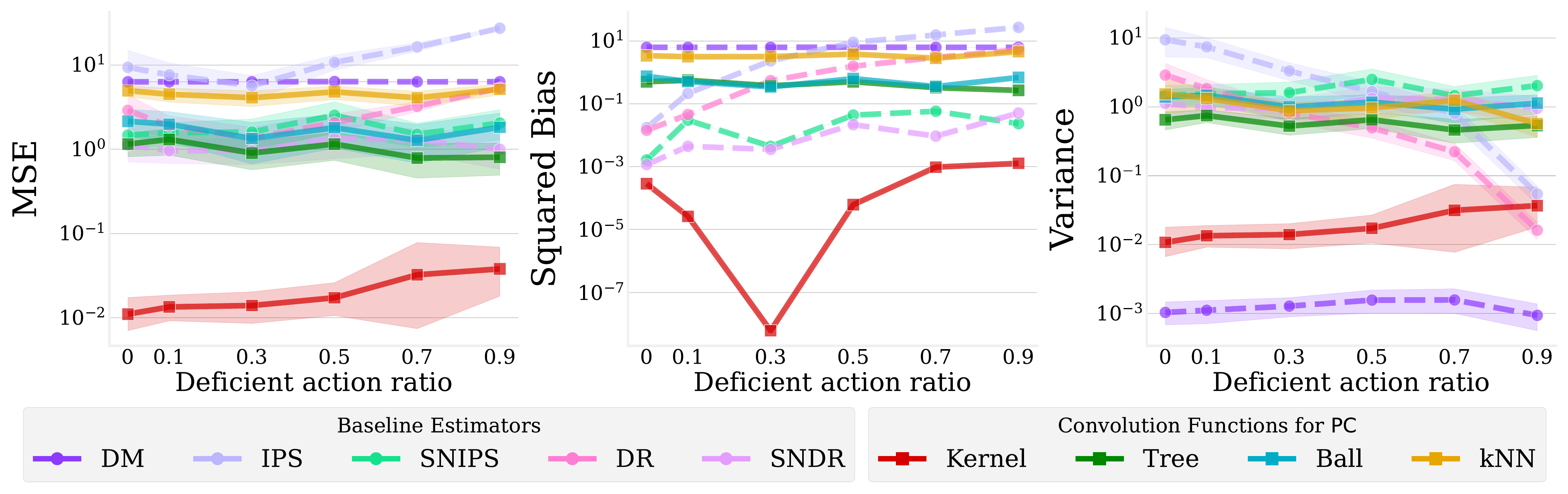}
    \vspace{0.15cm}
    \includegraphics[width=0.9\linewidth,trim={0 0 0 22cm},clip]{figures/varying_beta/movielens_eps_0.05.pdf}
    \caption{Change in MSE, Squared Bias, and Variance for \oracle-\ips and other baseline estimators while estimating $V(\pi_{\mathsf{good}})$ with varying support ($\log$-$\log$ scale) for the synthetic dataset (with $2000$ actions), using data logged by $\mu_{\mathsf{uniform}}$. Results for other backbones in \estimator, and estimating $V(\pi_{\mathsf{bad}})$ can be found in \cref{appendix:experiments}, \cref{fig:v_support_pi_good_mu_unif,fig:v_support_pi_bad_mu_unif}.}
    \label{fig:def_support}
\end{figure*}

\begin{figure*}[!t] 
    \centering
    \includegraphics[width=0.85\linewidth,trim={0 0 0 0},clip]{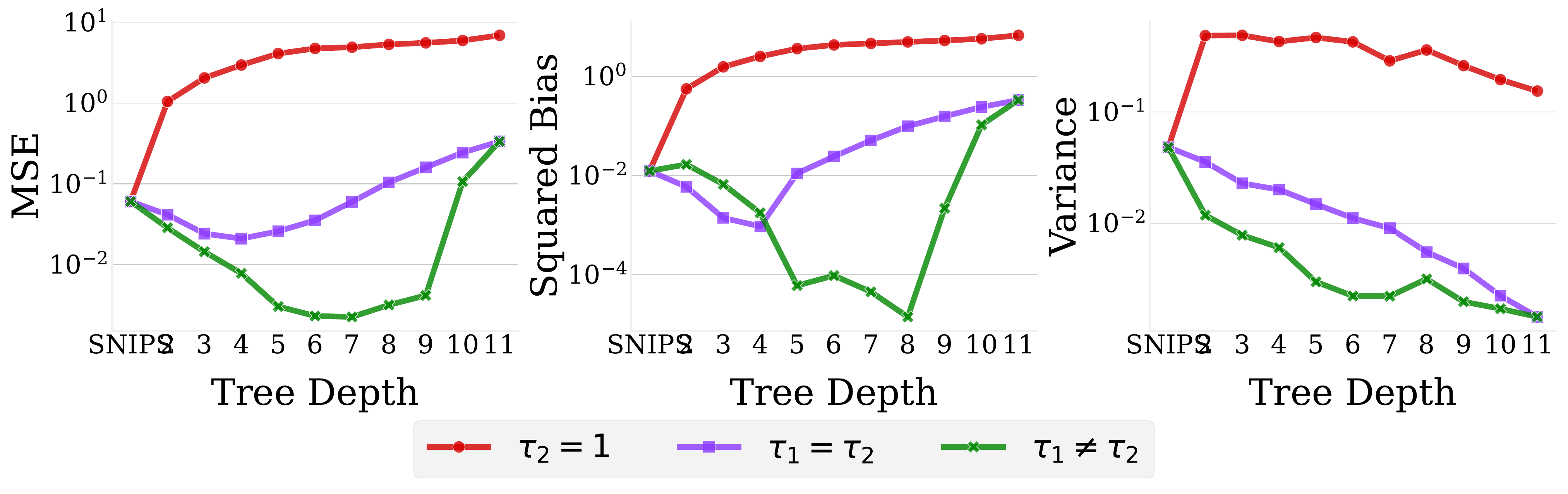} 
    \caption{Visualizing the bias-variance trade-off for \estimator-\snips (Tree pooling) with \textbf{varying amount of pooling}, while estimating $V(\pi_{\mathsf{good}})$ and using $\mu_{\mathsf{good}}$ for logging on the synthetic dataset (with $2000$ actions). When $\tau_1 \neq \tau_2$, we plot results for $\tau_1$ on the plot. The na\"ive \snips estimator is the left-most point, \ie, when there's \textbf{no} pooling. Results for other backbones, pooling methods, and the movielens dataset can be found in \cref{appendix:experiments}, \cref{fig:v_pooling_synthetic_pi_good_mu_good,fig:v_pooling_ml_pi_good_mu_good}.}
    \label{fig:bias_variance}
\end{figure*}

\paragraph{(\cref{fig:n_actions}) How does \estimator perform with a varying number of actions?} 
Observing the effect of increasing the number of actions ($|\mathcal{A}|$) on different estimators' performance while keeping the size of the available bandit feedback ($|\mathcal{D}|$) constant, we find that the MSE of all \is-based estimators deteriorates, in accordance with the $\Omega(|\mathcal{A}|)$ growth of their variance. We further note that utilizing $\mu_{\mathsf{good}}$ rather than $\mu_{\mathsf{uniform}}$ as the logging policy, results in an almost $2$-orders of magnitude reduction of MSE across all estimators due to the increased overlap between $\mu_{\mathsf{good}}$ and $\pi_{\mathsf{good}}$. Finally, a general trend that holds across various convolution strategies is that the improvement in MSE achieved by \estimator \wrt its respective backbone, significantly increases with increasing $|\mathcal{A}|$. Notably, in the extreme scenario of $10$k actions, \estimator-\dr outperforms \dr by up to $5$-orders of magnitude in terms of MSE. While we note that no single backbone estimator or pooling strategy is optimal in every scenario for \estimator, \estimator-\sndr and kernel \emph{or} kNN convolution strategies generally exhibit better performance than others.

\paragraph{(\cref{fig:policy_mismatch}) How does \estimator perform with varying amount of policy-mismatch?}
We analyze the impact of increasing the policy mismatch between the target and logging policies on off-policy estimation performance, specifically by tuning the $\beta$ parameter in \cref{eqn:synthetic_policies_defn} that controls the quality of the logging policy, keeping the target policy fixed. We observe that OPE becomes hardest on both ends of the spectrum, \ie, when the logging and target policies have large divergence.
The difficulties of OPE in such low-overlap scenarios have been well documented in the literature \cite{ips_support, pessimism_1, curse_ope}, and we observe that \estimator---through the use of latent structure amongst actions---is particularly helpful in such conditions. An analysis of the exact bias-variance trade-off provided in \cref{appendix:experiments}, \cref{fig:v_beta_synthetic_pi_bad} reveals that \estimator is able to effectively (1) reduce the variance when policy-mismatch is low, \ie, $|\beta| \approx 0$, and (2) counteract the bias introduced by \is when Assumption \ref{assumption:common_support} is violated, \ie, $|\beta|$ is large. Both of these improvements result in significantly better MSE for \estimator across the entire policy-mismatch spectrum, as depicted in \cref{fig:policy_mismatch}.

\paragraph{(\cref{fig:n_data}) How does \estimator perform with a varying amount of bandit feedback?}
Keeping all other factors fixed, we investigate the impact of increasing the size of the logged bandit feedback ($|\mathcal{D}|$) on the MSE of various off-policy estimators. Like other baseline estimators, we observe that \estimator exhibits consistency and is effectively able to balance the bias-variance trade-off.
Specifically, \estimator is most advantageous in the low-data regime, due to its variance reduction properties. However, as $|\mathcal{D}|$ continues to increase, \estimator converges to its respective backbone estimator, \ie, when $\tau_1=\tau_2=0$ represents the optimal bias-variance trade-off point. This pattern of a decreasing amount of optimal pooling ($\tau_1$, $\tau_2$) with increasing $|\mathcal{D}|$ is anticipated, as the variance of \is-based estimators naturally decreases with growing $|\mathcal{D}|$, and any reduction in variance at the cost of increased bias negatively impacts the overall MSE. We take further note of this observation for even more kinds of logging policies and backbone estimators in \cref{appendix:experiments}, \cref{fig:synthetic_pi_good_mu_unif,fig:synthetic_pi_good_mu_bad,fig:synthetic_pi_good_mu_good}.

\paragraph{(\cref{fig:def_support}) How does \estimator perform with a varying amount of deficient support?}
To understand the effect of varying amount of support (or overlap) between the logging and target policies on various estimators, we simulate such a scenario by explicitly forcing the logging policy to only have support (\ie, non-zero probability) over a smaller, random set of actions, and have zero probability for all other actions. Varying this deficient action ratio, we firstly observe an expected increase in the MSE and squared bias of baseline estimators like \ips, \snips, \dr, \etc due to the violation of Assumption \ref{assumption:common_support}, which has been shown to to add irrecoverable bias in importance sampling based estimators \cite{ips_support}. On the other hand, even with an increasing number of deficient actions, the MSE for \estimator tends to stay relatively constant, with kernel-based convolutions being the best approach. This goes to show that \estimator is able to accurately leverage action-embeddings as a guide for appropriately filling-in the blind spots while performing OPE.

\paragraph{(\cref{fig:bias_variance}) How does the amount of convolution affect the bias-variance trade-off?}
We examine the influence that the amount of convolution in \estimator has on the bias-variance trade-off for three variations of \estimator with the tree pooling function: (1) only the target policy is convolved, \ie, $\tau_2 = 1$ which is equivalent to the similarity estimator \cite{similarity_estimator}; (2) both the logging and target policies are convolved equally, \ie, $\tau_1 = \tau_2$ which is equivalent to the offCEM \cite{saito2023off} and groupIPS \cite{peng2023offline} estimators; and (3) both the logging and target policies are convolved and $\tau_1, \tau_2$ can be different. 
From \cref{fig:bias_variance}, we observe that the amount of convolution results in a bias-variance trade-off, where larger pooling leads to decreased variance, but  increased bias. 
It is worth mentioning that the initial decrease in bias with convolution is due to the use of $\mu_{\mathsf{good}}$ for logging, which partially violates Assumption \ref{assumption:common_support}. This results in a biased \snips estimate, which the \estimator family of estimators are able to effectively recover.
Further, we note that solely convolving the target policy (\ie, the similarity estimator) does not necessarily result in a suitable bias-variance trade-off, with the other two convolution strategies being significantly better, and having $\tau_1 \neq \tau_2$ consistently being the best approach.

\section{Related Work}
\paragraph{Off-policy evaluation.} 
A wide body of literature in operations research, causal inference, and reinforcement learning studies the problem of off-policy evaluation. 
Prominent off-policy estimators can be grouped into the following three categories: \textbf{(1) Model-based:} dubbed as the direct method (\dm), whose key idea is to use a parametric reward model to extrapolate the reward for unobserved (context, action) pairs \cite{dr}. \dm typically has a low variance, at the cost of uncontrollable bias due to model misspecification. \textbf{(2) Inverse propensity scoring (\ips)}: IPS uses the propensity ratio between the target and logging policies to account for the distribution mismatch \cite{ips}. Though unbiased under mild assumptions, \ips suffers from large variance.  Typical remedies for the large variance are propensity clipping \cite{clipped_ips, cab} or self-normalization \cite{snips}, which might introduce bias. \textbf{(3) Hybrid:} some estimators (e.g, the doubly robust estimator~\cite{dr}) combine \dm and \ips together to leverage the benefits of both worlds~\cite{dr, dr_shrinkage, dr_ips_variance, mrdr, sndr1, sndr2}. However, these estimators still suffer from the large variance problem due to the large propensities especially when the action space is large. 

\paragraph{Off-policy evaluation for large action spaces.}
Two kinds of major problems occur when attempting to perform OPE in large action spaces. Firstly, the variance of any importance sampling method grows linearly \wrt the size of the action-space \cite{ope_slate}, and the common support assumption tends to become impractical \cite{ips_support} leading to irrecoverable bias in estimation. Recent work \cite{mips, saito2023off, peng2023offline, similarity_estimator} attempts to use some notion of latent structure in the action-space to address both of the aforementioned limitations. The MIPS estimator \cite{mips} builds on the randomness in the available action embeddings to improve OPE. However, in a setting where only a 1:1 action-to-embedding mapping is available (as in this paper), MIPS reduces to vanilla \ips. Further, as we discussed in \cref{sec:pc}, offCEM \cite{saito2023off}, similarity estimator \cite{similarity_estimator}, and groupIPS \cite{peng2023offline} are all specific instantiations of our \estimator family of estimators. 

\paragraph{Off-policy evaluation for continuous action spaces.}
Another line of work builds off-policy estimators when the action-space is continuous, \eg, the dosage of a treatment. 
If we discretize the action-space into a fixed number of bins as per some resolution, the action-space becomes too large for typical off-policy estimators to work well \cite{discrete_continuous_action_space}. Naive use of importance sampling based estimators would be vacuous in this setting, since the probability of selecting any action can be zero for a policy that samples actions according to some probability density function. 
To this end, typical approaches extend the discrete rejection sampling idea into a smooth rejection operation using standard kernel functions \cite{kernel_ips, kernel_metric_learning_ope, wang2023oracle}, with the implicit assumption that similar actions (in terms of distance in the continues action space) should lead to similar reward. Our proposed \estimator also leverages the similarity information between actions through action embeddings, but for problem with discrete and large action spaces. 


\section{Conclusion \& Future Work}
In this paper, we proposed the \estimatorfull (\estimator) family of estimators which leverage latent action structure specified via action embeddings to perform off-policy evaluation in large action spaces. More specifically, \estimator convolves both the target and logging policies according to an action-action convolution function, which posits a new kind of bias-variance tradeoff controlled by the amount of convolution.

Conducting empirical evaluation over a diverse set of off-policy estimation scenarios, we observe that the estimators from the \estimator framework enjoy up to 5 orders of magnitude improvement over existing baseline estimators in terms of MSE, especially when (1) the action-space is large, (2) the policy mismatch between logging and target policies is high, or (3) the common support assumption for importance sampling is violated.
We believe that our findings can expand the potential use of off-policy estimators into new and practical scenarios, and also encourage further exploration into the use of additional structure for efficient OPE.

We also discuss limitations and unexplored directions in this paper that we believe are promising for future work. Firstly, having a deeper formal understanding about the statistical properties of \estimator might help in designing more robust off-policy estimators. Next, even though we propose four different action convolution functions, having a better understanding of the inductive biases that various convolution functions posit might guide us in designing even better and more principled OPE approaches. Finally, understanding and developing principled techniques for automatically selecting the level of convolution to conduct on the target and logging policies is an interesting research direction \cite{slope, model_selection_offline_rl}.

\bibliographystyle{plain}
{\small
\bibliography{references}}

\clearpage
\appendix
\section{Appendix: Further Details} \label{appendix:pc_estimators}

For the sake of clarity, we provide a formal definition of \estimator with using Self-Normalized IPS (\snips, \cref{sec:snips}), Doubly Robust (\dr, \cref{sec:dr}), Self-Normalized DR (\sndr, \cref{sec:sndr}) as backbone estimators:

\begin{itemize}[leftmargin=0.2in]
    \listheader{\estimator-\snips.} Defined as follows:
    \begin{align*}
        \hat{V}_{\estimator-\snips}(\pi) &\triangleq \expectation{(x,a,r) \sim \mathcal{D}}{\frac{(\pi(\cdot|x) \ast f_{\tau_1})(a)}{\rho \cdot (\mu(\cdot|x) \ast f_{\tau_2})(a)} \cdot r} \\
        \text{s.t.} ~~ \rho &\triangleq \expectation{(x,a,\cdot) \sim \mathcal{D}}{\frac{(\pi(\cdot|x) \ast f_{\tau_1})(a)}{(\mu(\cdot|x) \ast f_{\tau_2})(a)}}
        .
    \end{align*}

    \listheader{\estimator-\dr.} Defined as follows:

    \begin{align*}
        \hat{V}_{\estimator-\dr}(\pi) &\triangleq  \expectation{(x,a,r) \sim \mathcal{D}}{\frac{(\pi(\cdot|x) \ast f_{\tau_1})(a)}{(\mu(\cdot|x) \ast f_{\tau_2})(a)} \cdot (r - \hat{\delta}(a, x)) + \Delta(\pi, x)} \\
        \text{s.t.} ~~~~ \Delta(\pi, x) &\triangleq \sum_{a' \in \mathcal{A}} \pi(a'|x) \cdot \hat{\delta}(a', x)
        .
    \end{align*}
\end{itemize}

\ 

\section{Appendix: Hyper-Parameters} \label{appendix:details}

\def\arraystretch{1.3}
\begin{center}
\begin{tabularx}{\textwidth}{XX|XX|XX}
    \hline
    \textbf{Hyper-param} & \textbf{Value} & \textbf{Hyper-param} & \textbf{Value} & \textbf{Hyper-param} & \textbf{Value} \\
    \hline
    $|\mathcal{A}|$ & 2,000 & $|\mathcal{D}|$ & 10,000 & | Test data | & 100,000 \\
    $\beta$ & 0.0 & $\epsilon$ & 0.05 & \# Seeds & 50 \\
    $\dim($context$)$ & 32 & $\dim($action-embed$)$ & 16 & $\dim($noise$)$ & 8 \\
    \hline
\end{tabularx}
\end{center}

\newpage

~

\begin{itemize}[leftmargin=0.2in]
    \listheader{\estimator-\sndr.} Defined as follows:
    \begin{align*}
        \hat{V}_{\estimator-\sndr}(\pi) &\triangleq \expectation{(x,a,r) \sim \mathcal{D}}{\frac{(\pi(\cdot|x) \ast f_{\tau_1})(a)}{\rho \cdot (\mu(\cdot|x) \ast f_{\tau_2})(a)} \cdot (r - \hat{\delta}(a, x)) + \Delta(\pi, x)} \\
        \text{s.t.} ~~~~ \rho &\triangleq \expectation{(x,a,\cdot) \sim \mathcal{D}}{\frac{(\pi(\cdot|x) \ast f_{\tau_1})(a)}{(\mu(\cdot|x) \ast f_{\tau_2})(a)}} \\
        \text{s.t.} ~~~~ \Delta(\pi, x) &\triangleq \sum_{a \in \mathcal{A}} \pi(a|x) \cdot \hat{\delta}(a, x)
        .
    \end{align*}
\end{itemize}

\clearpage

\section{Appendix: Additional Results} \label{appendix:experiments}

\newcommand{\figref}[1]{Page \pageref{#1}, \cref{#1}}
\def\arraystretch{1.15}
\begin{center}
\begin{tabularx}{\linewidth}{>{\raggedleft\arraybackslash}m{1.03in} m{1.2in} m{2.3in} m{1.1in}}
    \\[0.1in]
    & \multicolumn{2}{c}{\hspace{1cm}\textbf{Varying Number of Actions $(|\mathcal{A}|)$}} \\[0.1in]

    1. & Synthetic Dataset & Estimating $V(\pi_{\mathsf{bad}})$ using $\mu_{\mathsf{uniform}}$ \dotfill & \figref{fig:v_actions_pi_bad_mu_unif} \\
    2. & Synthetic Dataset & Estimating $V(\pi_{\mathsf{bad}})$ using $\mu_{\mathsf{good}}$ \dotfill & \figref{fig:v_actions_pi_bad_mu_good} \\

    \\[-0.01in] & \multicolumn{2}{c}{\hspace{1cm}\textbf{Varying Policy-mismatch through $\mu (\beta)$}} \\[0.1in]
    
    1. & Synthetic Dataset & Estimating $V(\pi_{\mathsf{good}})$ \dotfill & \figref{fig:v_beta_synthetic_pi_good} \\
    2. & Synthetic Dataset & Estimating $V(\pi_{\mathsf{bad}})$ \dotfill & \figref{fig:v_beta_synthetic_pi_bad} \\
    3. & Movielens Dataset & Estimating $V(\pi_{\mathsf{good}})$ \dotfill & \figref{fig:v_beta_ml_pi_good} \\
    4. & Movielens Dataset & Estimating $V(\pi_{\mathsf{bad}})$ \dotfill & \figref{fig:v_beta_ml_pi_bad} \\

    \\[-0.01in] & \multicolumn{2}{c}{\hspace{1cm}\textbf{Varying Policy-mismatch  through $\pi (\epsilon)$}} \\[0.1in]

    1. & Synthetic Dataset & Using $\mu_{\mathsf{uniform}}$ \dotfill & \figref{fig:v_eps_synthetic_mu_unif} \\
    2. & Synthetic Dataset & Using $\mu_{\mathsf{good}}$ \dotfill & \figref{fig:v_eps_synthetic_mu_good} \\
    3. & Movielens Dataset & Using $\mu_{\mathsf{uniform}}$ \dotfill & \figref{fig:v_eps_ml_mu_unif} \\
    4. & Movielens Dataset & Using $\mu_{\mathsf{good}}$ \dotfill & \figref{fig:v_eps_ml_mu_good} \\

    \\[-0.01in] & \multicolumn{2}{c}{\hspace{1cm}\textbf{Varying Bandit Feedback $(|\mathcal{D}|)$}} \\[0.1in]

    1. & Synthetic Dataset & Estimating $V(\pi_{\mathsf{good}})$ \dotfill & \figref{fig:v_data_synthetic_pi_good} \\
    2. & Synthetic Dataset & Estimating $V(\pi_{\mathsf{bad}})$ \dotfill & \figref{fig:v_data_synthetic_pi_bad} \\
    3. & Movielens Dataset & Estimating $V(\pi_{\mathsf{good}})$ \dotfill & \figref{fig:v_data_ml_pi_good} \\
    4. & Movielens Dataset & Estimating $V(\pi_{\mathsf{bad}})$ \dotfill & \figref{fig:v_data_ml_pi_bad} \\
    
    \\[-0.01in] & \multicolumn{2}{c}{\hspace{1cm}\textbf{Varying Deficient Support}} \\[0.1in]

    1. & Synthetic Dataset & Estimating $V(\pi_{\mathsf{good}})$ using $\mu_{\mathsf{uniform}}$ \dotfill & \figref{fig:v_support_pi_good_mu_unif} \\
    2. & Synthetic Dataset & Estimating $V(\pi_{\mathsf{bad}})$ using $\mu_{\mathsf{uniform}}$ \dotfill & \figref{fig:v_support_pi_bad_mu_unif} \\

    \\[-0.01in] & \multicolumn{2}{c}{\hspace{1cm}\textbf{Varying Action Embedding Size}} \\[0.1in]

    1. & Synthetic Dataset & Estimating $V(\pi_{\mathsf{good}})$ using $\mu_{\mathsf{uniform}}$ \dotfill & \figref{fig:v_embed_size_synthetic_pi_good_mu_unif} \\
    2. & Movielens Dataset & Estimating $V(\pi_{\mathsf{good}})$ using $\mu_{\mathsf{uniform}}$ \dotfill & \figref{fig:v_embed_size_ml_pi_good_mu_unif} \\

    \\[-0.01in] & \multicolumn{2}{c}{\hspace{1cm}\textbf{Data-Driven \vs Oracle Action Embeddings}} \\[0.1in]

    1. & Synthetic Dataset & Estimating $V(\pi_{\mathsf{good}})$ using $\mu_{\mathsf{uniform}}$ \dotfill & \figref{fig:v_embed_type_pi_good_mu_unif} \\
    2. & Synthetic Dataset & Estimating $V(\pi_{\mathsf{bad}})$ using $\mu_{\mathsf{uniform}}$ \dotfill & \figref{fig:v_embed_type_pi_bad_mu_unif} \\

    \\[-0.01in] & \multicolumn{2}{c}{\hspace{1cm}\textbf{Varying Amount of Pooling}} \\[0.1in]

    1. & Synthetic Dataset & Estimating $V(\pi_{\mathsf{good}})$ using $\mu_{\mathsf{good}}$ \dotfill & \figref{fig:v_pooling_synthetic_pi_good_mu_good} \\
    2. & Movielens Dataset & Estimating $V(\pi_{\mathsf{good}})$ using $\mu_{\mathsf{good}}$ \dotfill & \figref{fig:v_pooling_ml_pi_good_mu_good} \\

    \\[-0.01in] & \multicolumn{2}{c}{\hspace{1cm}\textbf{Optimal Amount of Pooling}} \\[0.1in]

    1. & Synthetic Dataset & Estimating $V(\pi_{\mathsf{good}})$ using $\mu_{\mathsf{bad}}$ \dotfill & \figref{fig:synthetic_pi_good_mu_bad} \\
    2. & Synthetic Dataset & Estimating $V(\pi_{\mathsf{good}})$ using $\mu_{\mathsf{uniform}}$ \dotfill & \figref{fig:synthetic_pi_good_mu_unif} \\
    3. & Synthetic Dataset & Estimating $V(\pi_{\mathsf{good}})$ using $\mu_{\mathsf{good}}$ \dotfill & \figref{fig:synthetic_pi_good_mu_good} \\
\end{tabularx}
\end{center}

\newpage

\begin{figure*}
\begin{minipage}[c][\textheight][c]{\textwidth}
    \centering
    \begin{subfigure}{0.49\textwidth}
        \caption{Bias-Variance trade-off for \oracle-\ips}
        \includegraphics[width=\linewidth,trim={23cm 5.5cm 0 0},clip]{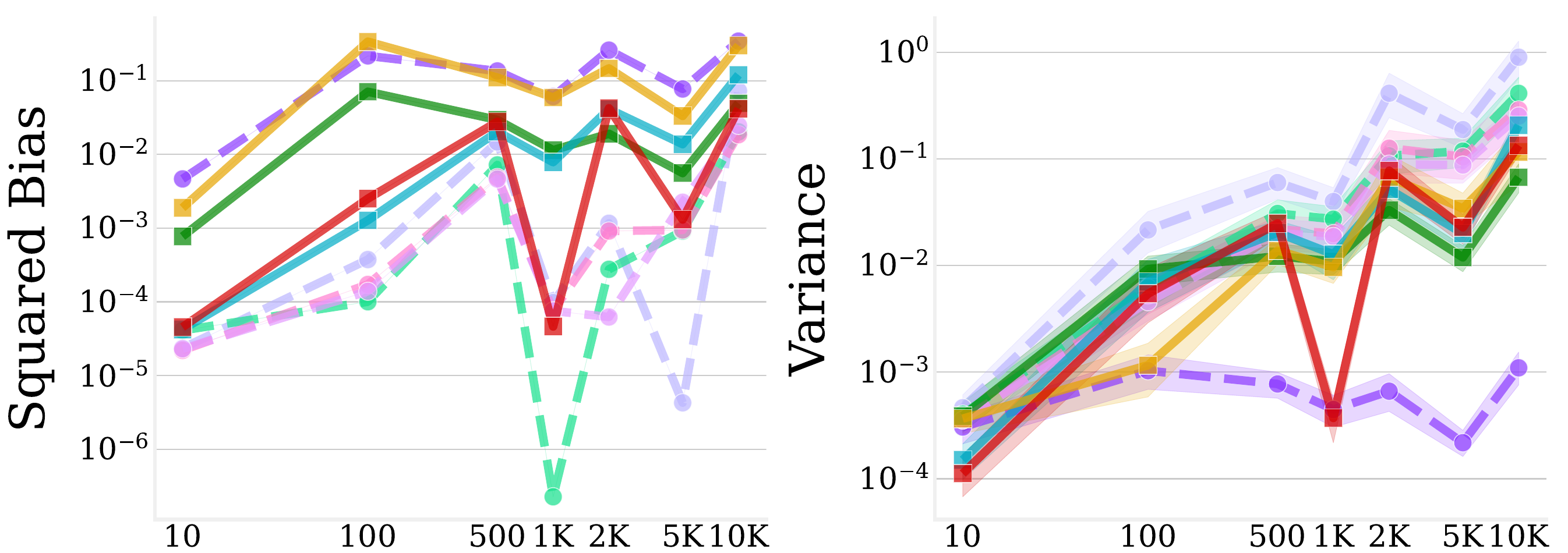}
    \end{subfigure} \hfill %
    \begin{subfigure}{0.49\textwidth}
        \caption{Bias-Variance trade-off for \oracle-\snips}
        \includegraphics[width=\linewidth,trim={23cm 5.5cm 0 0},clip]{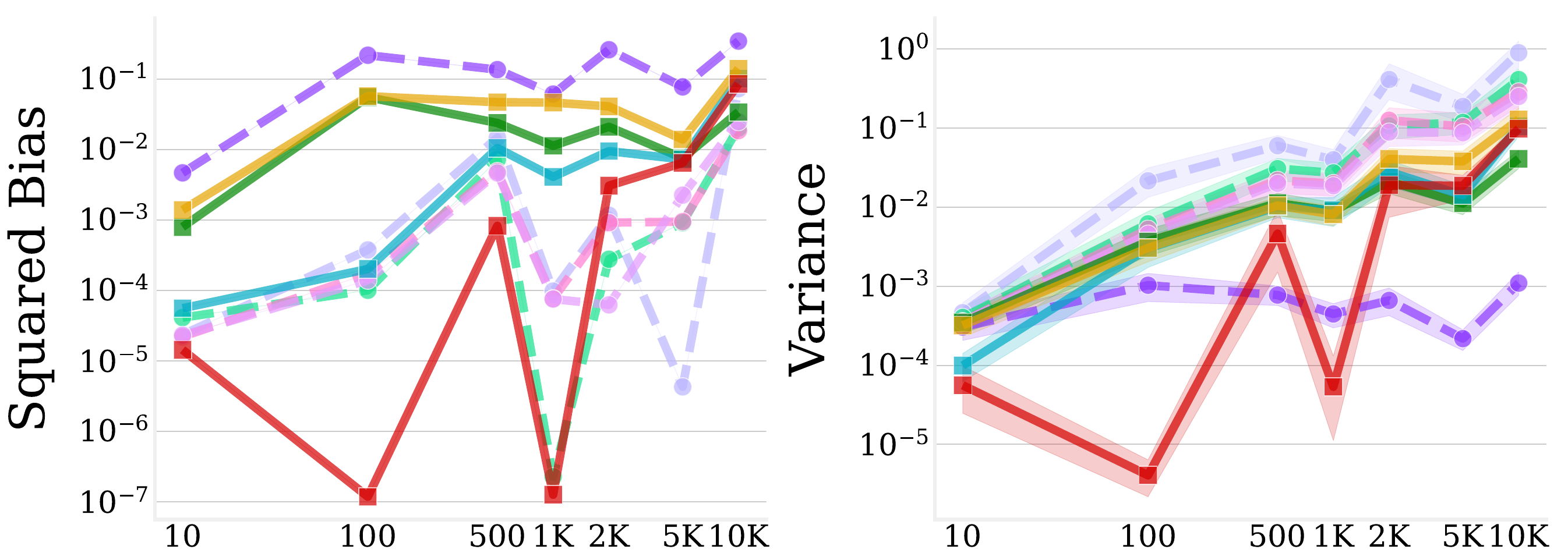}
    \end{subfigure}
    \begin{subfigure}{0.49\textwidth}
        \caption{Bias-Variance trade-off for \oracle-\dr}
        \includegraphics[width=\linewidth,trim={23cm 5.5cm 0 0},clip]{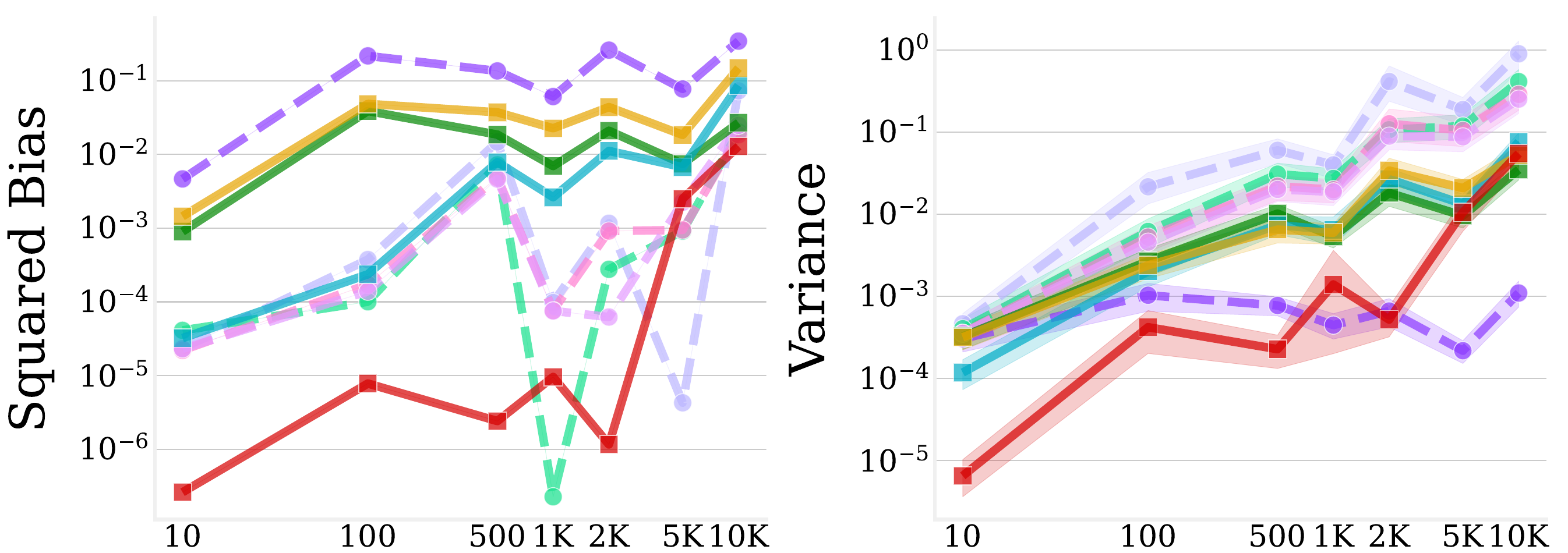}
    \end{subfigure} \hfill %
    \begin{subfigure}{0.49\textwidth}
        \caption{Bias-Variance trade-off for \oracle-\sndr}
        \includegraphics[width=\linewidth,trim={23cm 5.5cm 0 0},clip]{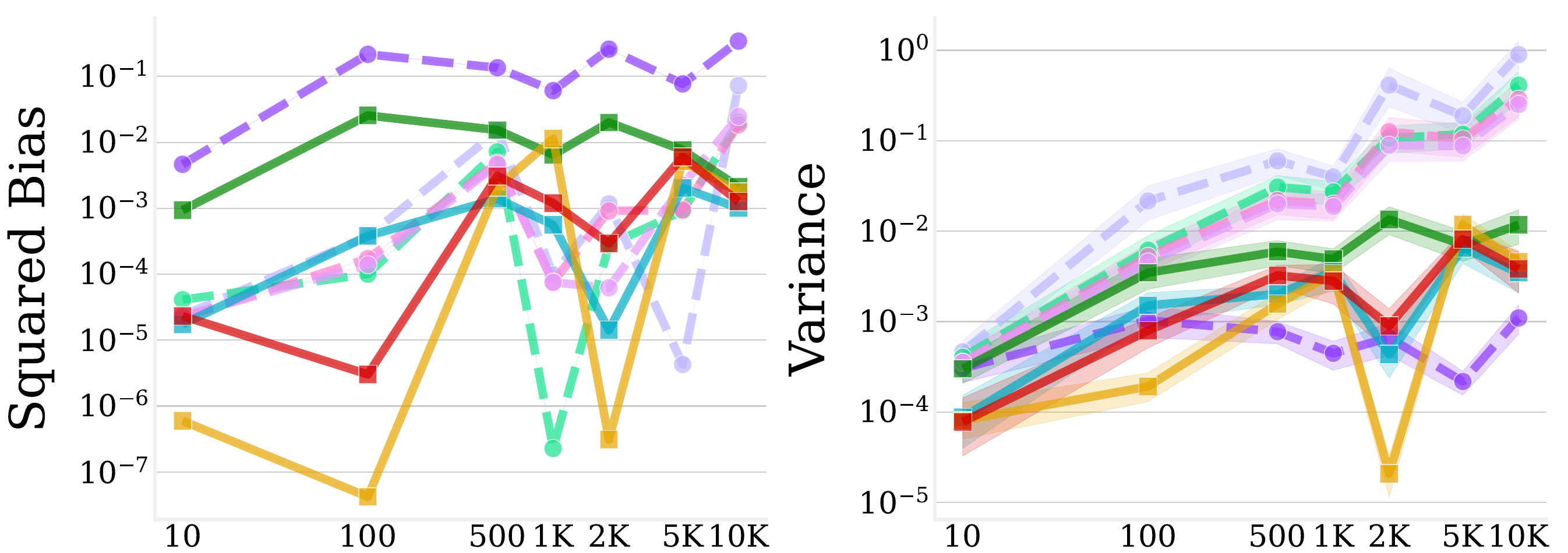}
    \end{subfigure}

    \vspace{0.3cm}
    
    \begin{subfigure}{\textwidth}
        \includegraphics[width=\linewidth,trim={0 0 0 0},clip]{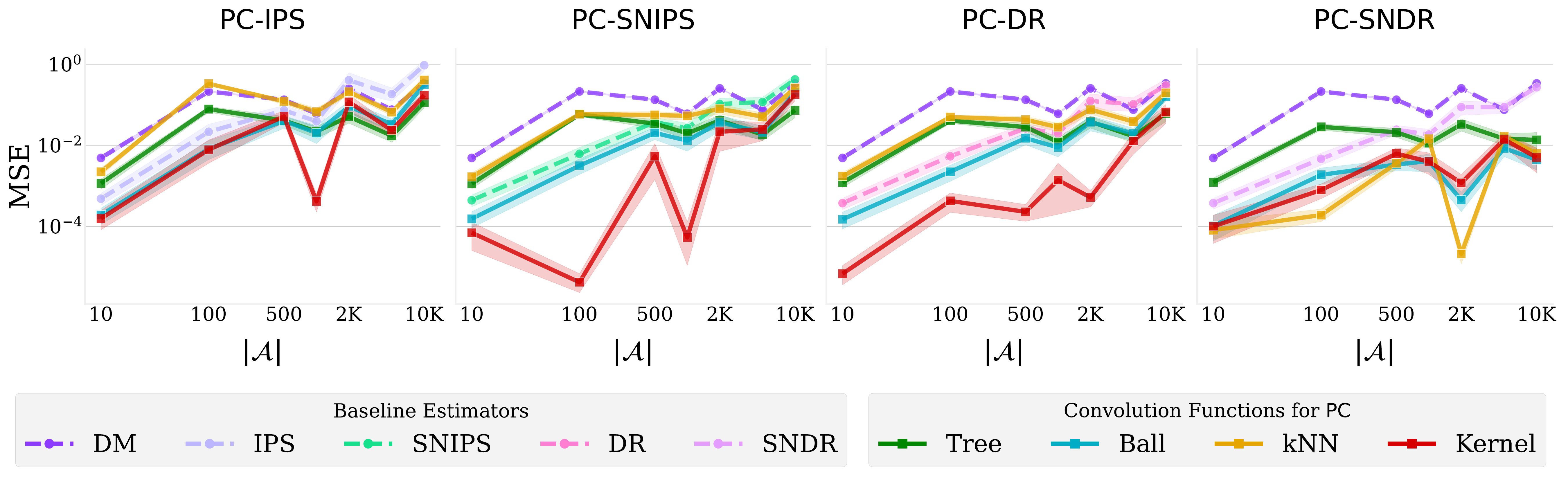}
    \end{subfigure} 
    \caption{Change in MSE, Squared Bias, and Variance while estimating $V(\pi_{\mathsf{bad}})$ with varying number of actions ($\log$-$\log$ scale) for the synthetic dataset, using data logged by $\mu_{\mathsf{uniform}}$.}
    \label{fig:v_actions_pi_bad_mu_unif}
\end{minipage}
\end{figure*}

\begin{figure*}
\begin{minipage}[c][\textheight][c]{\textwidth}
    \centering
    \begin{subfigure}{0.49\textwidth}
        \caption{Bias-Variance trade-off for \oracle-\ips}
        \includegraphics[width=\linewidth,trim={23cm 5.5cm 0 0},clip]{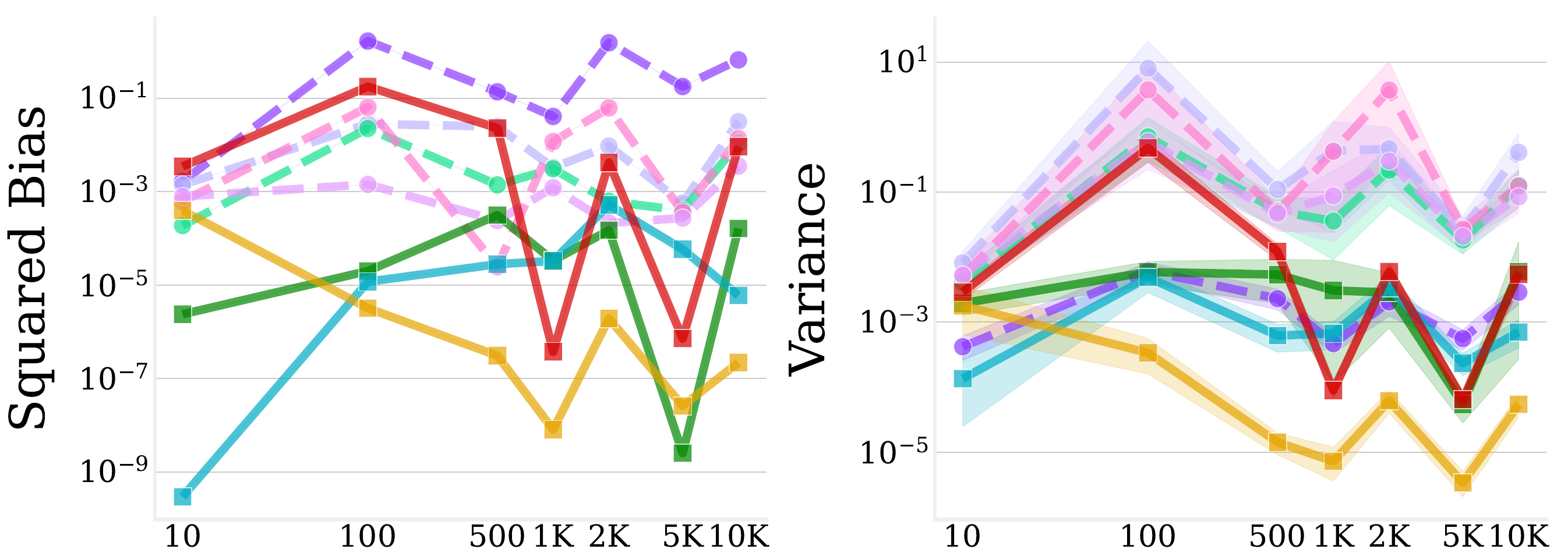}
    \end{subfigure} \hfill %
    \begin{subfigure}{0.49\textwidth}
        \caption{Bias-Variance trade-off for \oracle-\snips}
        \includegraphics[width=\linewidth,trim={23cm 5.5cm 0 0},clip]{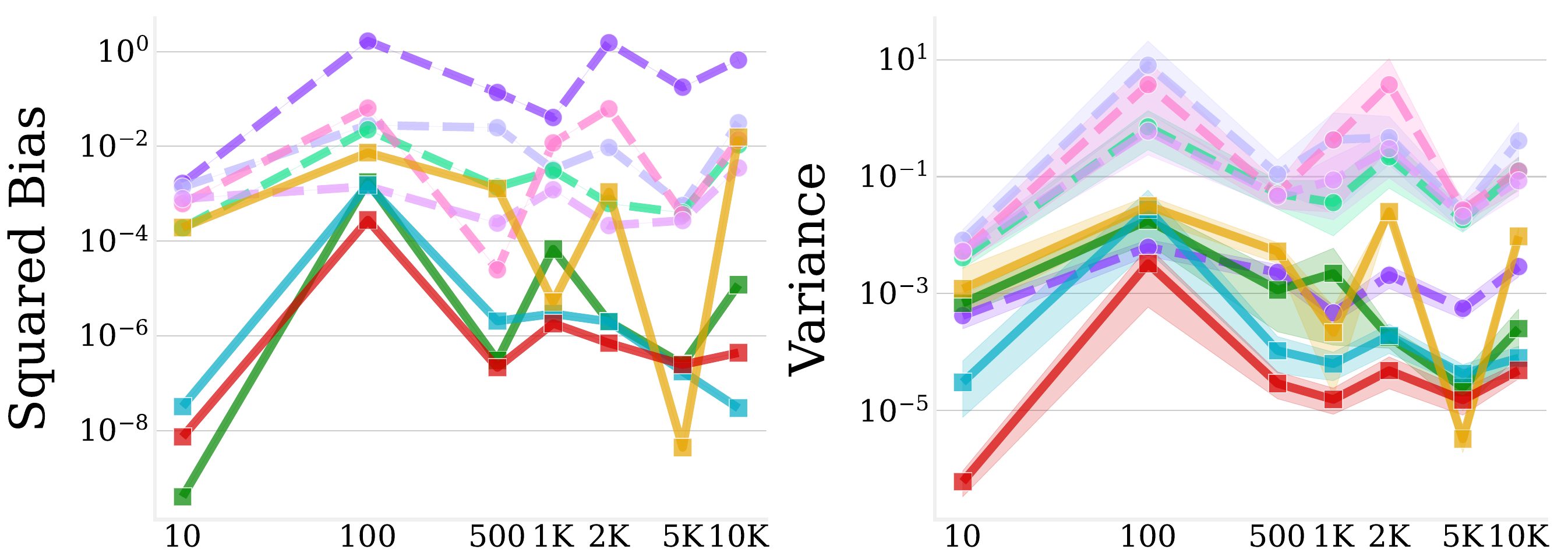}
    \end{subfigure}
    \begin{subfigure}{0.49\textwidth}
        \caption{Bias-Variance trade-off for \oracle-\dr}
        \includegraphics[width=\linewidth,trim={23cm 5.5cm 0 0},clip]{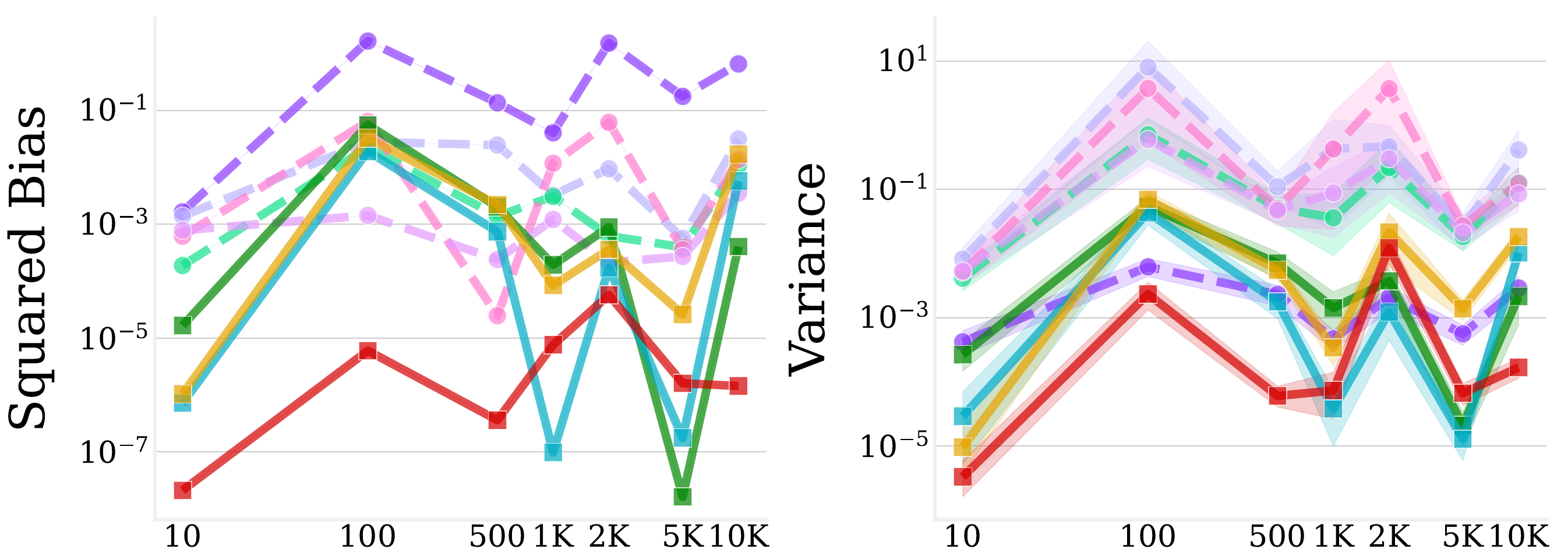}
    \end{subfigure} \hfill %
    \begin{subfigure}{0.49\textwidth}
        \caption{Bias-Variance trade-off for \oracle-\sndr}
        \includegraphics[width=\linewidth,trim={23cm 5.5cm 0 0},clip]{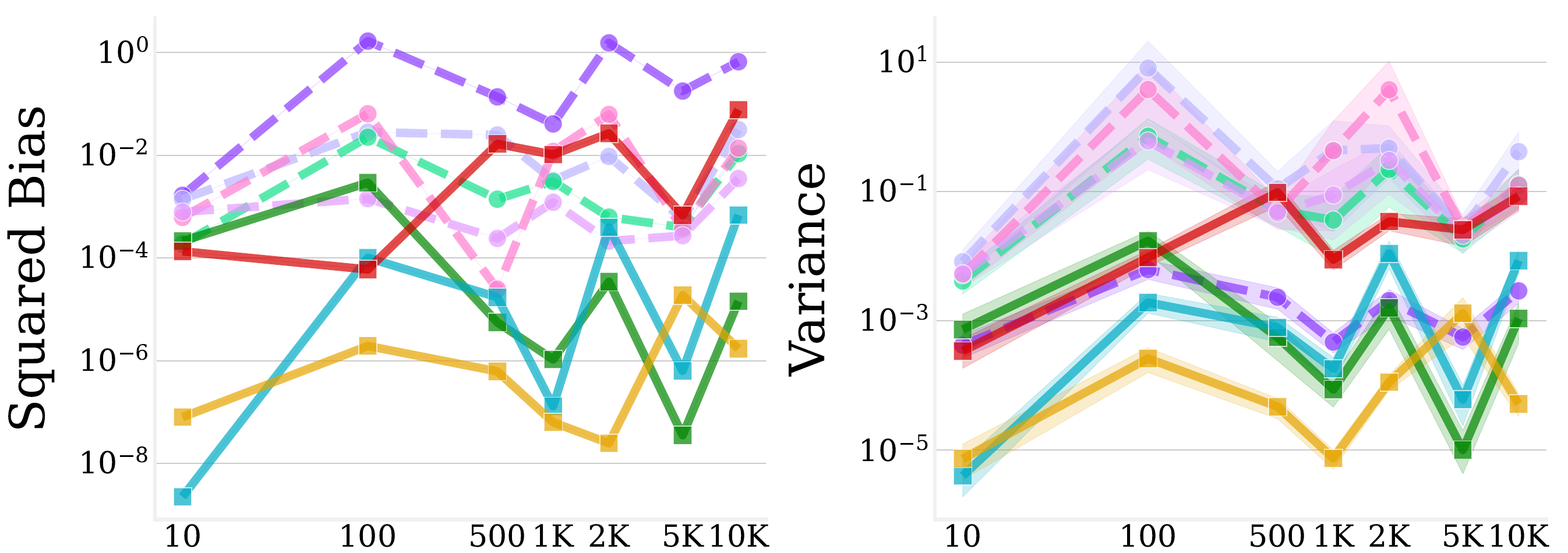}
    \end{subfigure}

    \vspace{0.3cm}
    
    \begin{subfigure}{\textwidth}
        \includegraphics[width=\linewidth,trim={0 0 0 0},clip]{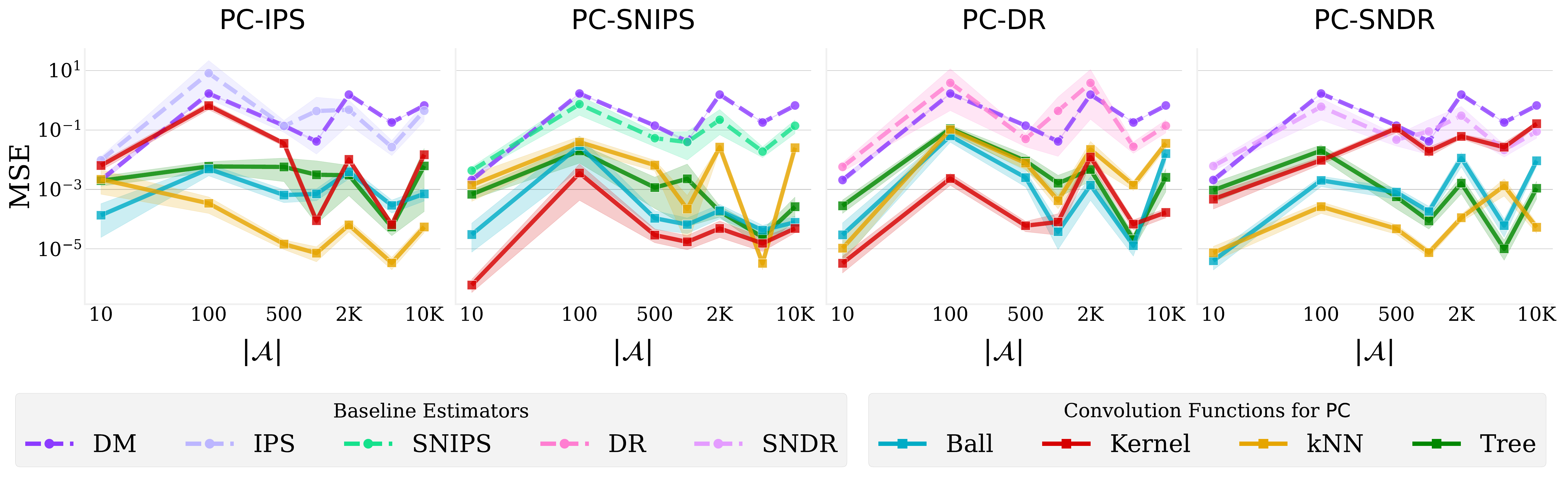}
    \end{subfigure} 
    \caption{Change in MSE, Squared Bias, and Variance while estimating $V(\pi_{\mathsf{bad}})$ with varying number of actions ($\log$-$\log$ scale) for the synthetic dataset, using data logged by $\mu_{\mathsf{good}}$.}
    \label{fig:v_actions_pi_bad_mu_good}
\end{minipage}
\end{figure*}

\clearpage

\begin{figure*}
\begin{minipage}[c][\textheight][c]{\textwidth}
    \centering
    \begin{subfigure}{0.49\textwidth}
        \caption{Bias-Variance trade-off for \oracle-\ips}
        \includegraphics[width=\linewidth,trim={23cm 5.5cm 0 0},clip]{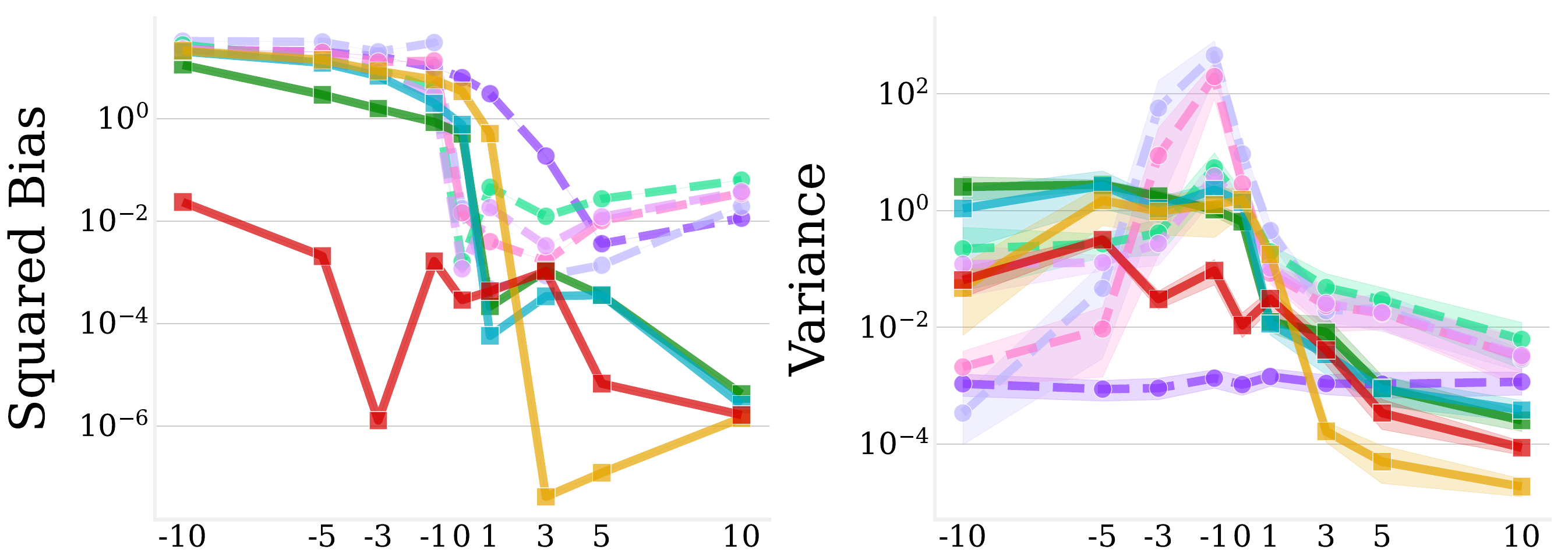}
    \end{subfigure} \hfill %
    \begin{subfigure}{0.49\textwidth}
        \caption{Bias-Variance trade-off for \oracle-\snips}
        \includegraphics[width=\linewidth,trim={23cm 5.5cm 0 0},clip]{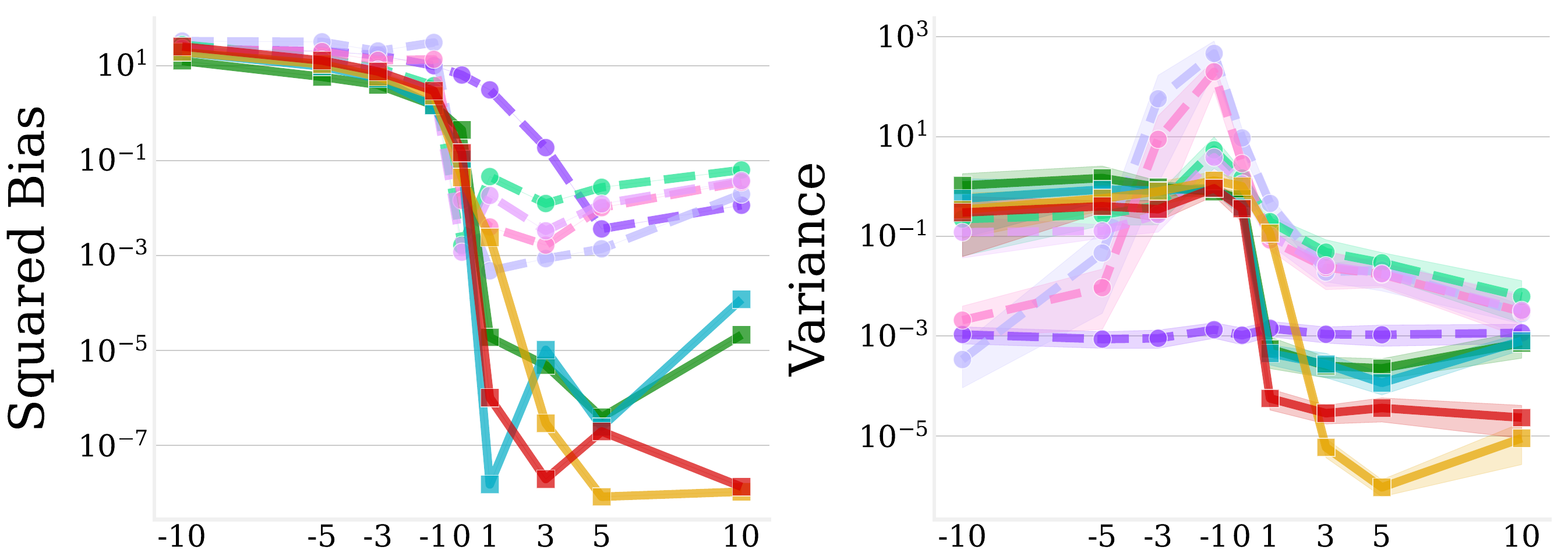}
    \end{subfigure}
    \begin{subfigure}{0.49\textwidth}
        \caption{Bias-Variance trade-off for \oracle-\dr}
        \includegraphics[width=\linewidth,trim={23cm 5.5cm 0 0},clip]{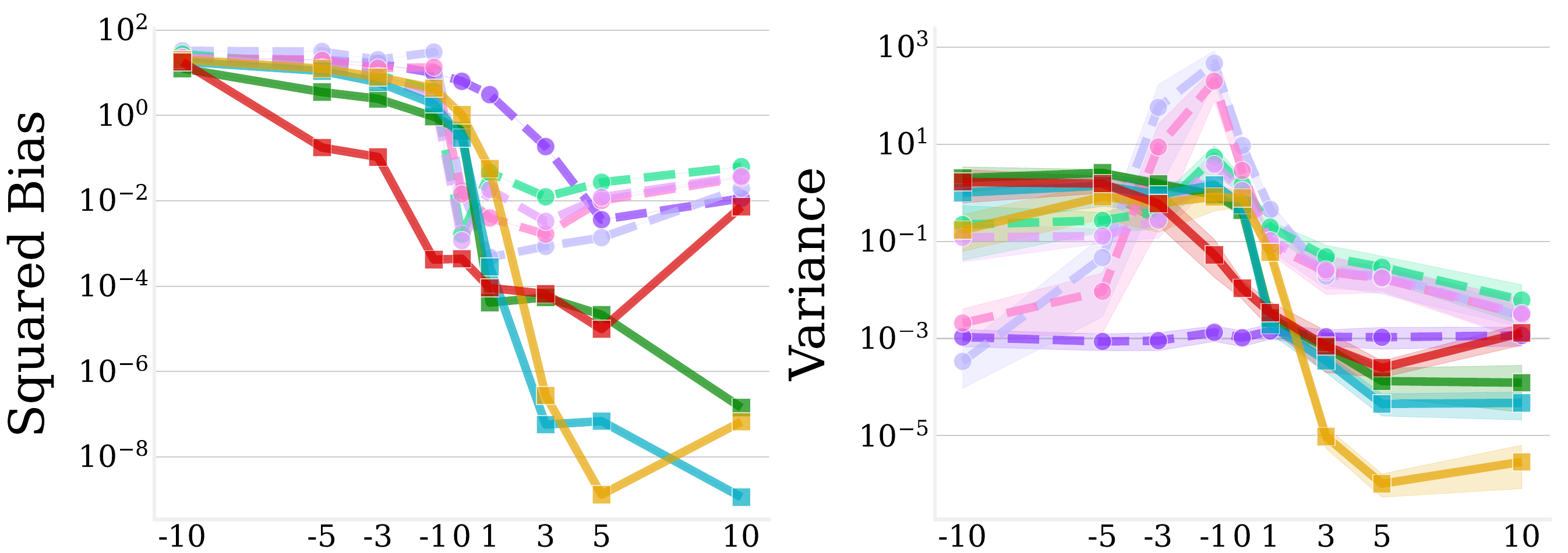}
    \end{subfigure} \hfill %
    \begin{subfigure}{0.49\textwidth}
        \caption{Bias-Variance trade-off for \oracle-\sndr}
        \includegraphics[width=\linewidth,trim={23cm 5.5cm 0 0},clip]{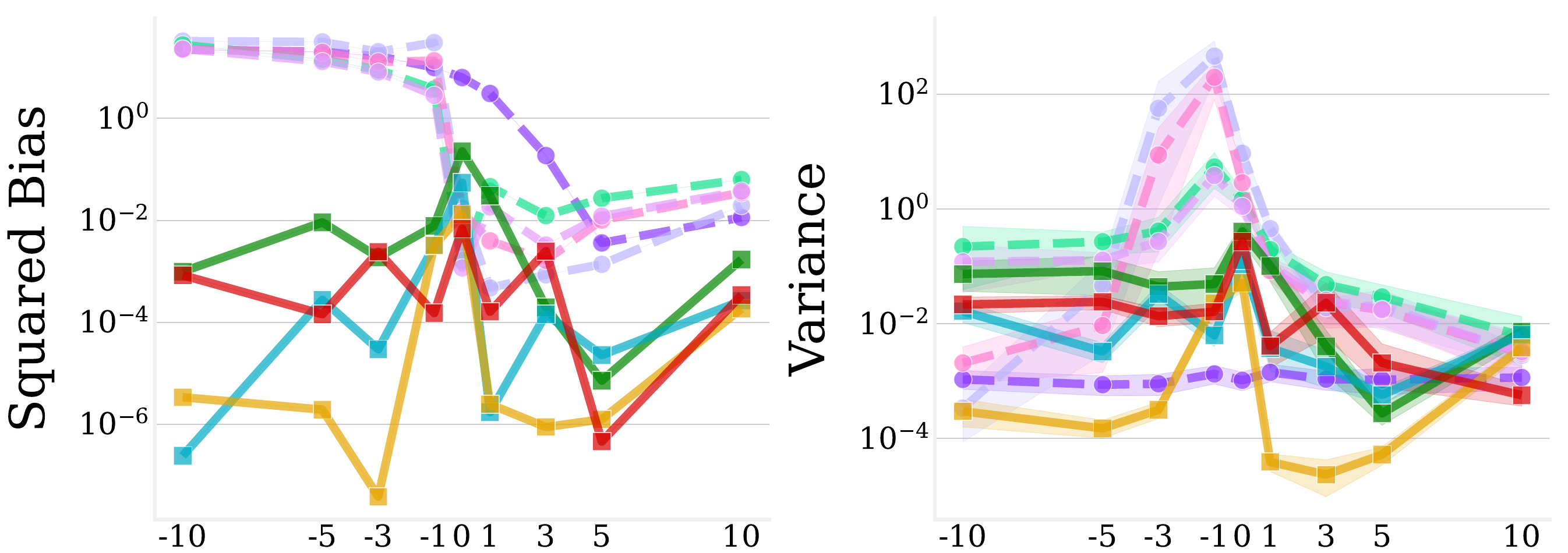}
    \end{subfigure}

    \vspace{0.3cm}
    
    \begin{subfigure}{\textwidth}
        \includegraphics[width=\linewidth,trim={0 0 0 0},clip]{figures/varying_beta/synthetic_eps_0.05.pdf}
    \end{subfigure} 
    \caption{Change in MSE, Squared Bias, and Variance while estimating $V(\pi_{\mathsf{good}})$ with varying policy-mismatch ($\log$ scale) for the synthetic dataset.}
    \label{fig:v_beta_synthetic_pi_good}
\end{minipage}
\end{figure*}

\begin{figure*}
\begin{minipage}[c][\textheight][c]{\textwidth}
    \centering
    \begin{subfigure}{0.49\textwidth}
        \caption{Bias-Variance trade-off for \oracle-\ips}
        \includegraphics[width=\linewidth,trim={23cm 5.5cm 0 0},clip]{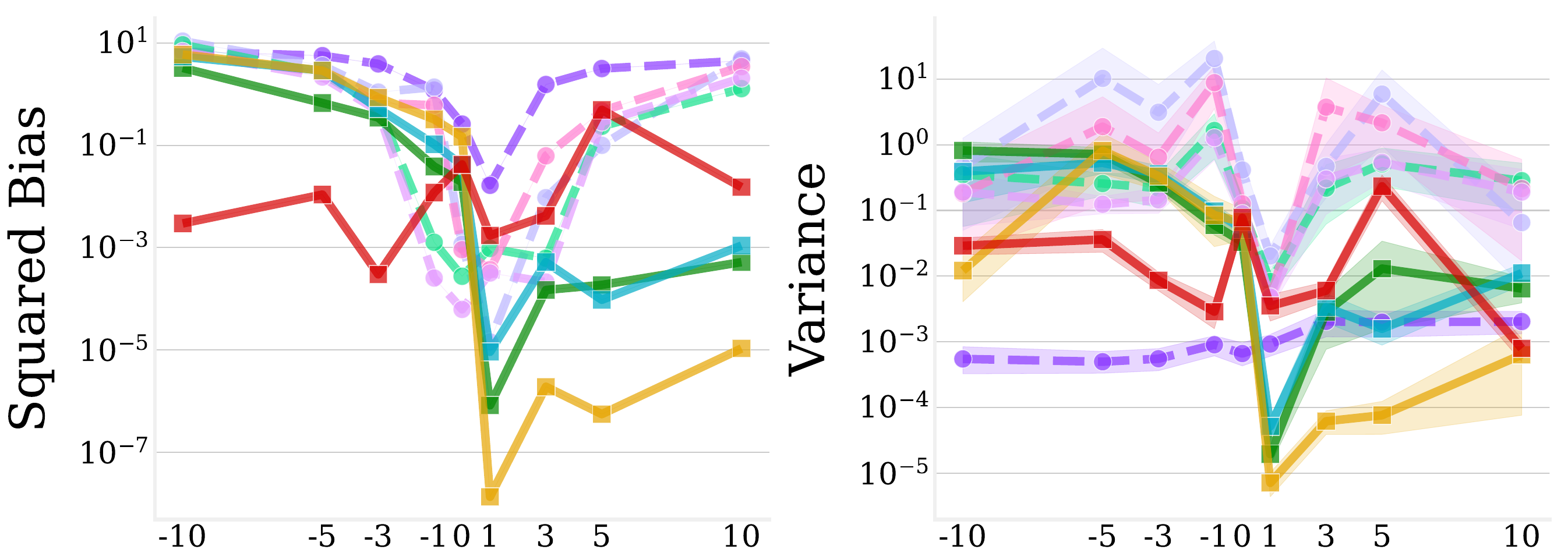}
    \end{subfigure} \hfill %
    \begin{subfigure}{0.49\textwidth}
        \caption{Bias-Variance trade-off for \oracle-\snips}
        \includegraphics[width=\linewidth,trim={23cm 5.5cm 0 0},clip]{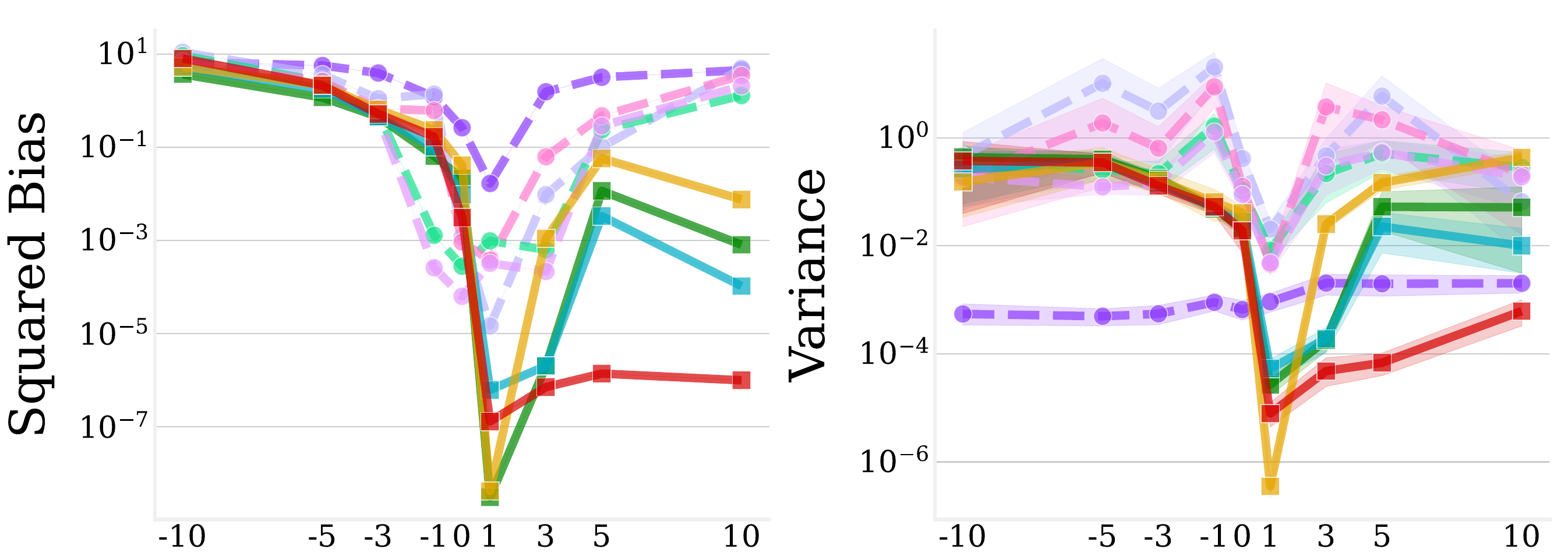}
    \end{subfigure}
    \begin{subfigure}{0.49\textwidth}
        \caption{Bias-Variance trade-off for \oracle-\dr}
        \includegraphics[width=\linewidth,trim={23cm 5.5cm 0 0},clip]{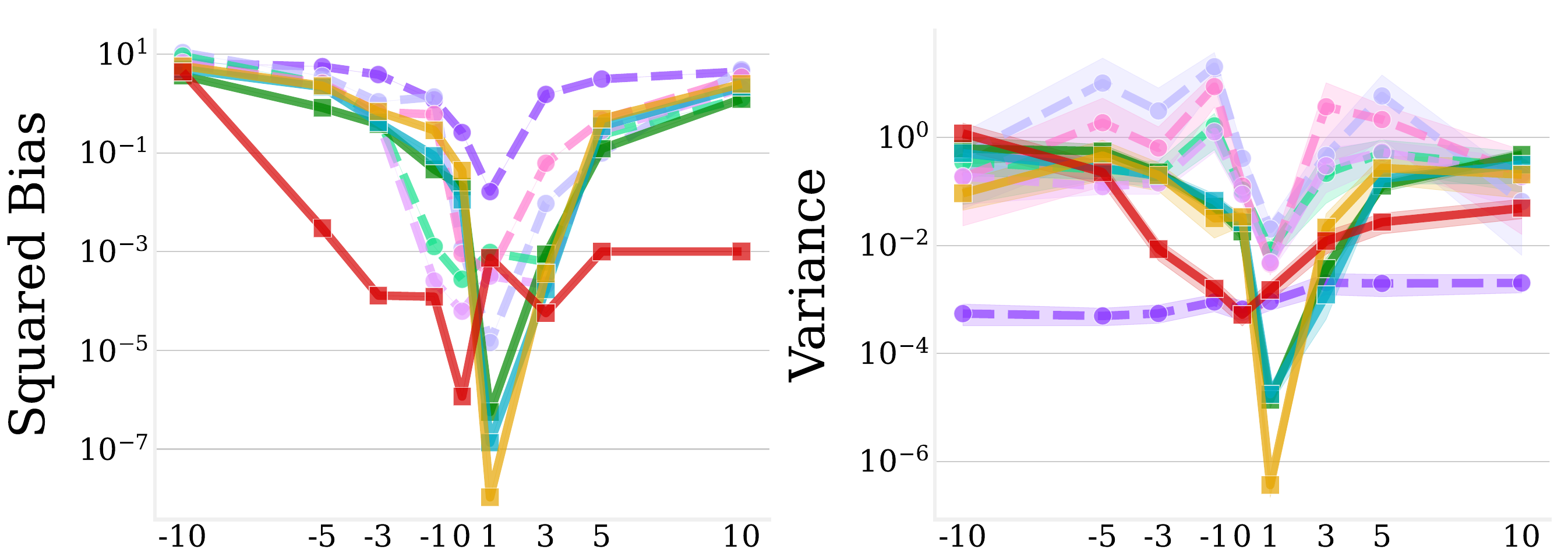}
    \end{subfigure} \hfill %
    \begin{subfigure}{0.49\textwidth}
        \caption{Bias-Variance trade-off for \oracle-\sndr}
        \includegraphics[width=\linewidth,trim={23cm 5.5cm 0 0},clip]{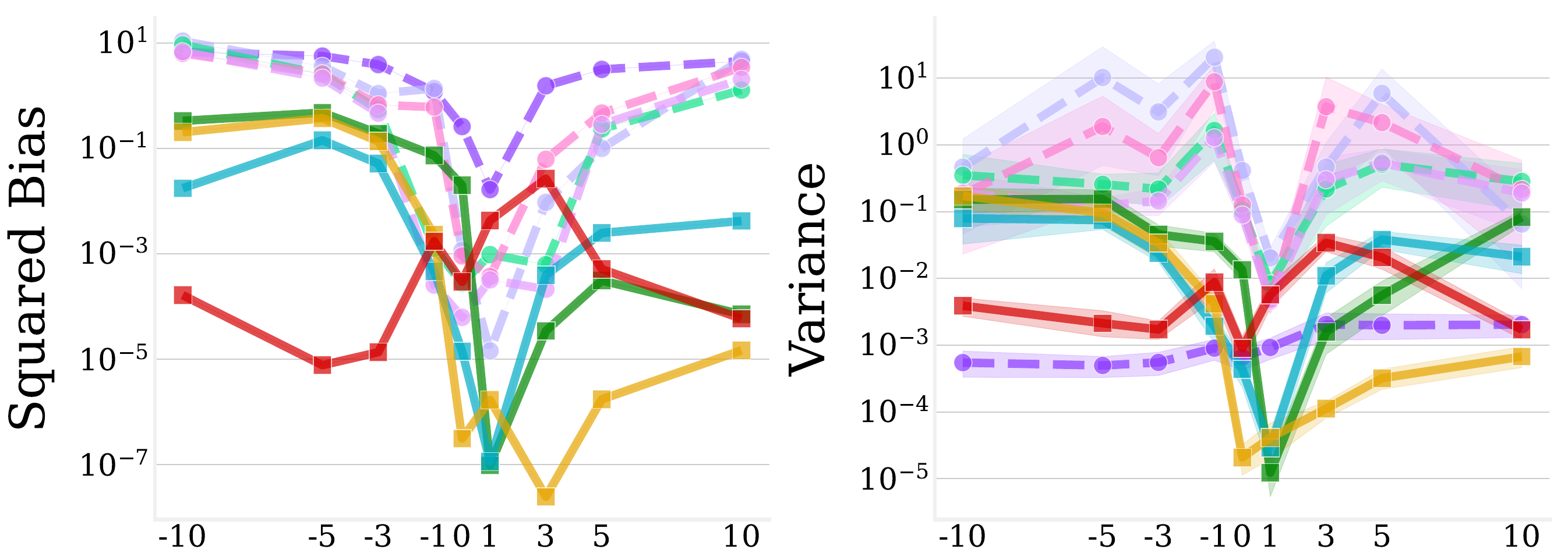}
    \end{subfigure}

    \vspace{0.3cm}
    
    \begin{subfigure}{\textwidth}
        \includegraphics[width=\linewidth,trim={0 0 0 0},clip]{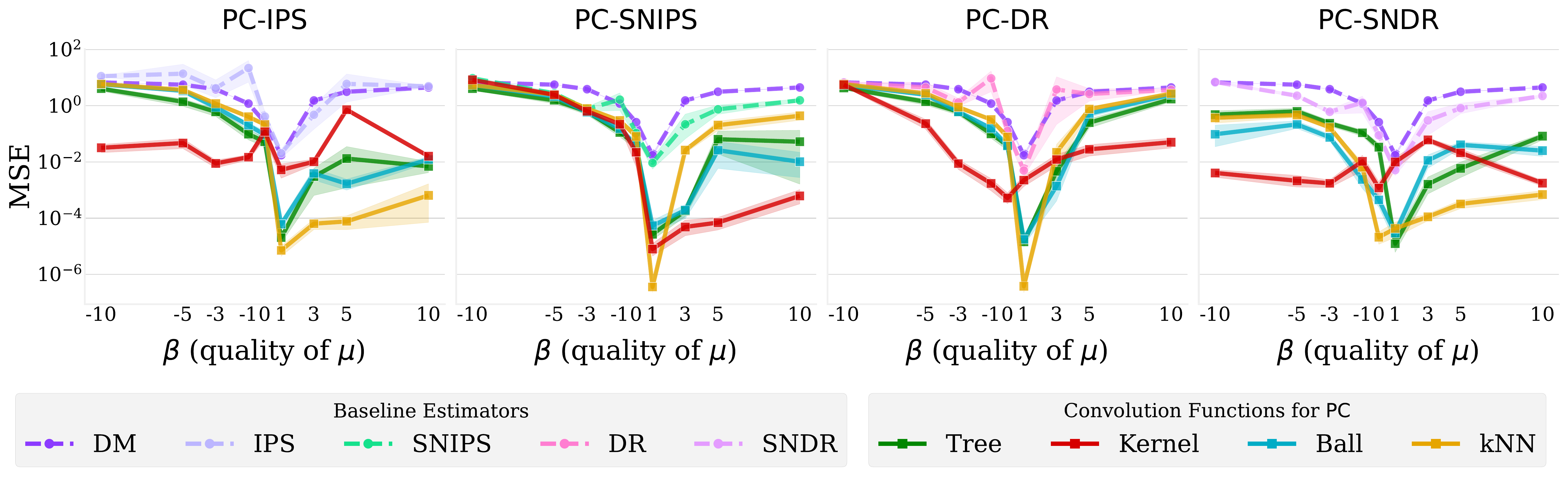}
    \end{subfigure} 
    \caption{Change in MSE, Squared Bias, and Variance while estimating $V(\pi_{\mathsf{bad}})$ with varying policy-mismatch ($\log$ scale) for the synthetic dataset.}
    \label{fig:v_beta_synthetic_pi_bad}
\end{minipage}
\end{figure*}

\begin{figure*}
\begin{minipage}[c][\textheight][c]{\textwidth}
    \centering
    \begin{subfigure}{0.49\textwidth}
        \caption{Bias-Variance trade-off for \oracle-\ips}
        \includegraphics[width=\linewidth,trim={23cm 5.5cm 0 0},clip]{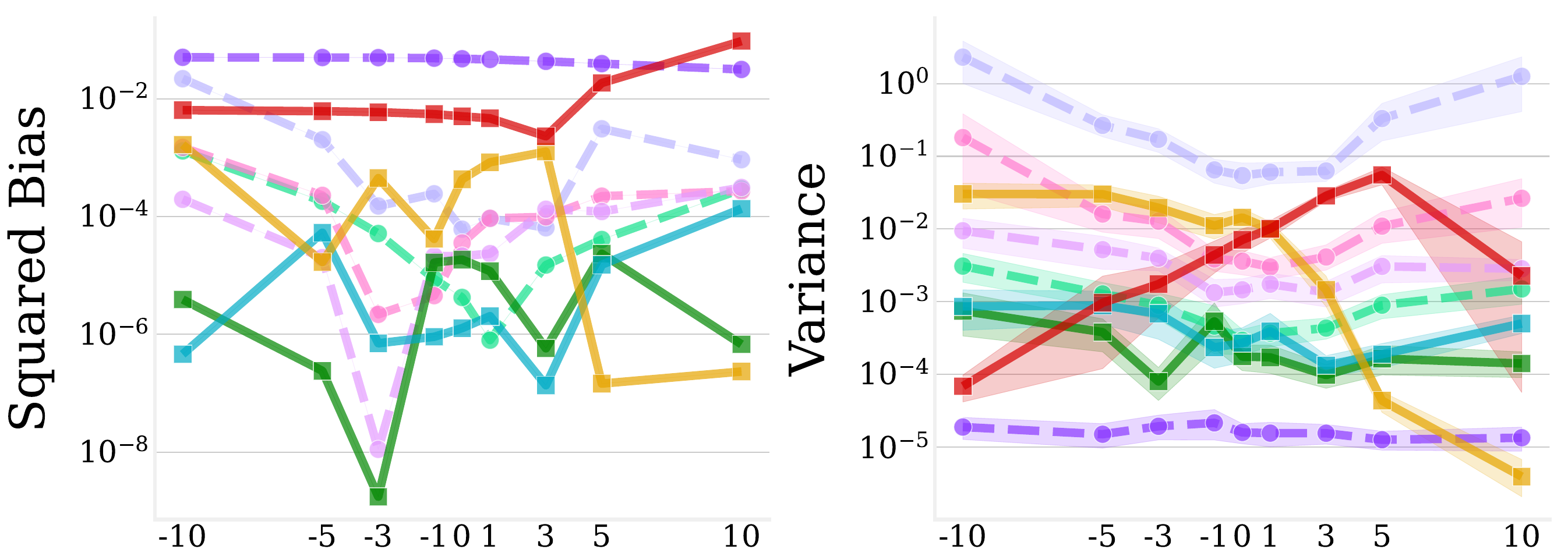}
    \end{subfigure} \hfill %
    \begin{subfigure}{0.49\textwidth}
        \caption{Bias-Variance trade-off for \oracle-\snips}
        \includegraphics[width=\linewidth,trim={23cm 5.5cm 0 0},clip]{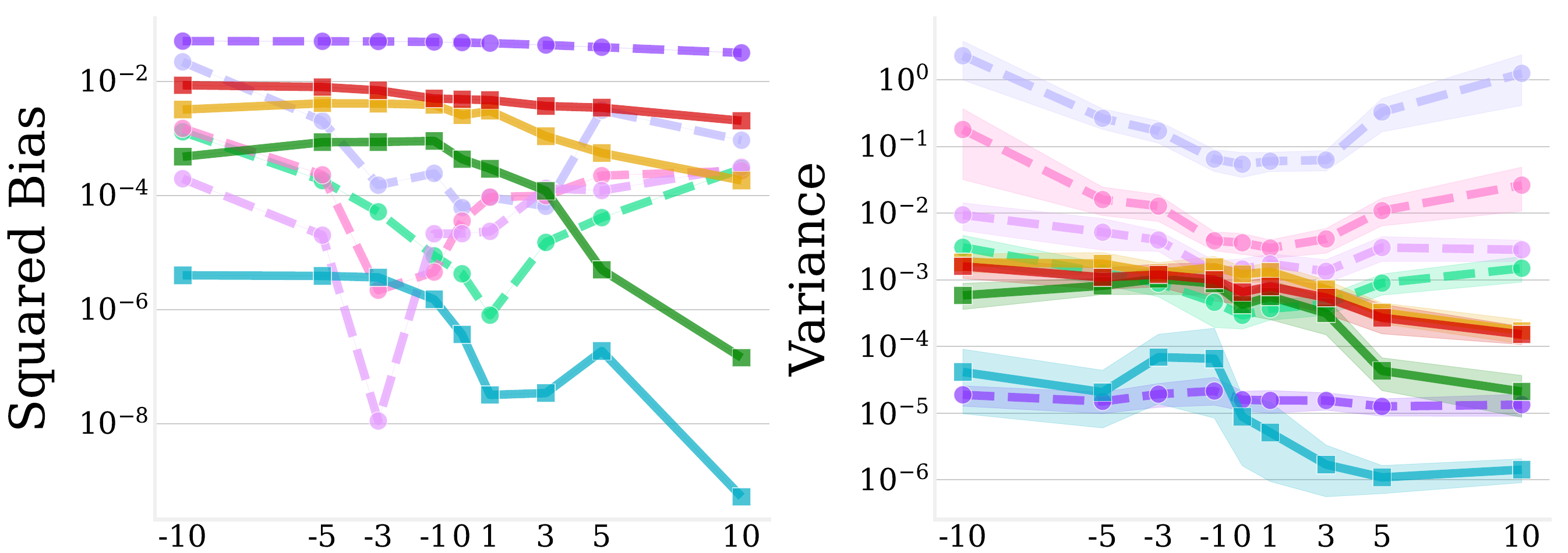}
    \end{subfigure}
    \begin{subfigure}{0.49\textwidth}
        \caption{Bias-Variance trade-off for \oracle-\dr}
        \includegraphics[width=\linewidth,trim={23cm 5.5cm 0 0},clip]{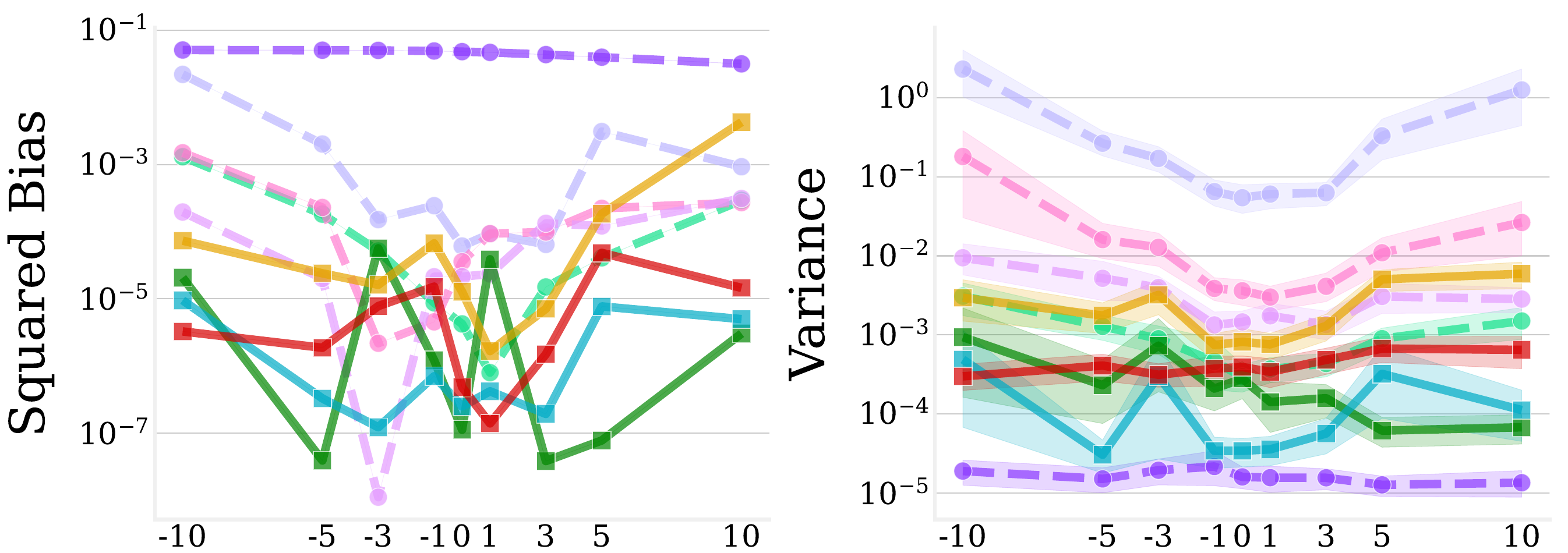}
    \end{subfigure} \hfill %
    \begin{subfigure}{0.49\textwidth}
        \caption{Bias-Variance trade-off for \oracle-\sndr}
        \includegraphics[width=\linewidth,trim={23cm 5.5cm 0 0},clip]{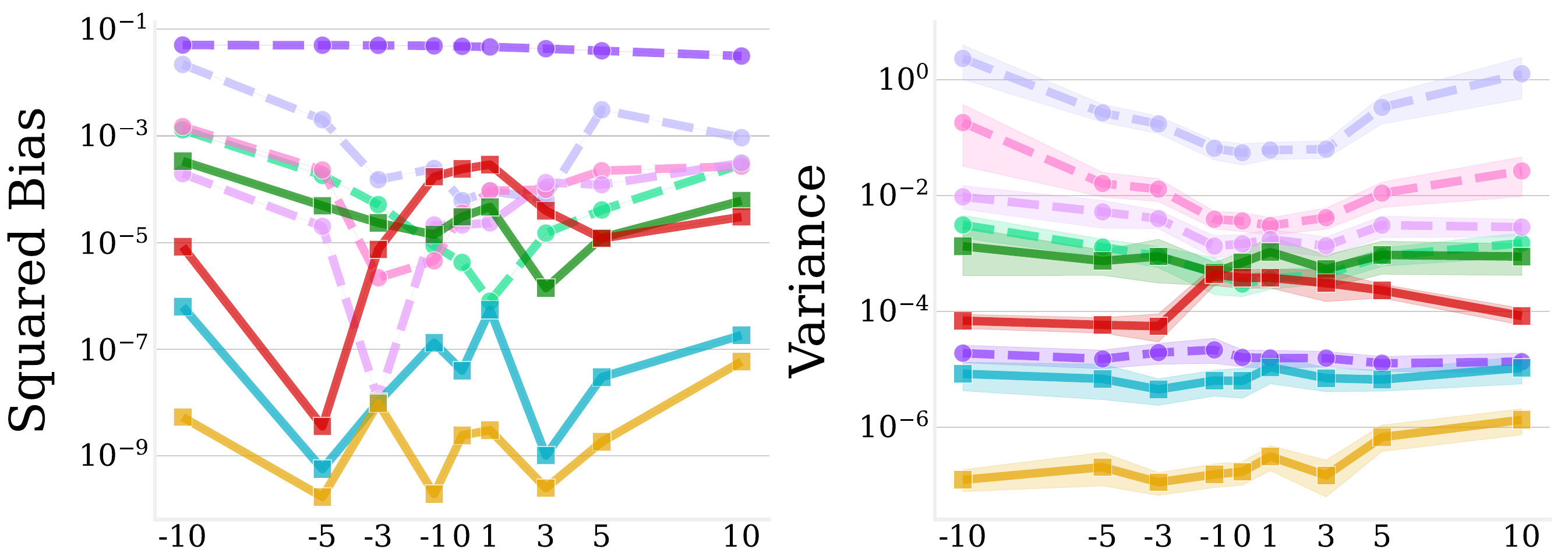}
    \end{subfigure}

    \vspace{0.3cm}
    
    \begin{subfigure}{\textwidth}
        \includegraphics[width=\linewidth,trim={0 0 0 0},clip]{figures/varying_beta/movielens_eps_0.05.pdf}
    \end{subfigure} 
    \caption{Change in MSE, Squared Bias, and Variance while estimating $V(\pi_{\mathsf{good}})$ with varying policy-mismatch ($\log$ scale) for the movielens dataset.}
    \label{fig:v_beta_ml_pi_good}
\end{minipage}
\end{figure*}

\begin{figure*}
\begin{minipage}[c][\textheight][c]{\textwidth}
    \centering
    \begin{subfigure}{0.49\textwidth}
        \caption{Bias-Variance trade-off for \oracle-\ips}
        \includegraphics[width=\linewidth,trim={23cm 5.5cm 0 0},clip]{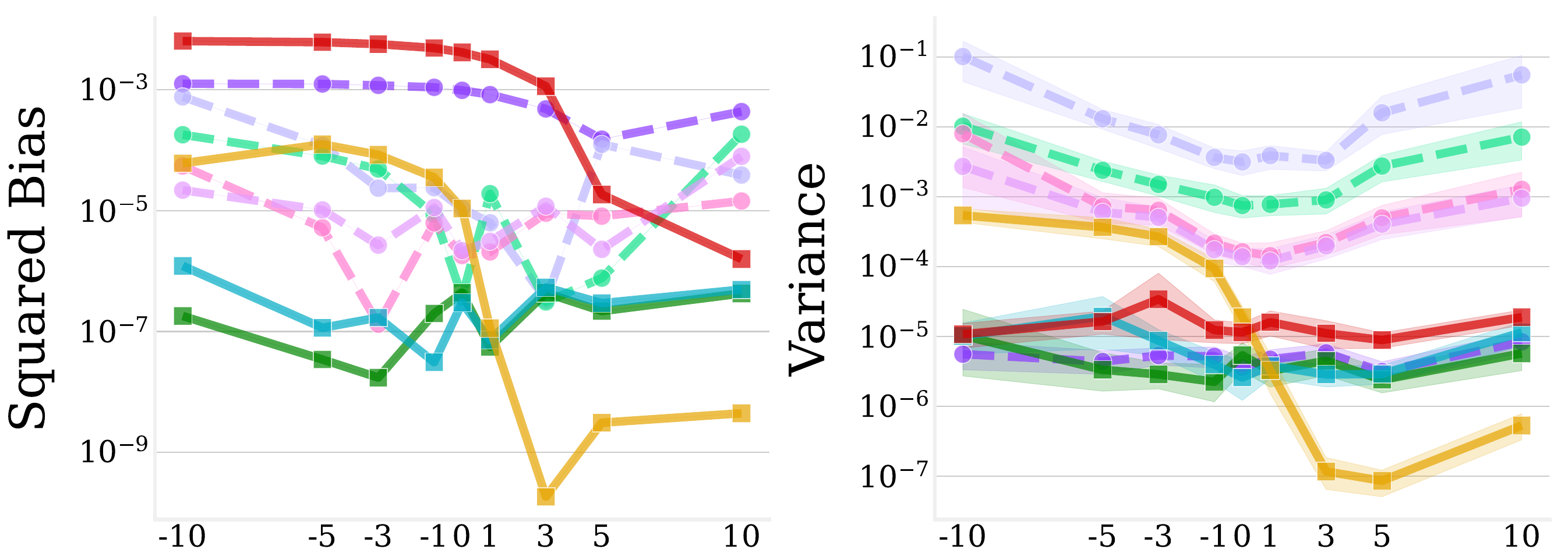}
    \end{subfigure} \hfill %
    \begin{subfigure}{0.49\textwidth}
        \caption{Bias-Variance trade-off for \oracle-\snips}
        \includegraphics[width=\linewidth,trim={23cm 5.5cm 0 0},clip]{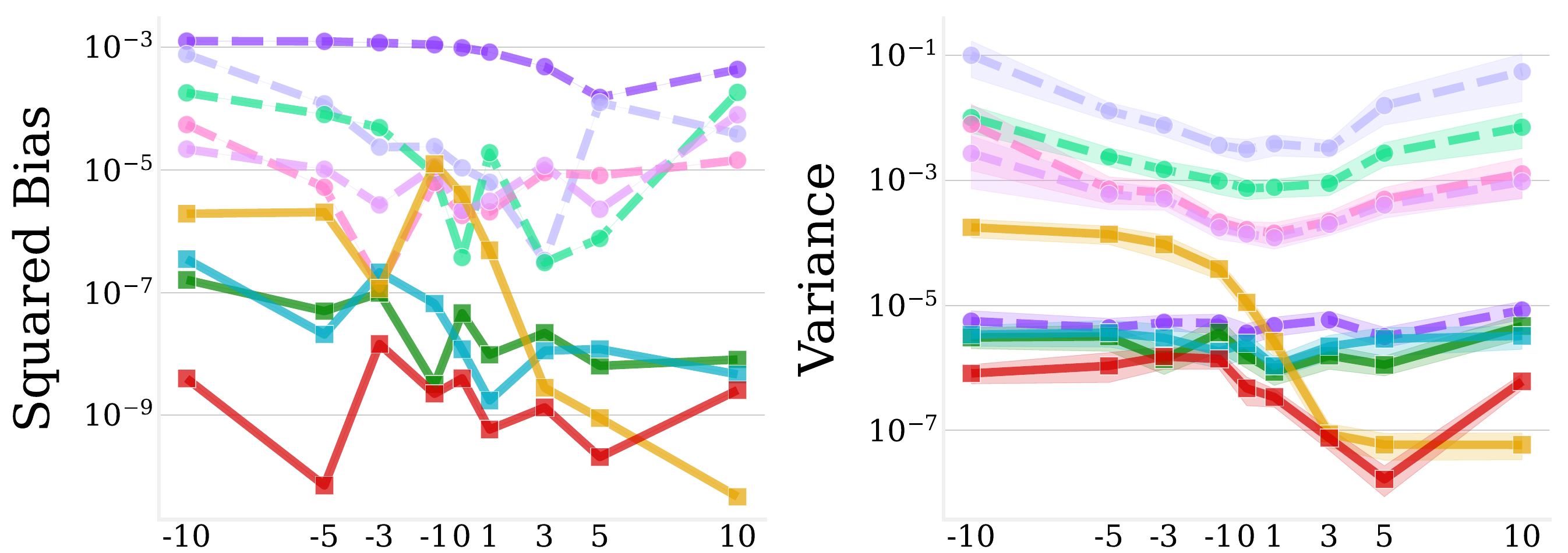}
    \end{subfigure}
    \begin{subfigure}{0.49\textwidth}
        \caption{Bias-Variance trade-off for \oracle-\dr}
        \includegraphics[width=\linewidth,trim={23cm 5.5cm 0 0},clip]{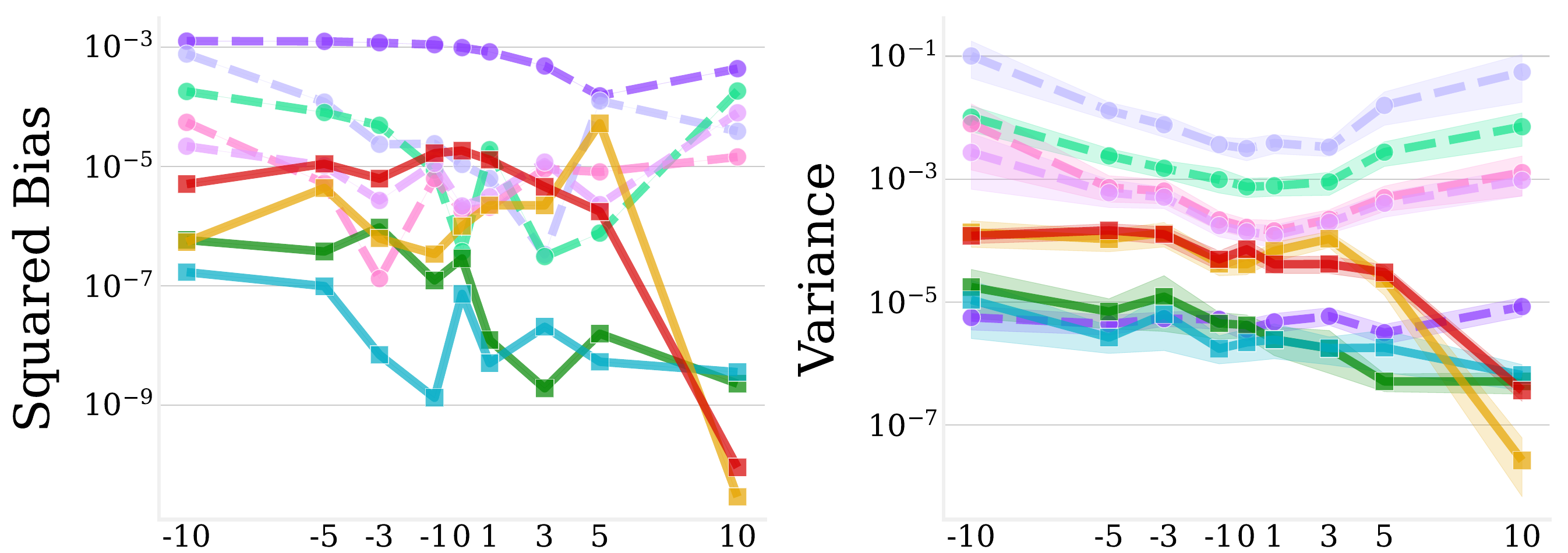}
    \end{subfigure} \hfill %
    \begin{subfigure}{0.49\textwidth}
        \caption{Bias-Variance trade-off for \oracle-\sndr}
        \includegraphics[width=\linewidth,trim={23cm 5.5cm 0 0},clip]{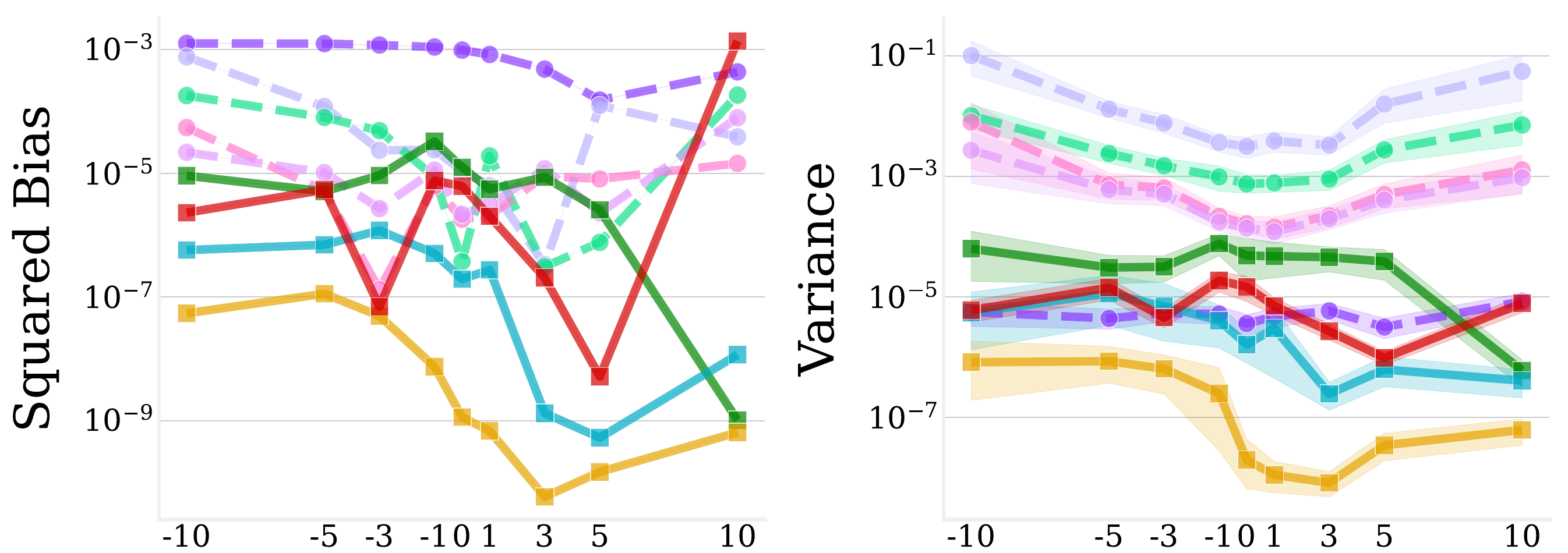}
    \end{subfigure}

    \vspace{0.3cm}
    
    \begin{subfigure}{\textwidth}
        \includegraphics[width=\linewidth,trim={0 0 0 0},clip]{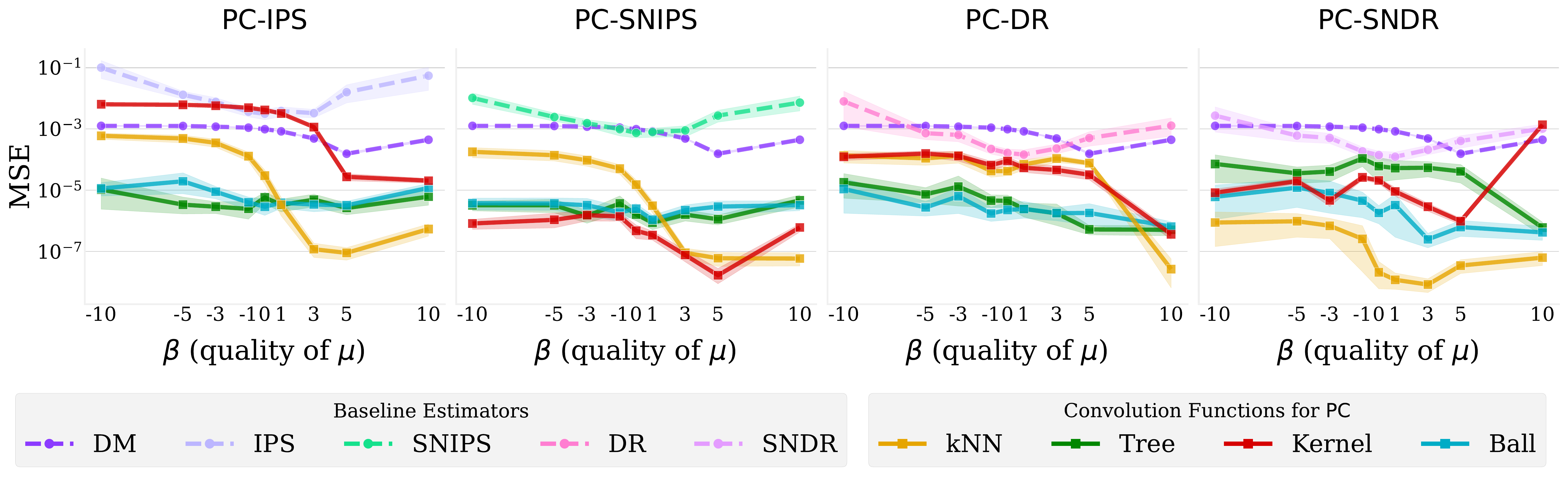}
    \end{subfigure} 
    \caption{Change in MSE, Squared Bias, and Variance while estimating $V(\pi_{\mathsf{bad}})$ with varying policy-mismatch ($\log$ scale) for the movielens dataset.}
    \label{fig:v_beta_ml_pi_bad}
\end{minipage}
\end{figure*}

\clearpage

\begin{figure*}
\begin{minipage}[c][\textheight][c]{\textwidth}
    \centering
    \begin{subfigure}{0.49\textwidth}
        \caption{Bias-Variance trade-off for \oracle-\ips}
        \includegraphics[width=\linewidth,trim={23cm 5.5cm 0 0},clip]{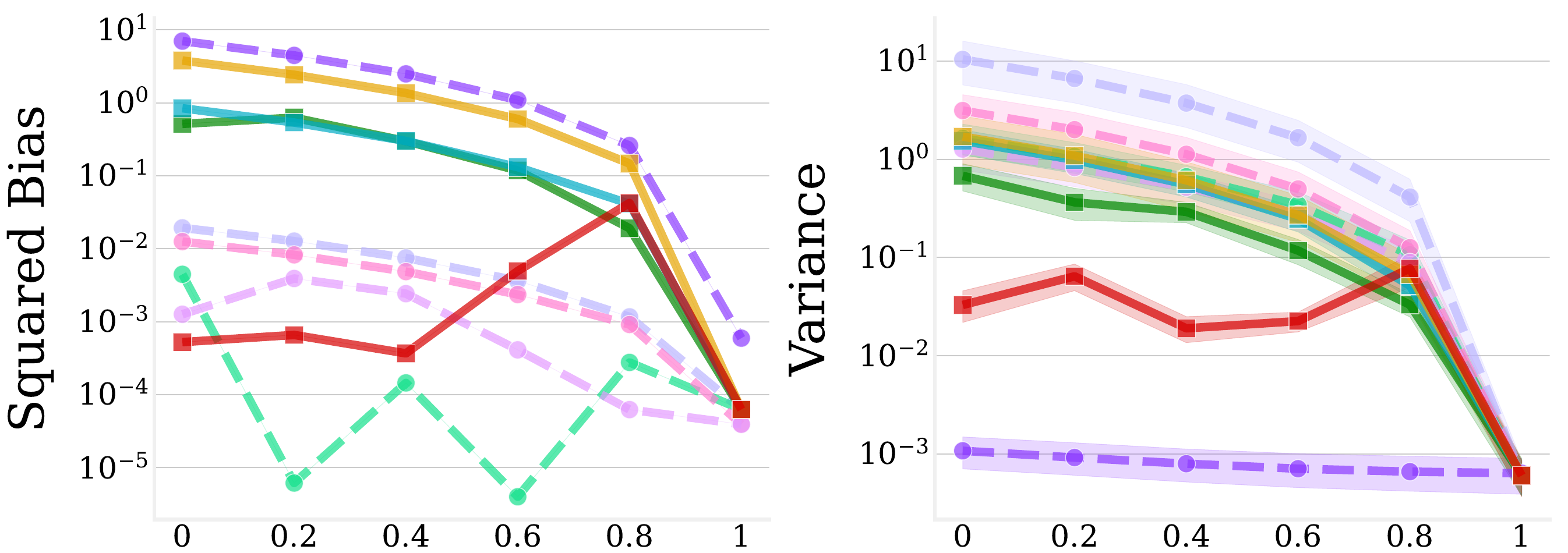}
    \end{subfigure} \hfill %
    \begin{subfigure}{0.49\textwidth}
        \caption{Bias-Variance trade-off for \oracle-\snips}
        \includegraphics[width=\linewidth,trim={23cm 5.5cm 0 0},clip]{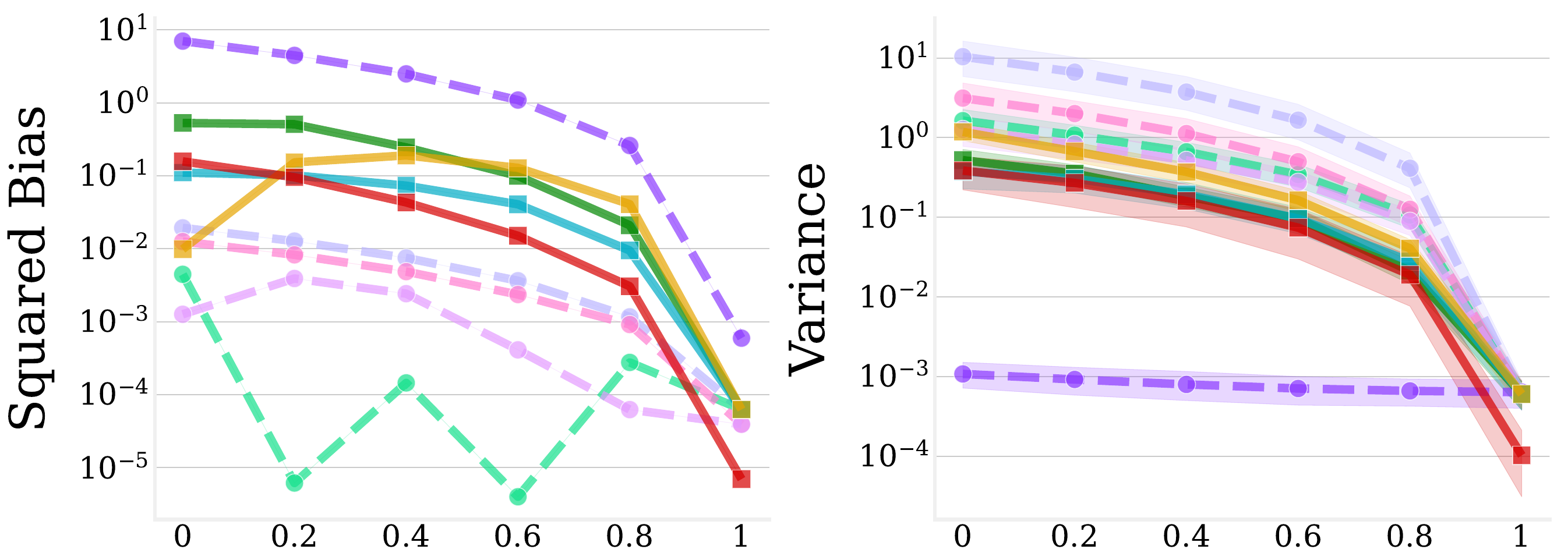}
    \end{subfigure}
    \begin{subfigure}{0.49\textwidth}
        \caption{Bias-Variance trade-off for \oracle-\dr}
        \includegraphics[width=\linewidth,trim={23cm 5.5cm 0 0},clip]{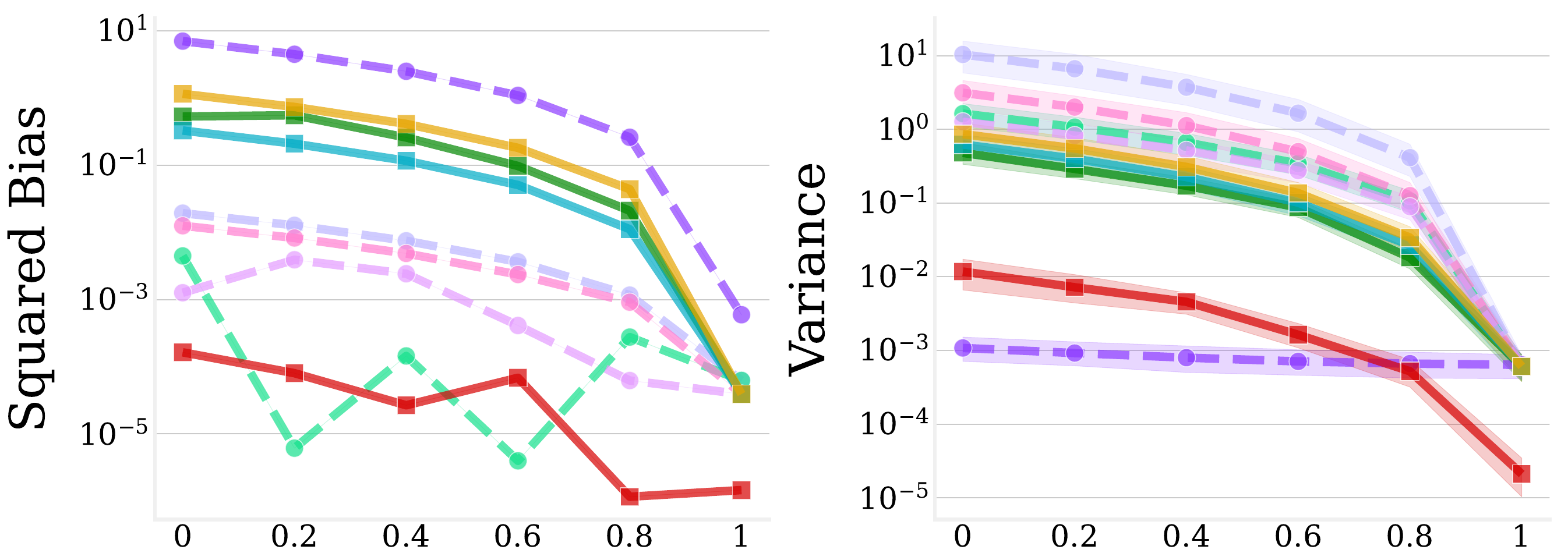}
    \end{subfigure} \hfill %
    \begin{subfigure}{0.49\textwidth}
        \caption{Bias-Variance trade-off for \oracle-\sndr}
        \includegraphics[width=\linewidth,trim={23cm 5.5cm 0 0},clip]{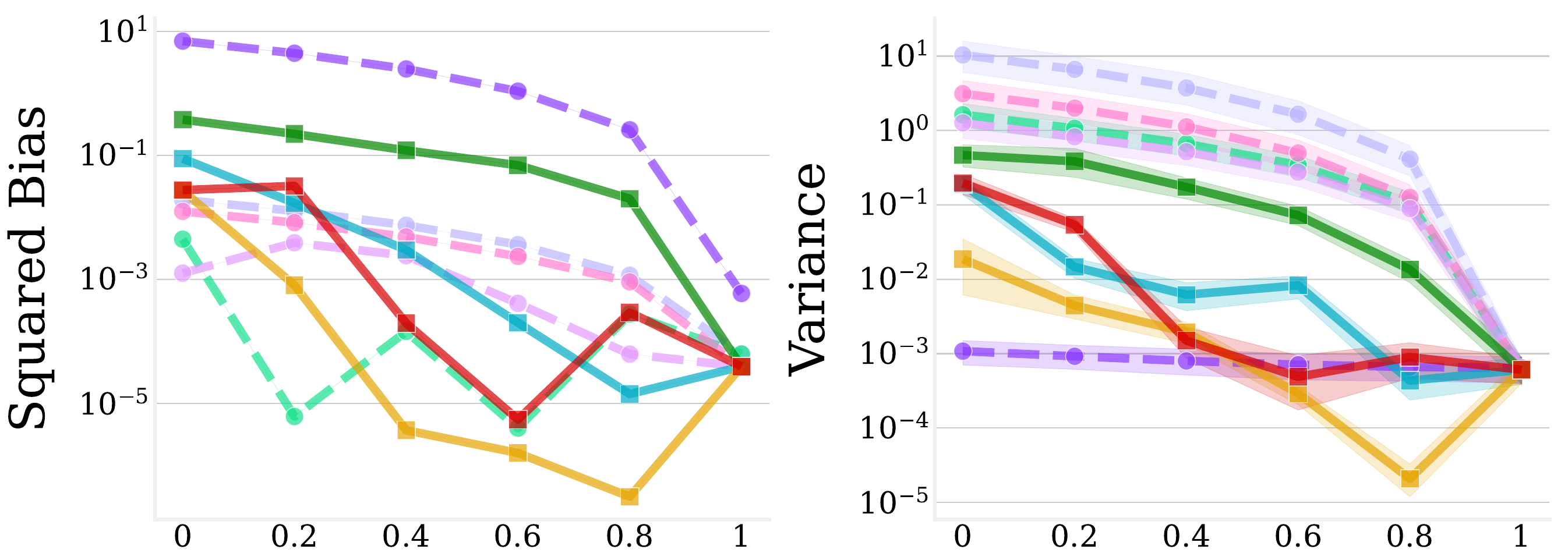}
    \end{subfigure}

    \vspace{0.3cm}
    
    \begin{subfigure}{\textwidth}
        \includegraphics[width=\linewidth,trim={0 0 0 0},clip]{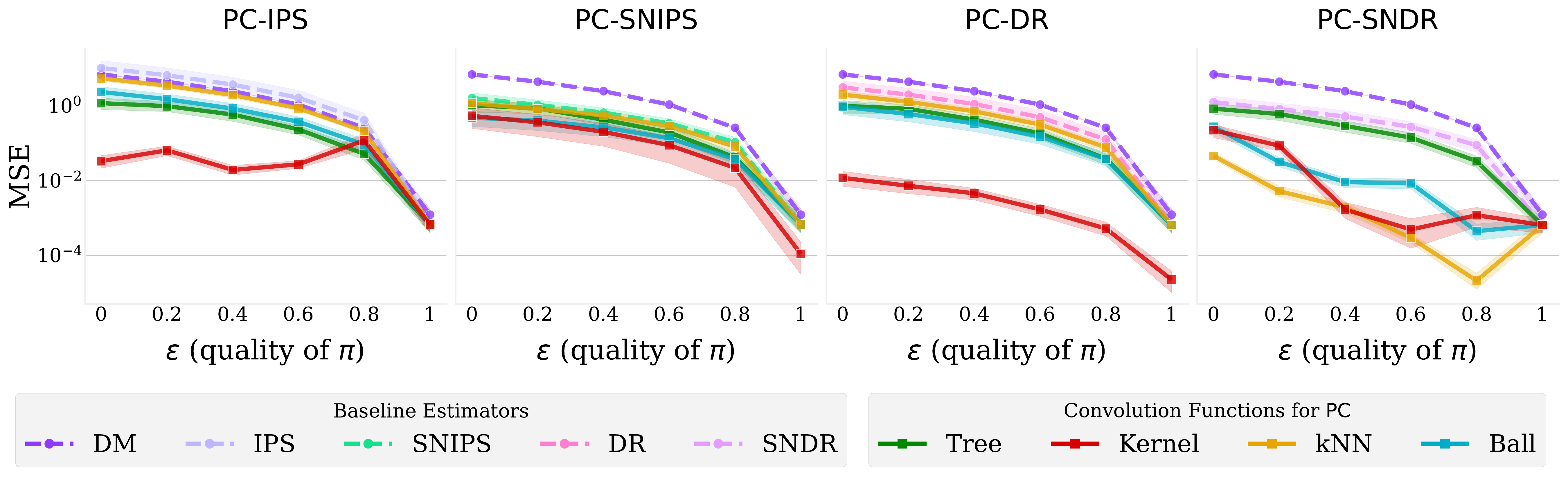}
    \end{subfigure} 
    \caption{Change in MSE, Squared Bias, and Variance while estimating various target policies ($\log$ scale) for the synthetic dataset, using data logged by $\mu_{\mathsf{uniform}}$.}
    \label{fig:v_eps_synthetic_mu_unif}
\end{minipage}
\end{figure*}

\begin{figure*}
\begin{minipage}[c][\textheight][c]{\textwidth}
    \centering
    \begin{subfigure}{0.49\textwidth}
        \caption{Bias-Variance trade-off for \oracle-\ips}
        \includegraphics[width=\linewidth,trim={23cm 5.5cm 0 0},clip]{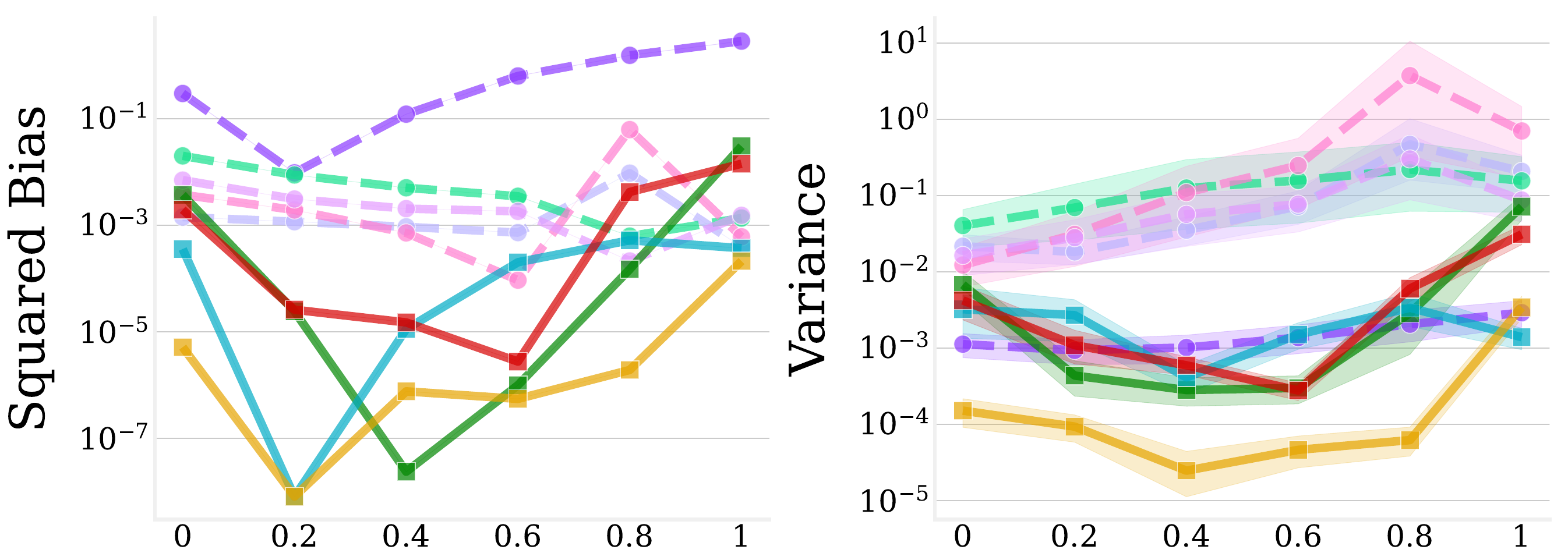}
    \end{subfigure} \hfill %
    \begin{subfigure}{0.49\textwidth}
        \caption{Bias-Variance trade-off for \oracle-\snips}
        \includegraphics[width=\linewidth,trim={23cm 5.5cm 0 0},clip]{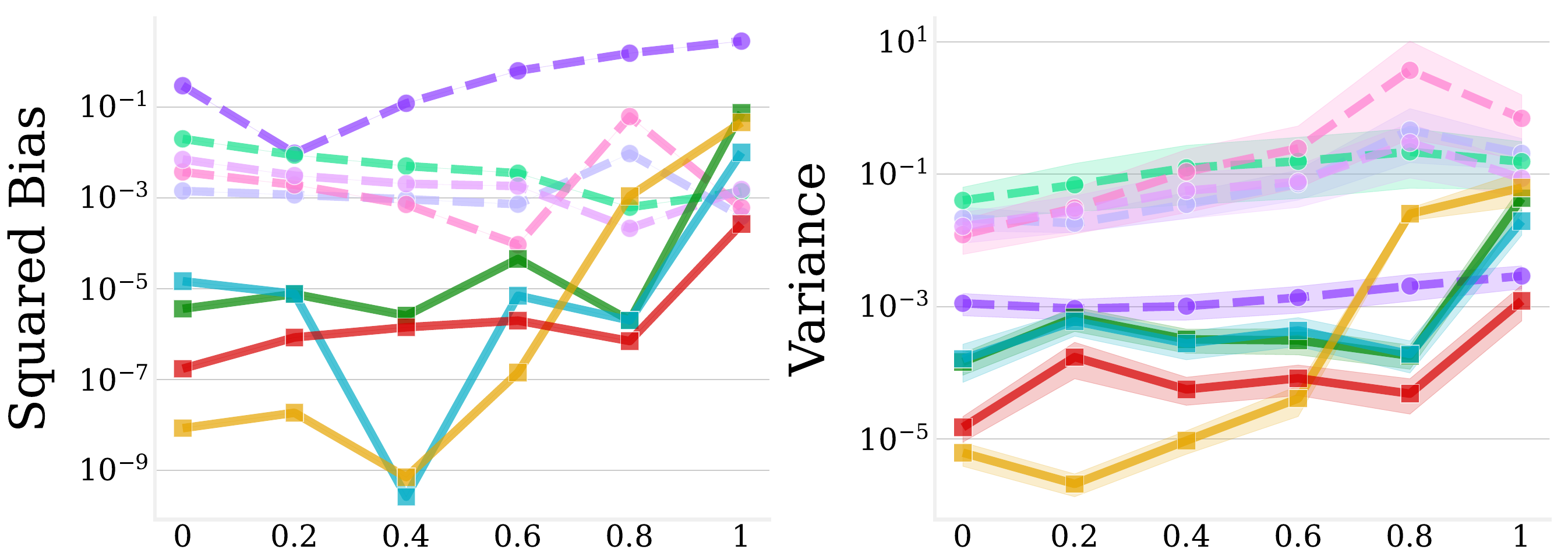}
    \end{subfigure}
    \begin{subfigure}{0.49\textwidth}
        \caption{Bias-Variance trade-off for \oracle-\dr}
        \includegraphics[width=\linewidth,trim={23cm 5.5cm 0 0},clip]{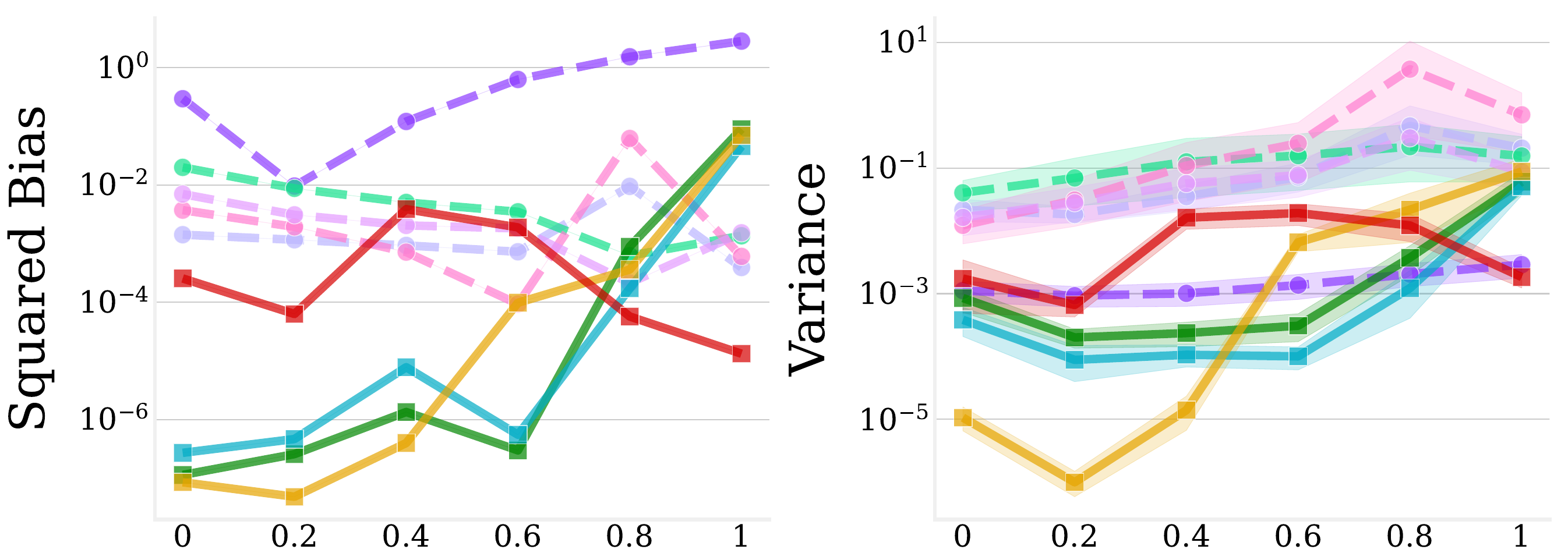}
    \end{subfigure} \hfill %
    \begin{subfigure}{0.49\textwidth}
        \caption{Bias-Variance trade-off for \oracle-\sndr}
        \includegraphics[width=\linewidth,trim={23cm 5.5cm 0 0},clip]{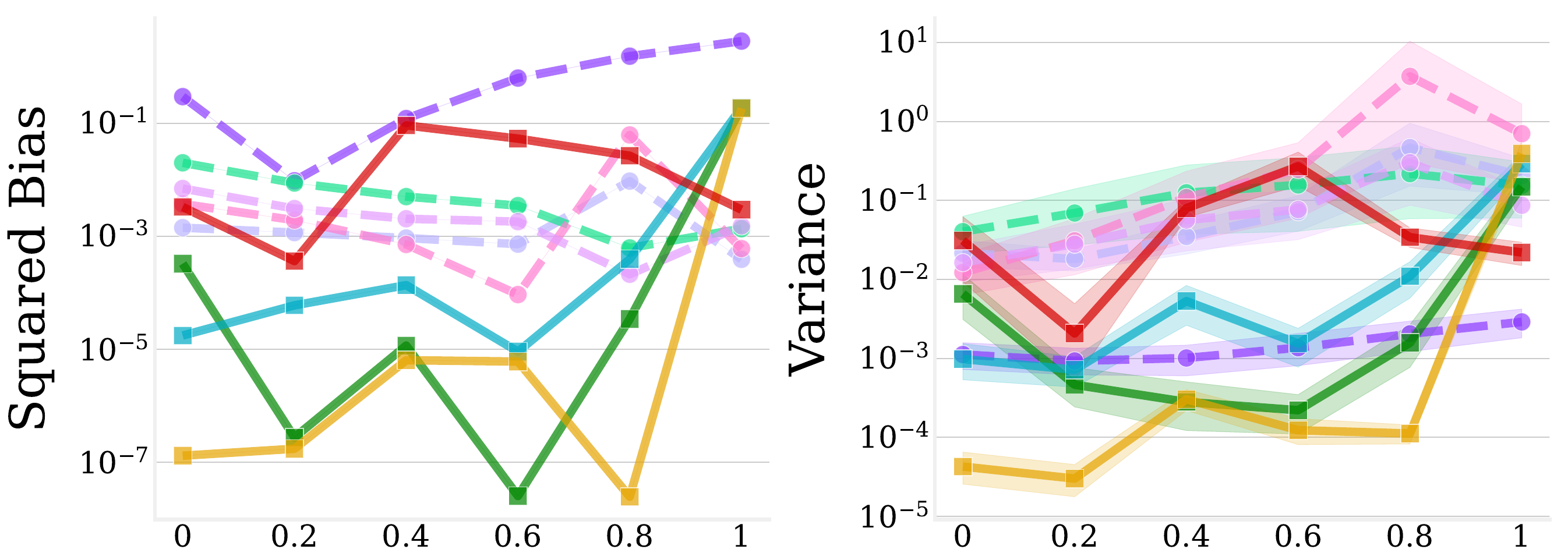}
    \end{subfigure}

    \vspace{0.3cm}
    
    \begin{subfigure}{\textwidth}
        \includegraphics[width=\linewidth,trim={0 0 0 0},clip]{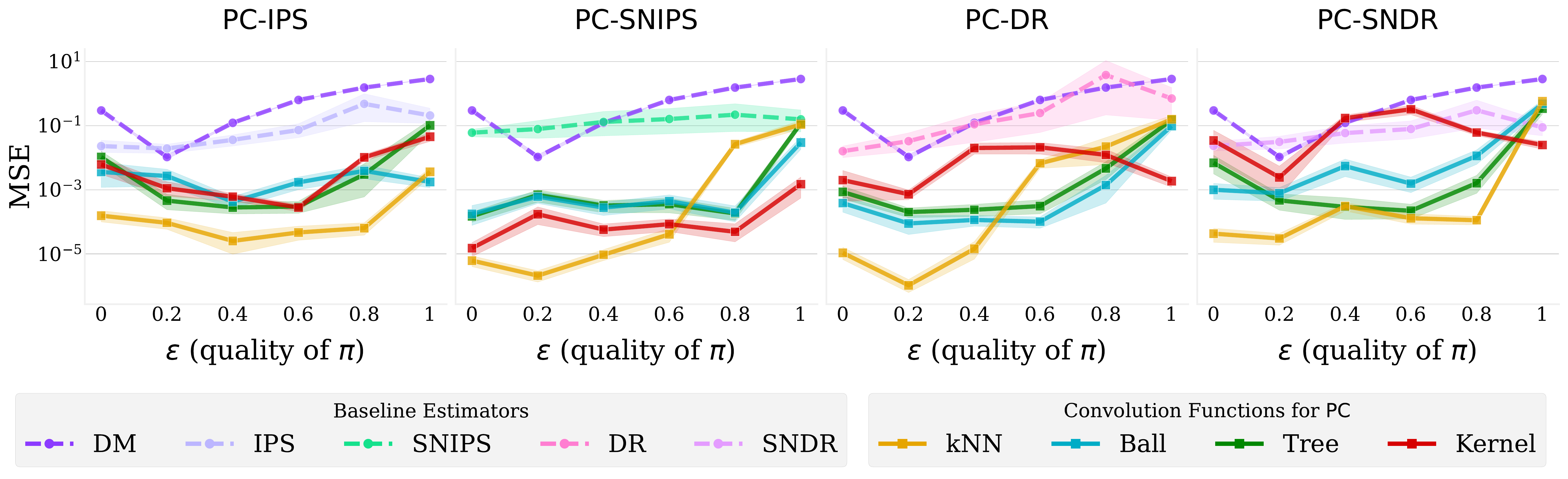}
    \end{subfigure} 
    \caption{Change in MSE, Squared Bias, and Variance while estimating various target policies ($\log$ scale) for the synthetic dataset, using data logged by $\mu_{\mathsf{good}}$.}
    \label{fig:v_eps_synthetic_mu_good}
\end{minipage}
\end{figure*}

\begin{figure*}
\begin{minipage}[c][\textheight][c]{\textwidth}
    \centering
    \begin{subfigure}{0.49\textwidth}
        \caption{Bias-Variance trade-off for \oracle-\ips}
        \includegraphics[width=\linewidth,trim={23cm 5.5cm 0 0},clip]{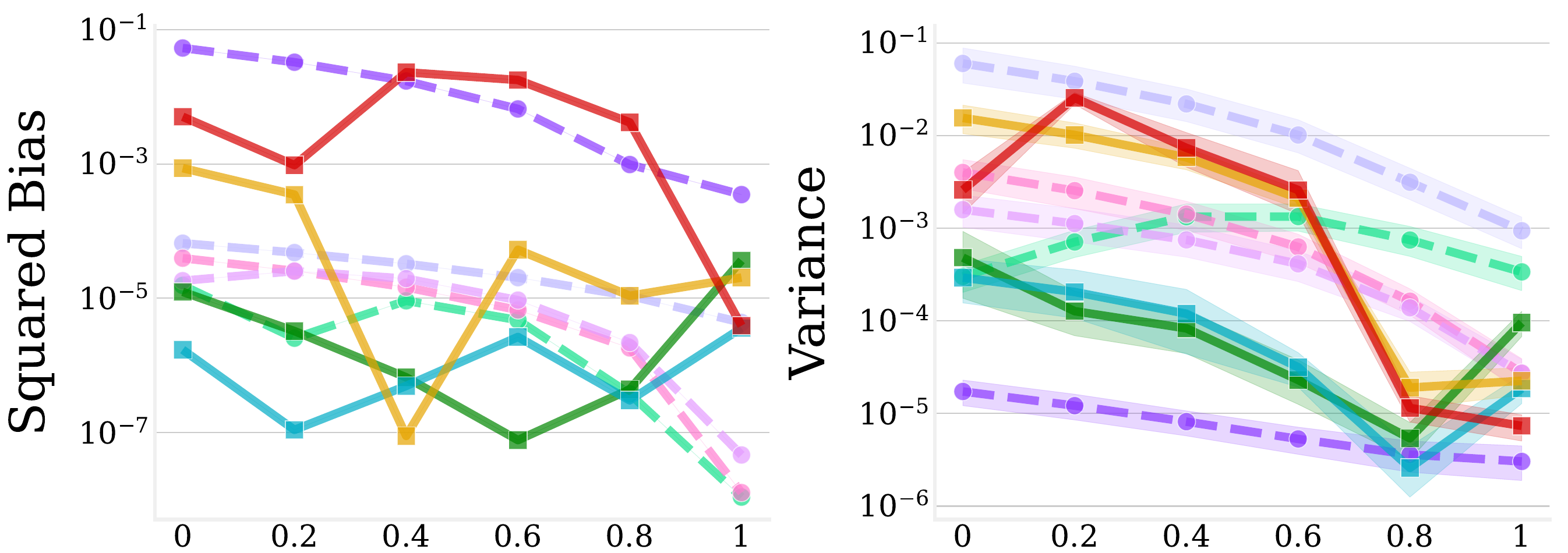}
    \end{subfigure} \hfill %
    \begin{subfigure}{0.49\textwidth}
        \caption{Bias-Variance trade-off for \oracle-\snips}
        \includegraphics[width=\linewidth,trim={23cm 5.5cm 0 0},clip]{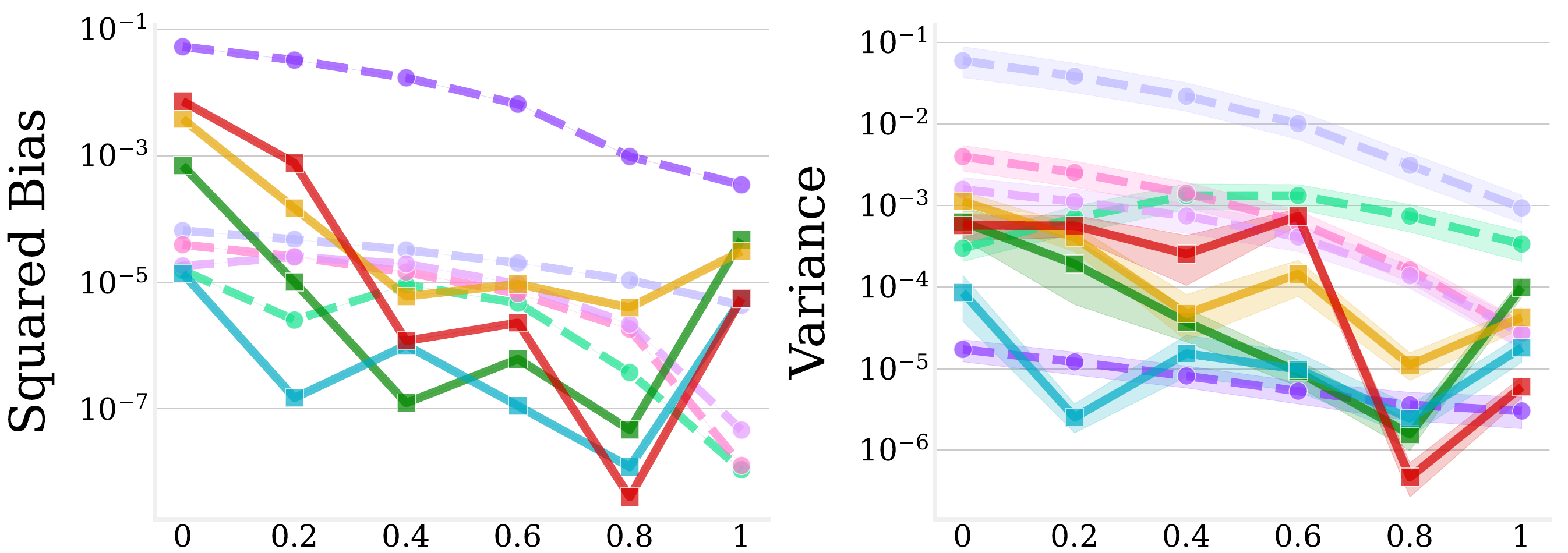}
    \end{subfigure}
    \begin{subfigure}{0.49\textwidth}
        \caption{Bias-Variance trade-off for \oracle-\dr}
        \includegraphics[width=\linewidth,trim={23cm 5.5cm 0 0},clip]{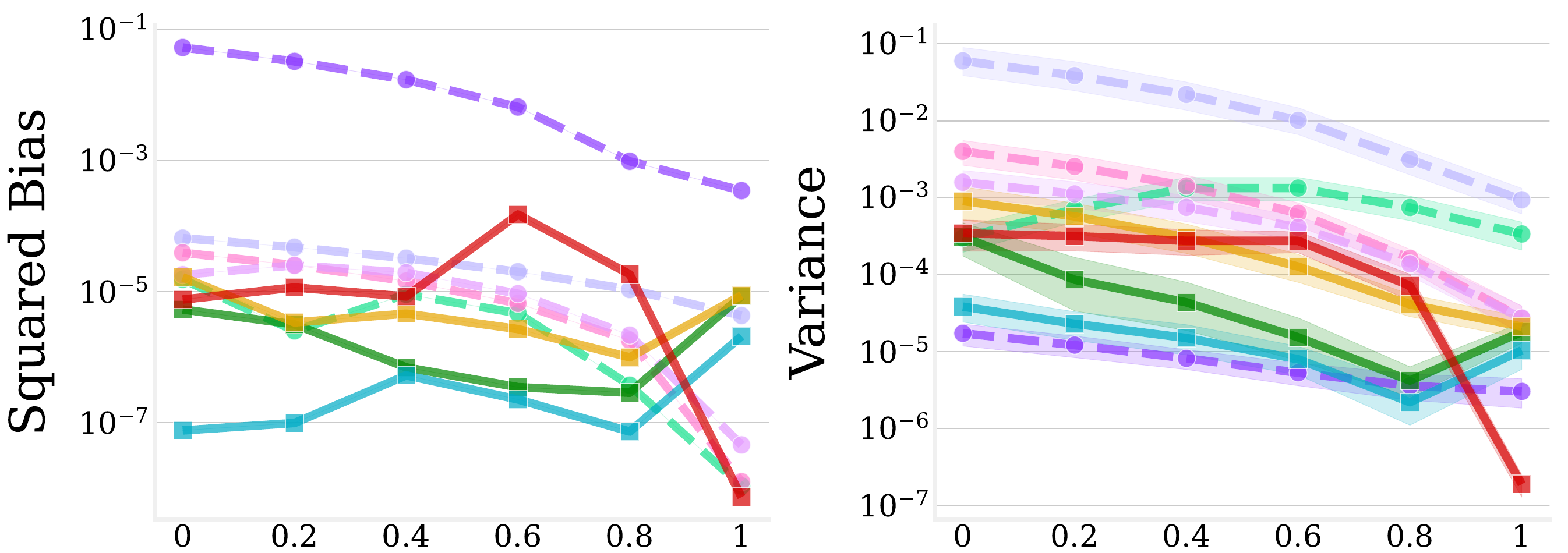}
    \end{subfigure} \hfill %
    \begin{subfigure}{0.49\textwidth}
        \caption{Bias-Variance trade-off for \oracle-\sndr}
        \includegraphics[width=\linewidth,trim={23cm 5.5cm 0 0},clip]{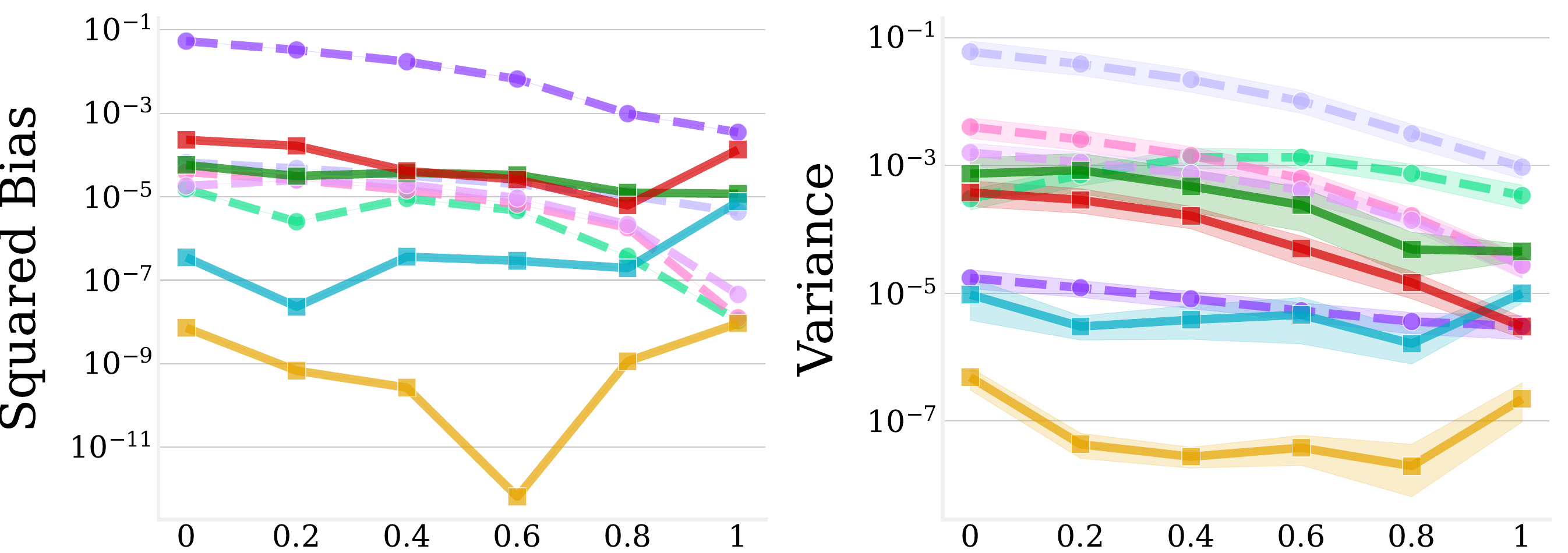}
    \end{subfigure}

    \vspace{0.3cm}
    
    \begin{subfigure}{\textwidth}
        \includegraphics[width=\linewidth,trim={0 0 0 0},clip]{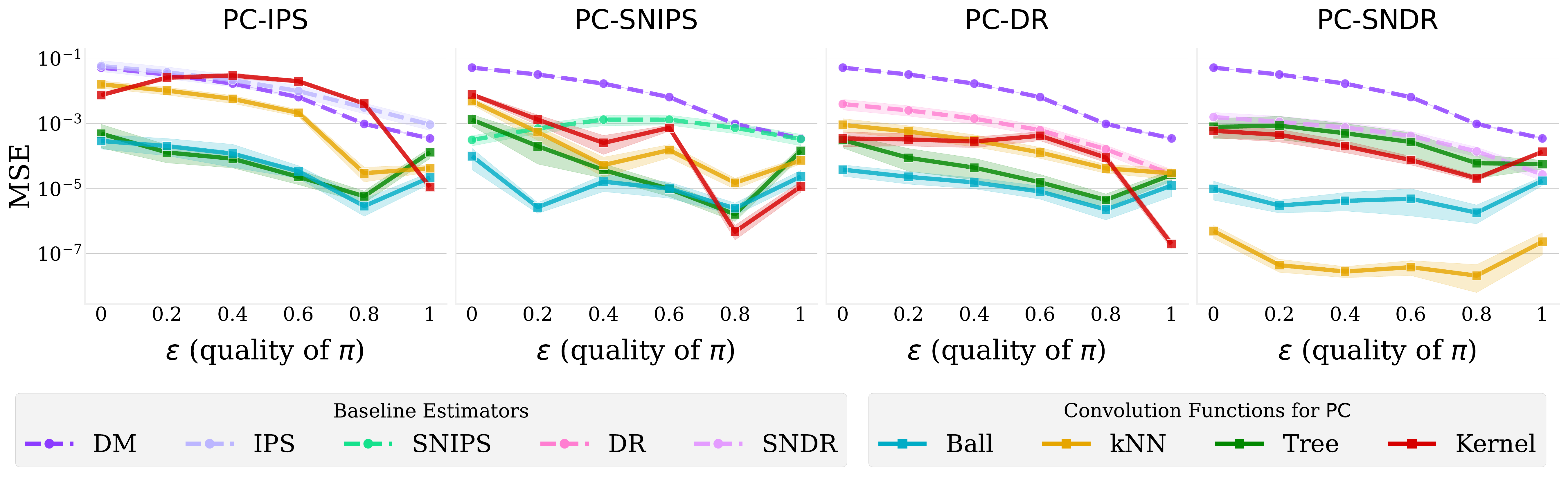}
    \end{subfigure} 
    \caption{Change in MSE, Squared Bias, and Variance while estimating various target policies ($\log$ scale) for the movielens dataset, using data logged by $\mu_{\mathsf{uniform}}$.}
    \label{fig:v_eps_ml_mu_unif}
\end{minipage}
\end{figure*}

\begin{figure*}
\begin{minipage}[c][\textheight][c]{\textwidth}
    \centering
    \begin{subfigure}{0.49\textwidth}
        \caption{Bias-Variance trade-off for \oracle-\ips}
        \includegraphics[width=\linewidth,trim={23cm 5.5cm 0 0},clip]{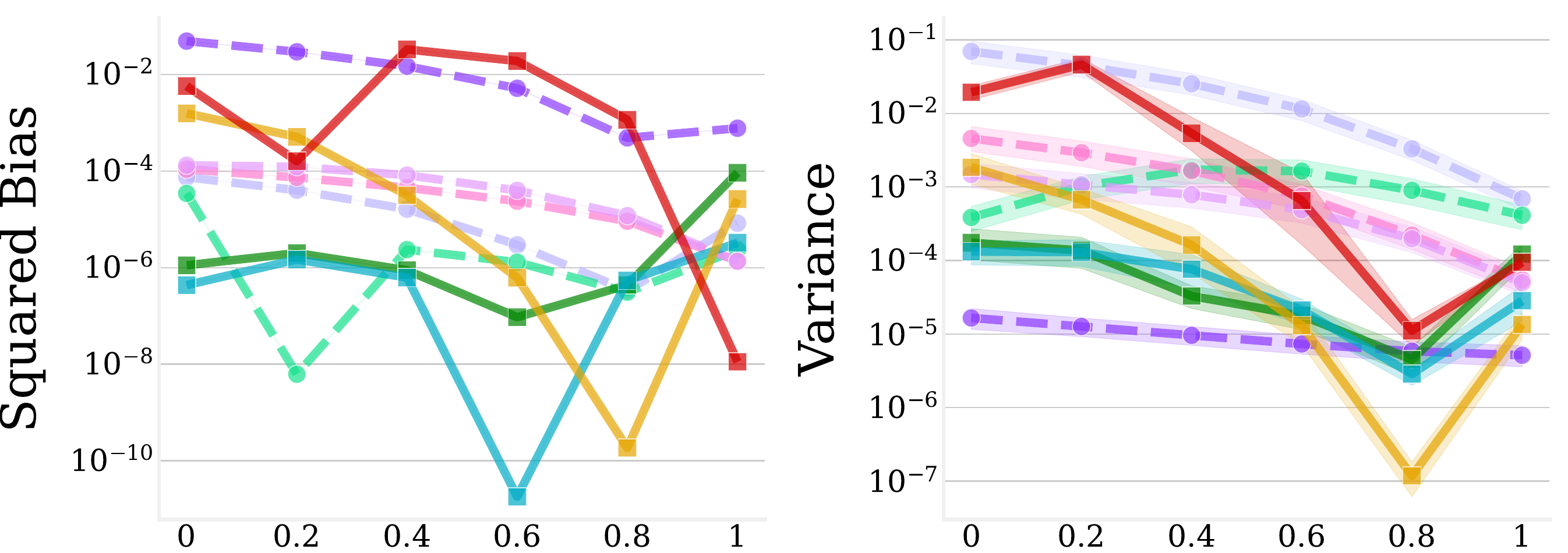}
    \end{subfigure} \hfill %
    \begin{subfigure}{0.49\textwidth}
        \caption{Bias-Variance trade-off for \oracle-\snips}
        \includegraphics[width=\linewidth,trim={23cm 5.5cm 0 0},clip]{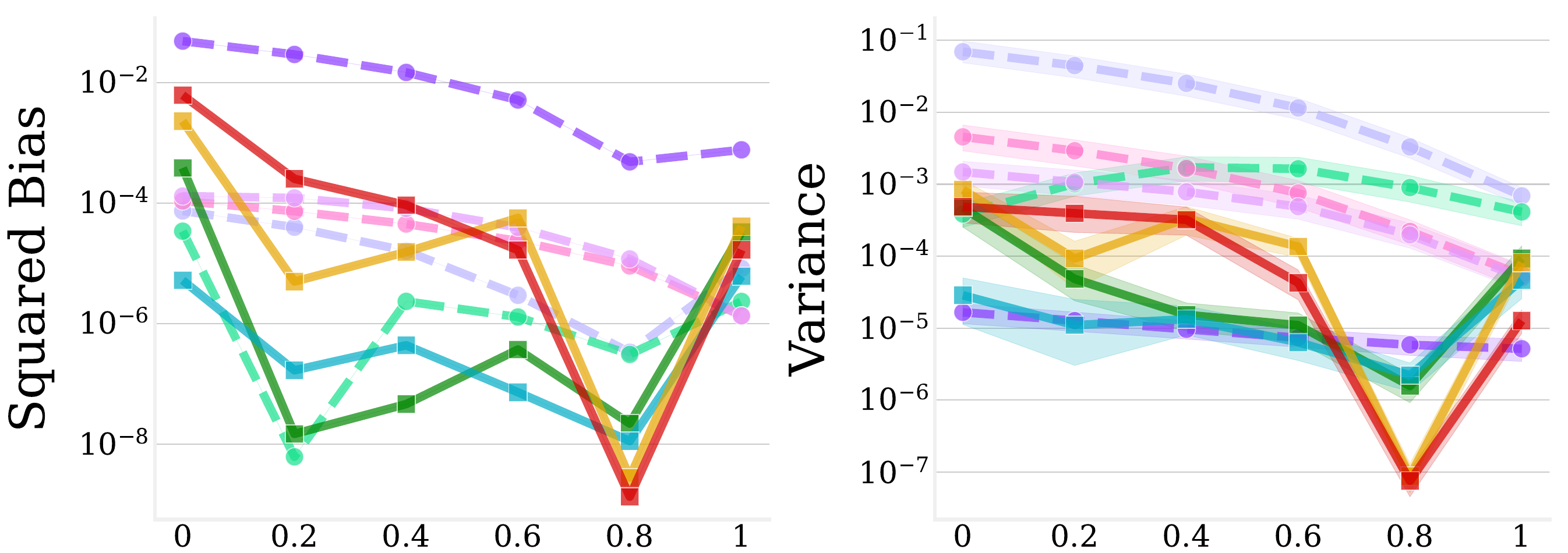}
    \end{subfigure}
    \begin{subfigure}{0.49\textwidth}
        \caption{Bias-Variance trade-off for \oracle-\dr}
        \includegraphics[width=\linewidth,trim={23cm 5.5cm 0 0},clip]{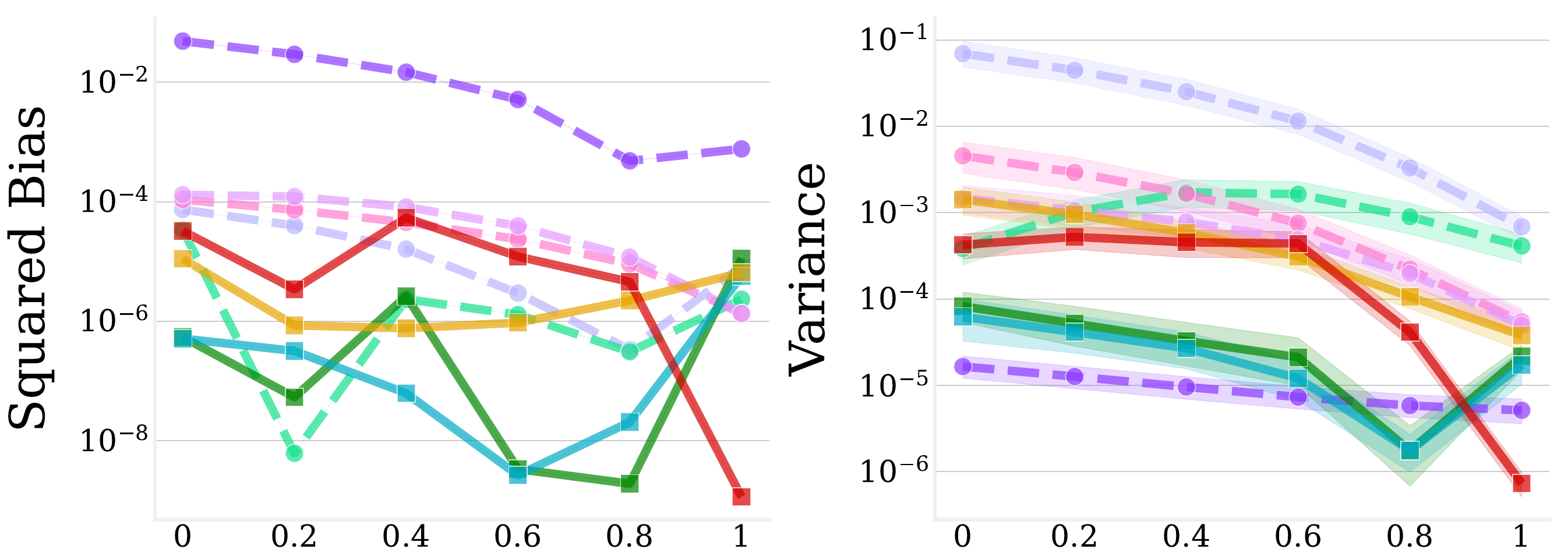}
    \end{subfigure} \hfill %
    \begin{subfigure}{0.49\textwidth}
        \caption{Bias-Variance trade-off for \oracle-\sndr}
        \includegraphics[width=\linewidth,trim={23cm 5.5cm 0 0},clip]{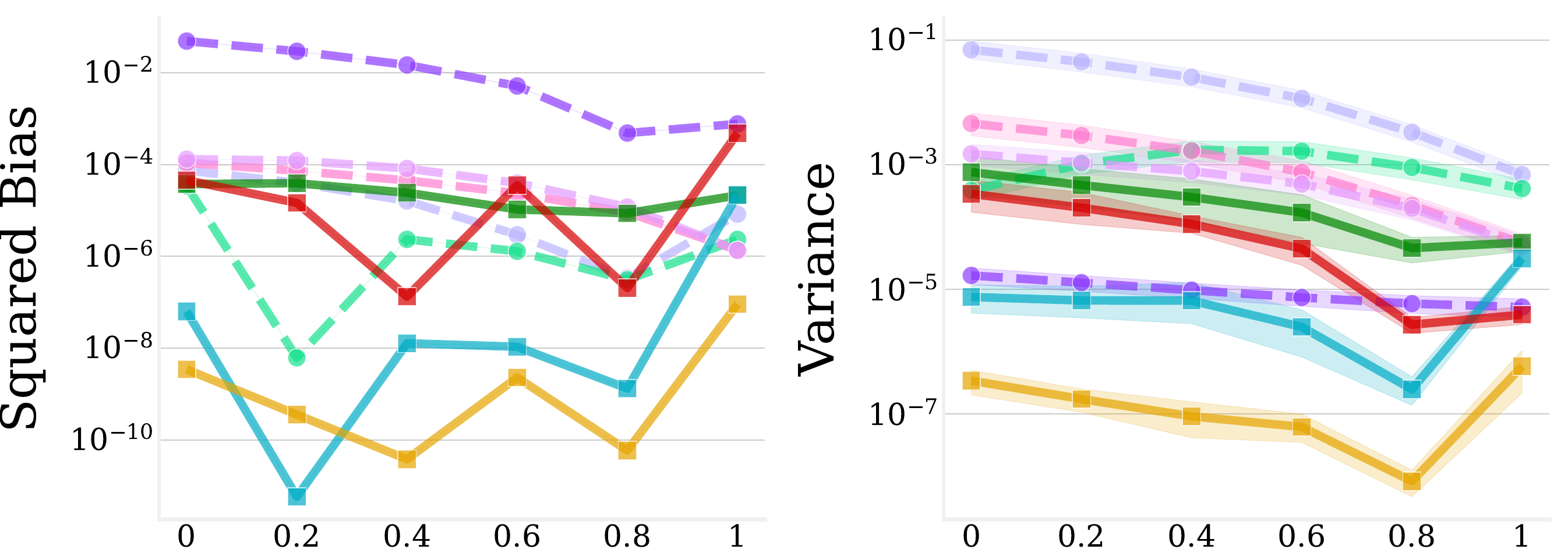}
    \end{subfigure}

    \vspace{0.3cm}
    
    \begin{subfigure}{\textwidth}
        \includegraphics[width=\linewidth,trim={0 0 0 0},clip]{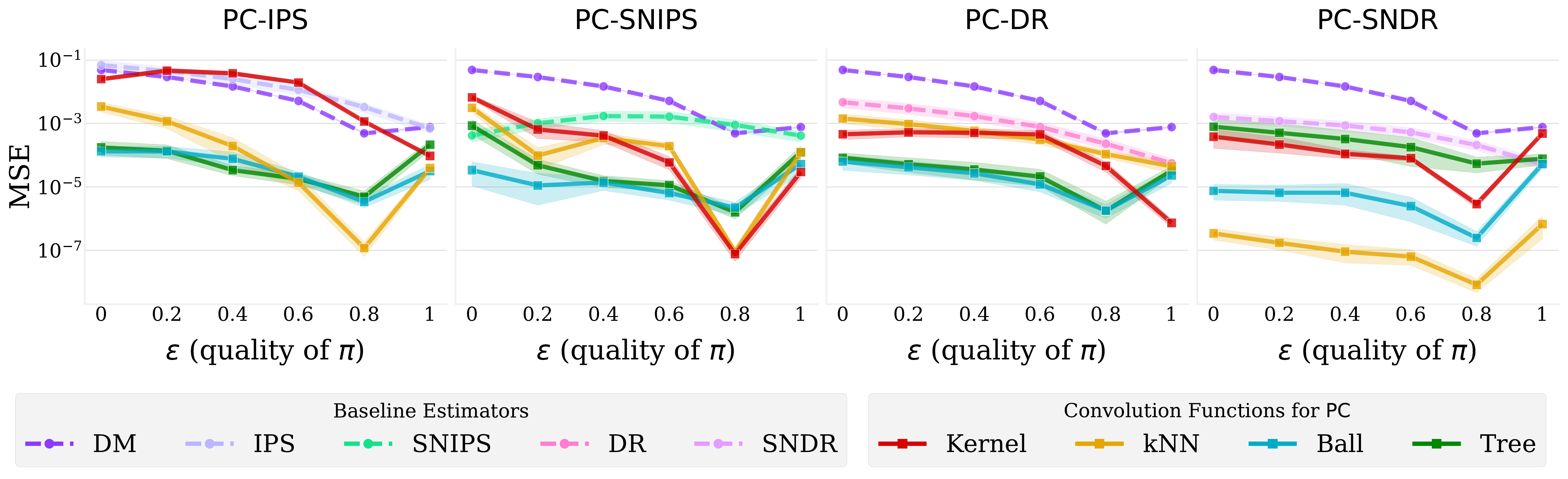}
    \end{subfigure} 
    \caption{Change in MSE, Squared Bias, and Variance while estimating various target policies ($\log$ scale) for the movielens dataset, using data logged by $\mu_{\mathsf{good}}$.}
    \label{fig:v_eps_ml_mu_good}
\end{minipage}
\end{figure*}

\clearpage

\begin{figure*}
\begin{minipage}[c][\textheight][c]{\textwidth}
    \centering
    \begin{subfigure}{\textwidth}
        \caption{\oracle-\ips}
        \vspace{0.2cm}
        \includegraphics[width=\linewidth,trim={0cm 5.2cm 0 0},clip]{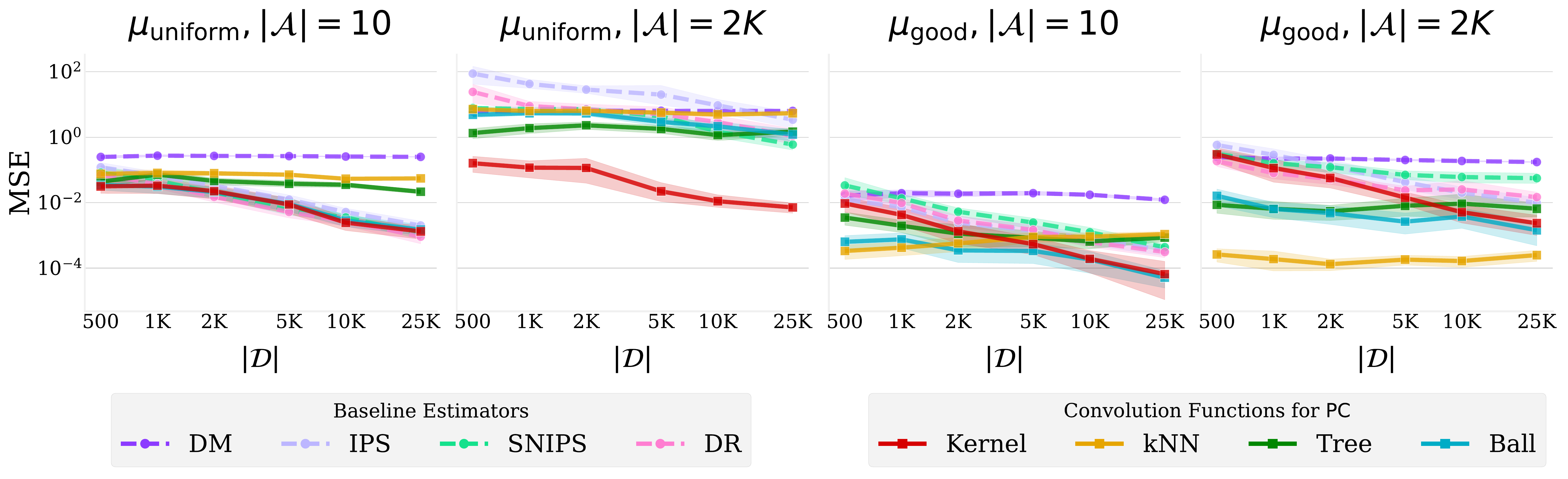}
    \end{subfigure}
    \begin{subfigure}{\textwidth}
        \caption{\oracle-\snips}
        \vspace{0.2cm}
        \includegraphics[width=\linewidth,trim={0cm 5.2cm 0 0},clip]{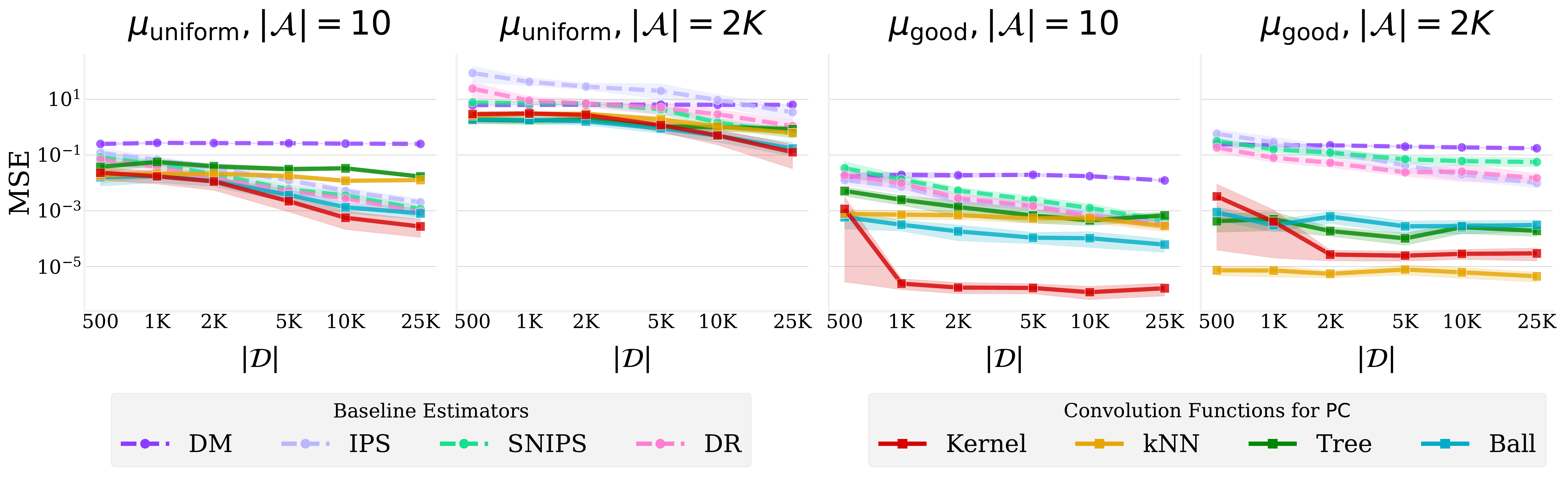}
    \end{subfigure}
    \begin{subfigure}{\textwidth}
        \caption{\oracle-\dr}
        \vspace{0.2cm}
        \includegraphics[width=\linewidth,trim={0cm 5.2cm 0 0},clip]{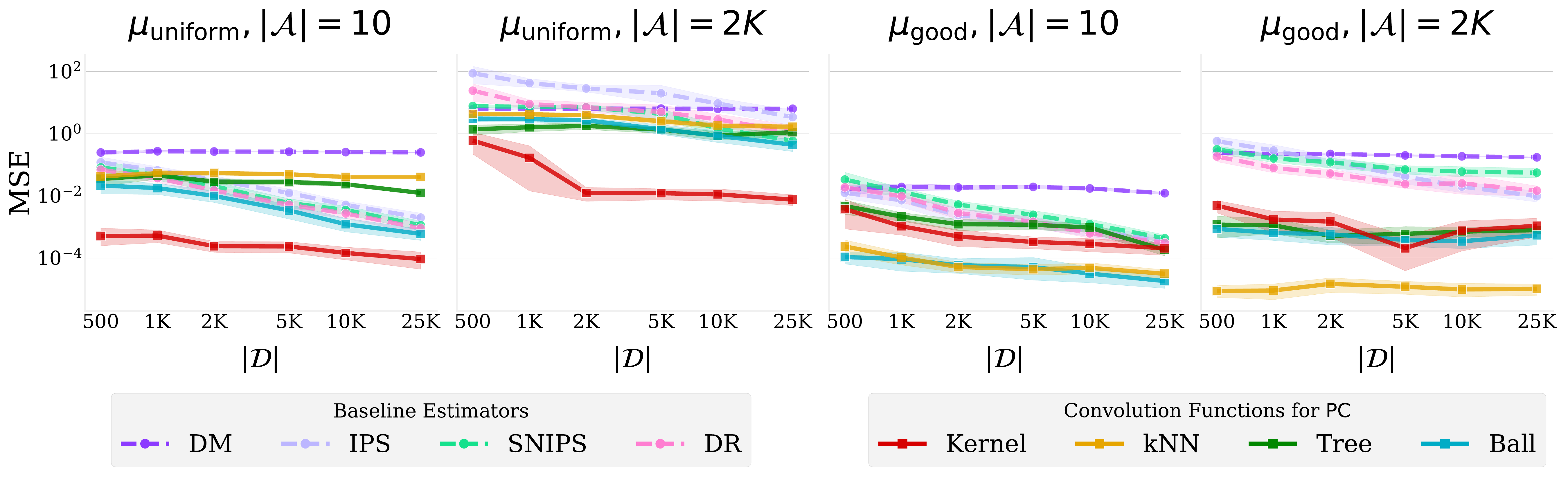}
    \end{subfigure}
    \begin{subfigure}{\textwidth}
        \caption{\oracle-\sndr}
        \vspace{0.2cm}
        \includegraphics[width=\linewidth,trim={0cm 0 0 0},clip]{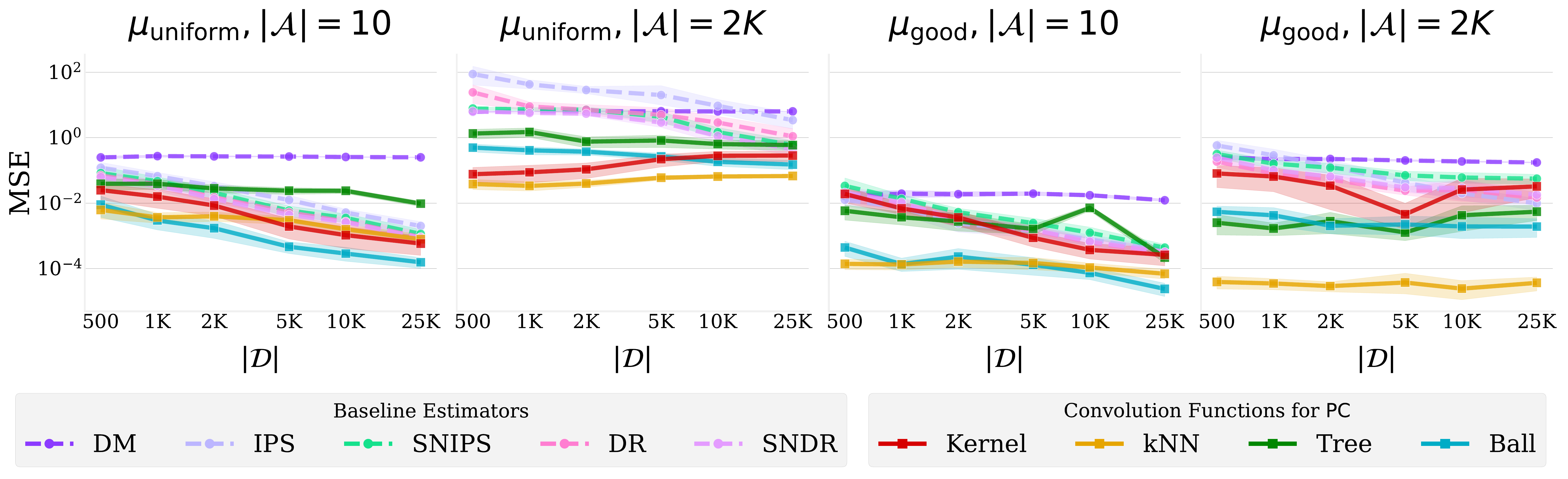}
    \end{subfigure}
    \vspace{0.001cm}
    \caption{Change in MSE while estimating $V(\pi_{\mathsf{good}})$ with varying amount of bandit feedback ($\log$-$\log$ scale) for the synthetic dataset.}
    \label{fig:v_data_synthetic_pi_good}
\end{minipage}
\end{figure*}

\begin{figure*}
\begin{minipage}[c][\textheight][c]{\textwidth}
    \centering
    \begin{subfigure}{\textwidth}
        \caption{\oracle-\ips}
        \vspace{0.2cm}
        \includegraphics[width=\linewidth,trim={0cm 5.2cm 0 0},clip]{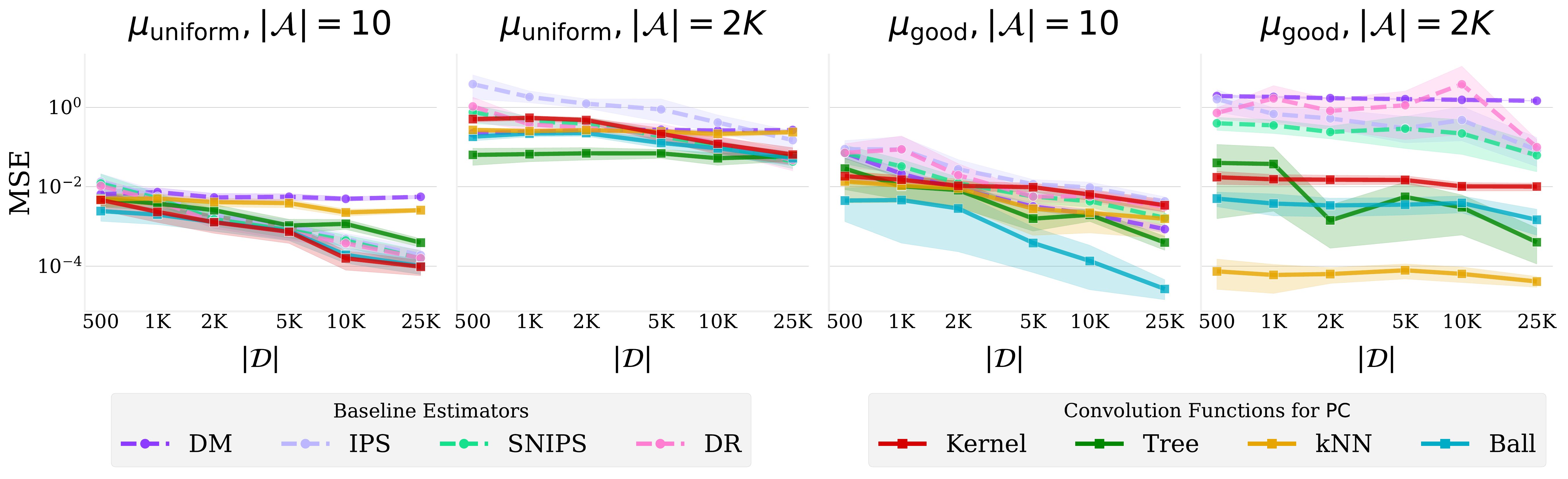}
    \end{subfigure}
    \begin{subfigure}{\textwidth}
        \caption{\oracle-\snips}
        \vspace{0.2cm}
        \includegraphics[width=\linewidth,trim={0cm 5.2cm 0 0},clip]{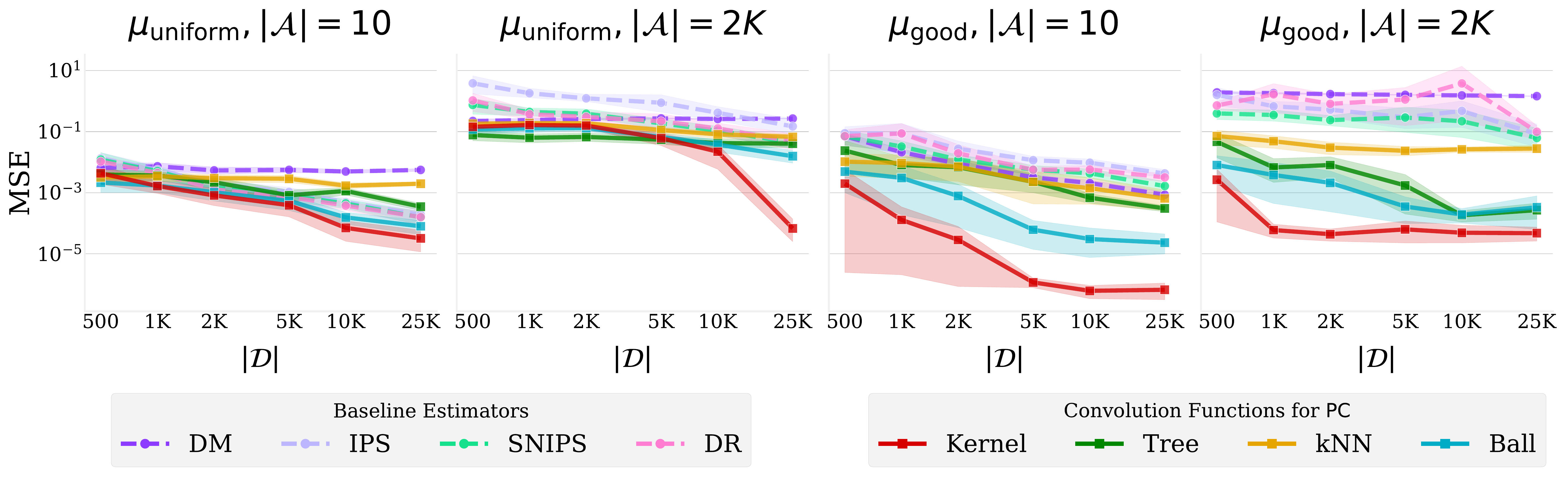}
    \end{subfigure}
    \begin{subfigure}{\textwidth}
        \caption{\oracle-\dr}
        \vspace{0.2cm}
        \includegraphics[width=\linewidth,trim={0cm 5.2cm 0 0},clip]{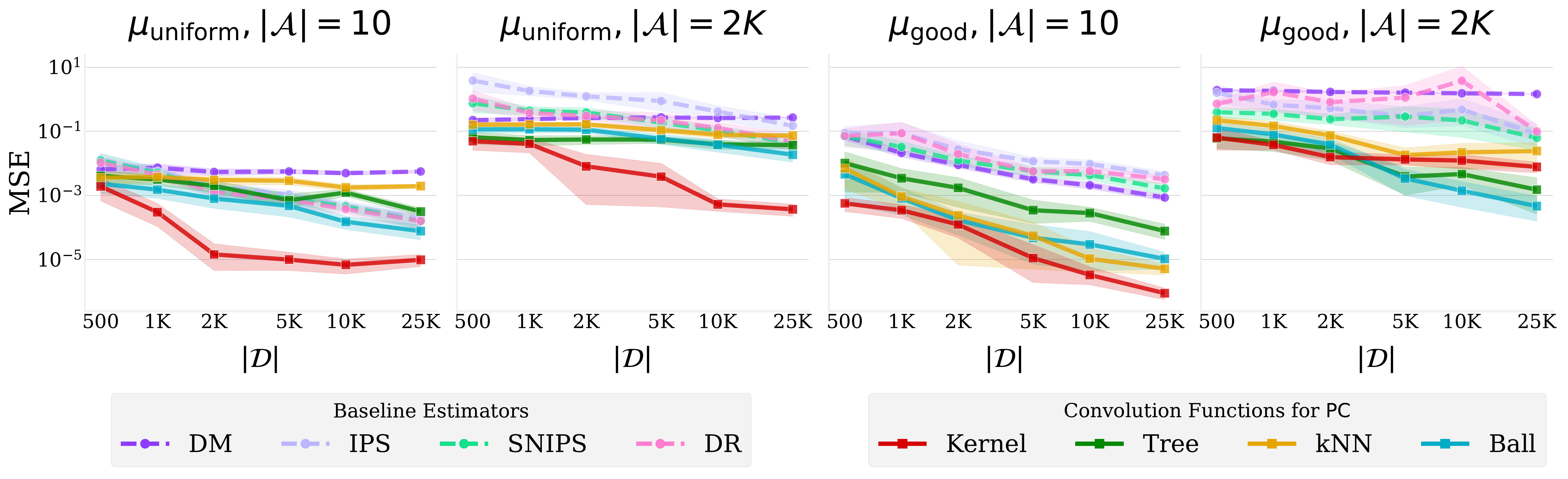}
    \end{subfigure}
    \begin{subfigure}{\textwidth}
        \caption{\oracle-\sndr}
        \vspace{0.2cm}
        \includegraphics[width=\linewidth,trim={0cm 0 0 0},clip]{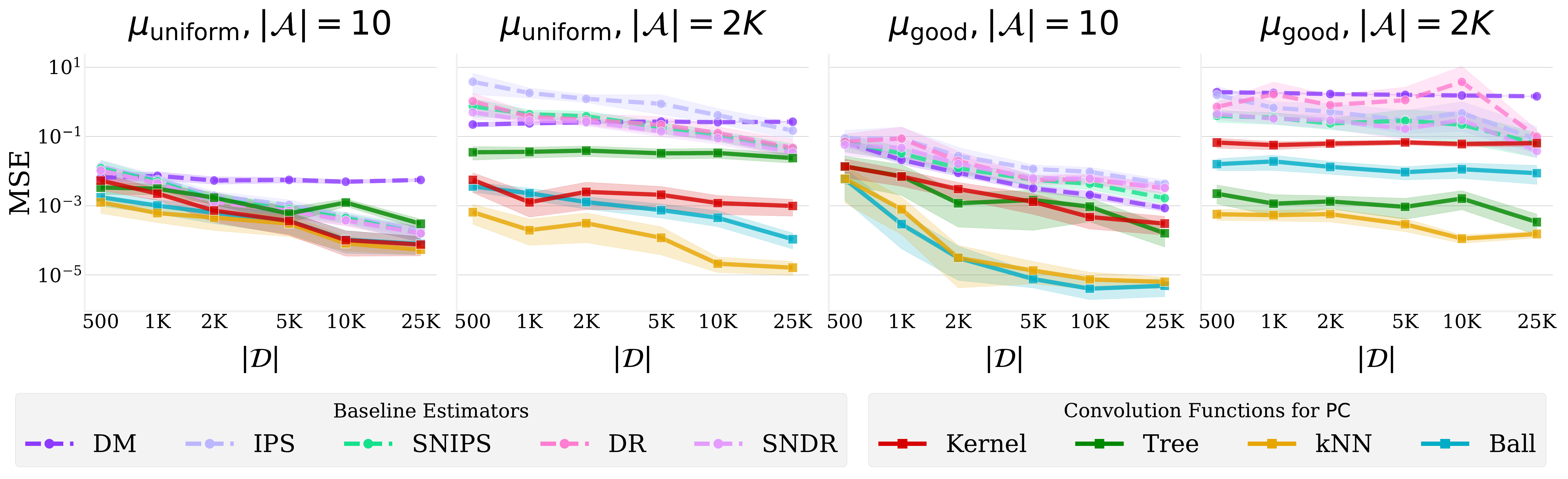}
    \end{subfigure}
    \vspace{0.001cm}
    \caption{Change in MSE while estimating $V(\pi_{\mathsf{bad}})$ with varying amount of bandit feedback ($\log$-$\log$ scale) for the synthetic dataset.}
    \label{fig:v_data_synthetic_pi_bad}
\end{minipage}
\end{figure*}

\begin{figure*}
\begin{minipage}[c][\textheight][c]{\textwidth}
    \centering
    \vspace{-0.3cm}
    \begin{subfigure}{0.495\textwidth}
        \caption{\oracle-\ips}
        \vspace{0.1cm}
        \includegraphics[width=\linewidth,trim={19.5cm 5.2cm 24cm 0},clip]{figures/custom_n_data/ml-100k/eps_0.02/IPS.pdf}
    \end{subfigure} \hfill %
    \begin{subfigure}{0.495\textwidth}
        \caption{\oracle-\snips}
        \vspace{0.1cm}
        \includegraphics[width=\linewidth,trim={19.5cm 5.2cm 24cm 0},clip]{figures/custom_n_data/ml-100k/eps_0.02/SNIPS.pdf}
    \end{subfigure}
    \vspace{-0.3cm}
    \begin{subfigure}{0.495\textwidth}
        \caption{\oracle-\dr}
        \vspace{0.1cm}
        \includegraphics[width=\linewidth,trim={19.5cm 4.7cm 24cm 0},clip]{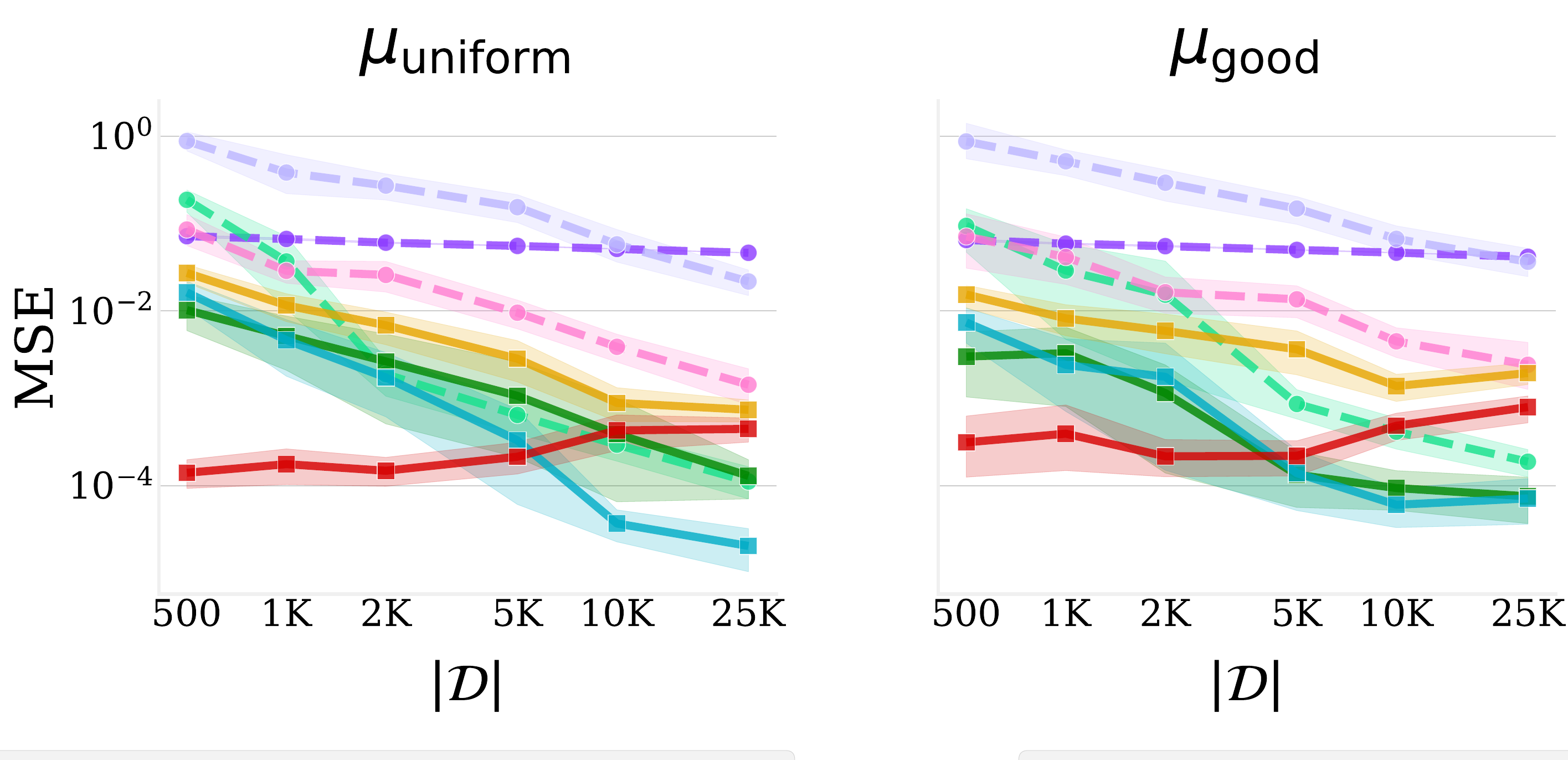}
    \end{subfigure} \hfill %
    \begin{subfigure}{0.495\textwidth}
        \caption{\oracle-\sndr}
        \vspace{0.1cm}
        \includegraphics[width=\linewidth,trim={19.5cm 4.7cm 24cm 0},clip]{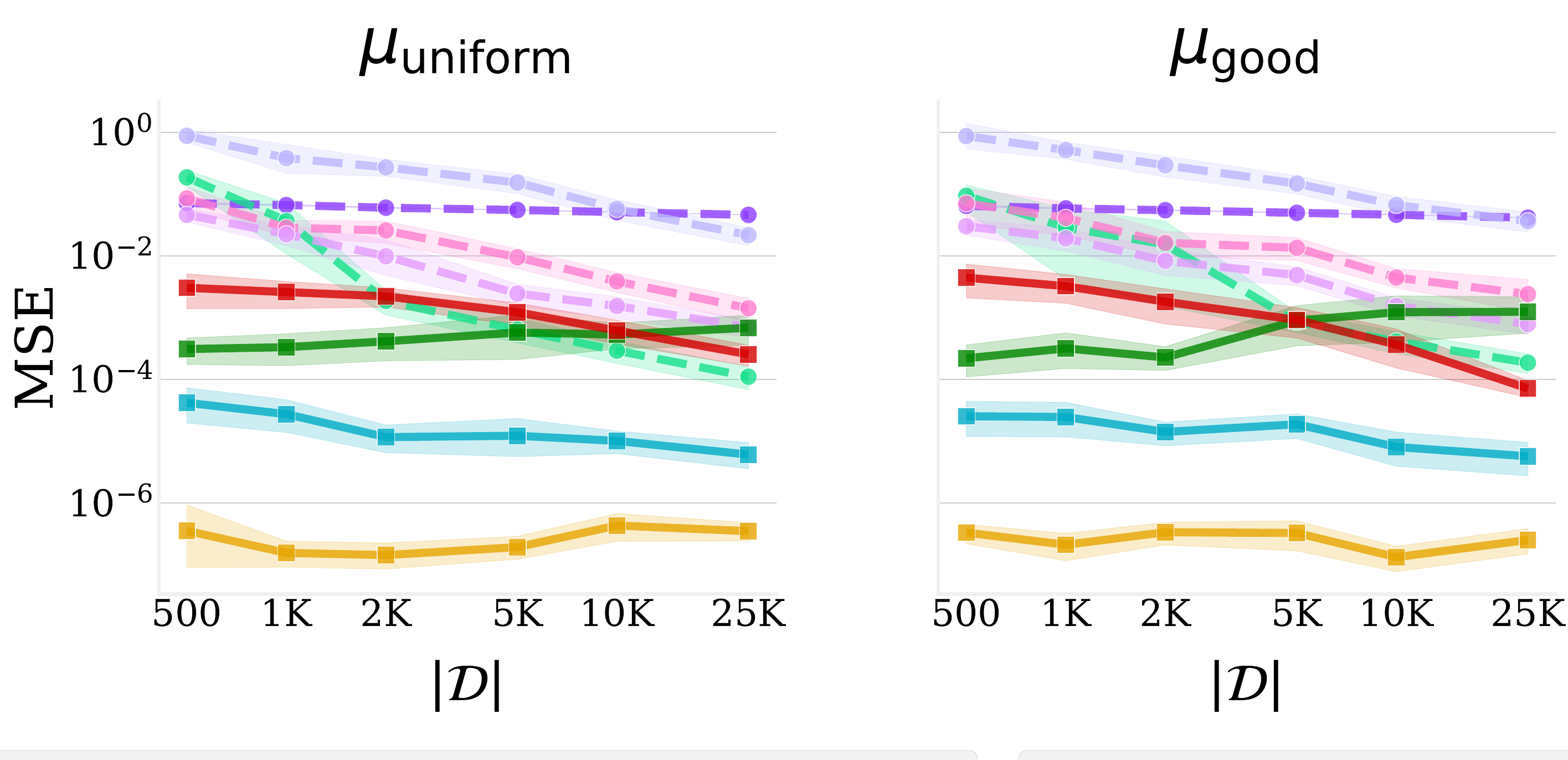}
    \end{subfigure}
    \includegraphics[width=\linewidth,trim={0 0 0 22cm},clip]{figures/custom_n_data/ml-100k/eps_0.02/SNDR.pdf}
    \vspace{0.001cm}
    \caption{Change in MSE while estimating $V(\pi_{\mathsf{good}})$ with varying amount of bandit feedback ($\log$-$\log$ scale) for the movielens dataset.}
    \label{fig:v_data_ml_pi_good}
\end{minipage}
\end{figure*}

\begin{figure*}
\begin{minipage}[c][\textheight][c]{\textwidth}
    \centering
    \vspace{-0.3cm}
    \begin{subfigure}{0.495\textwidth}
        \caption{\oracle-\ips}
        \vspace{0.1cm}
        \includegraphics[width=\linewidth,trim={19.5cm 5.2cm 24cm 0},clip]{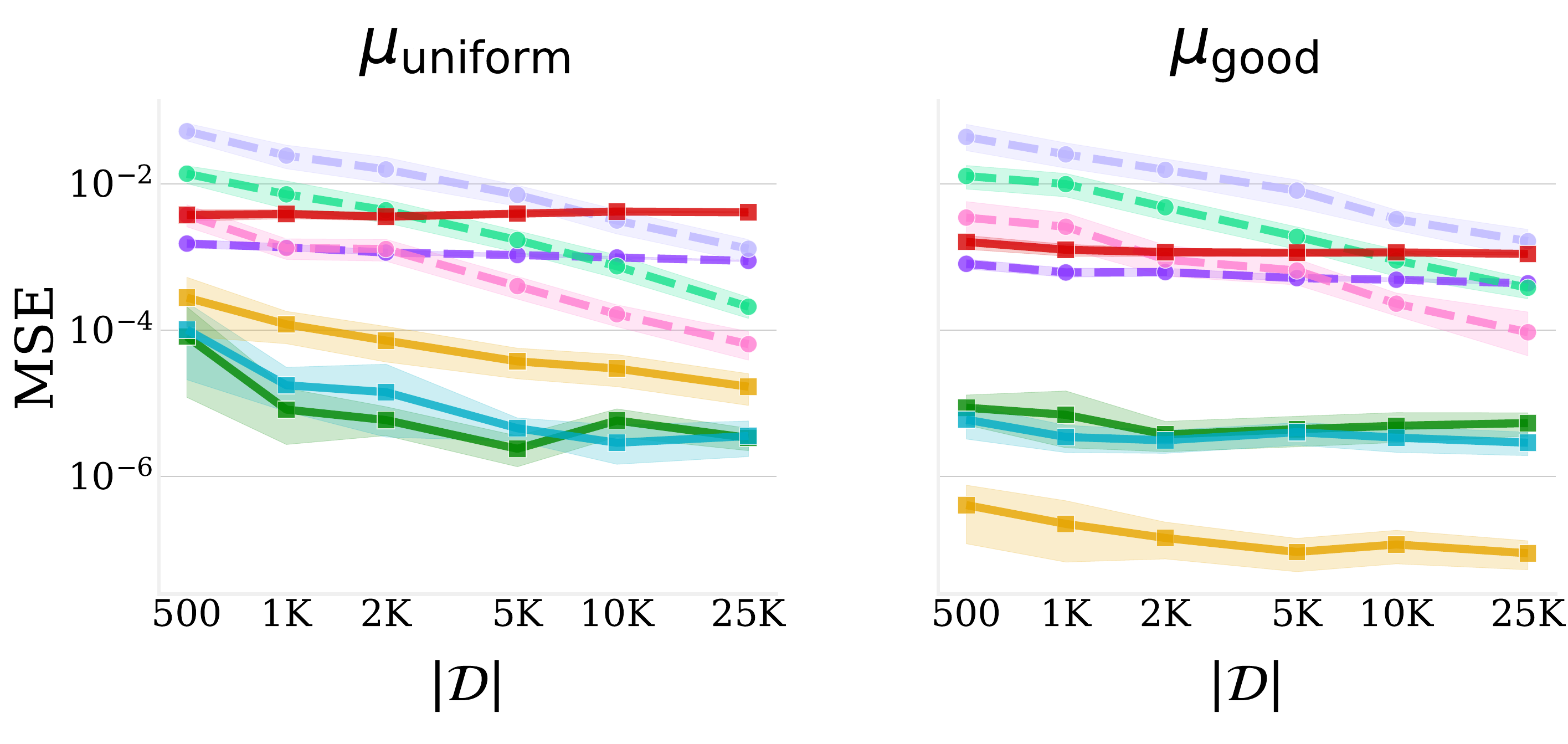}
    \end{subfigure} \hfill %
    \begin{subfigure}{0.495\textwidth}
        \caption{\oracle-\snips}
        \vspace{0.1cm}
        \includegraphics[width=\linewidth,trim={19.5cm 5.2cm 24cm 0},clip]{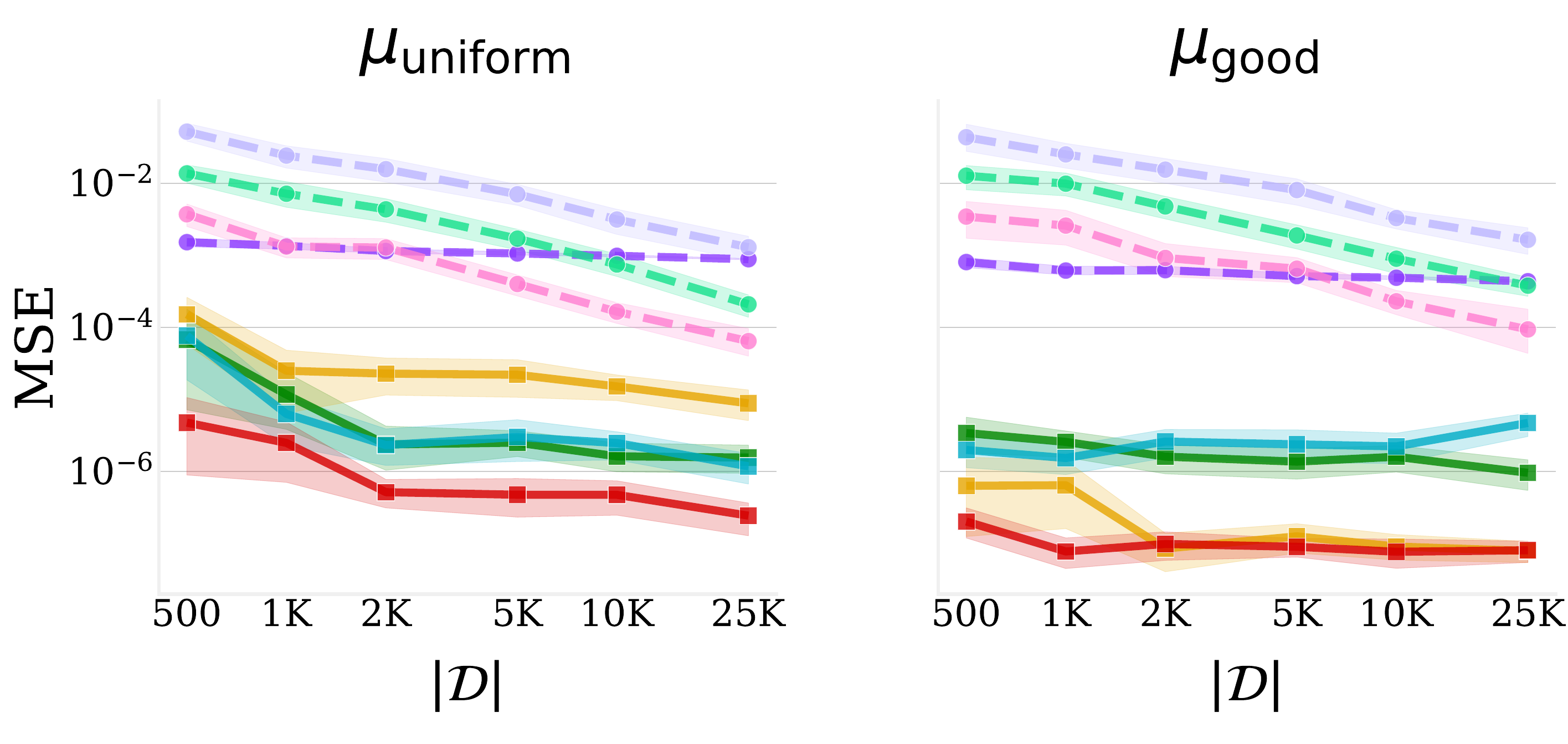}
    \end{subfigure}
    \vspace{-0.3cm}
    \begin{subfigure}{0.495\textwidth}
        \caption{\oracle-\dr}
        \vspace{0.1cm}
        \includegraphics[width=\linewidth,trim={19.5cm 4.7cm 24cm 0},clip]{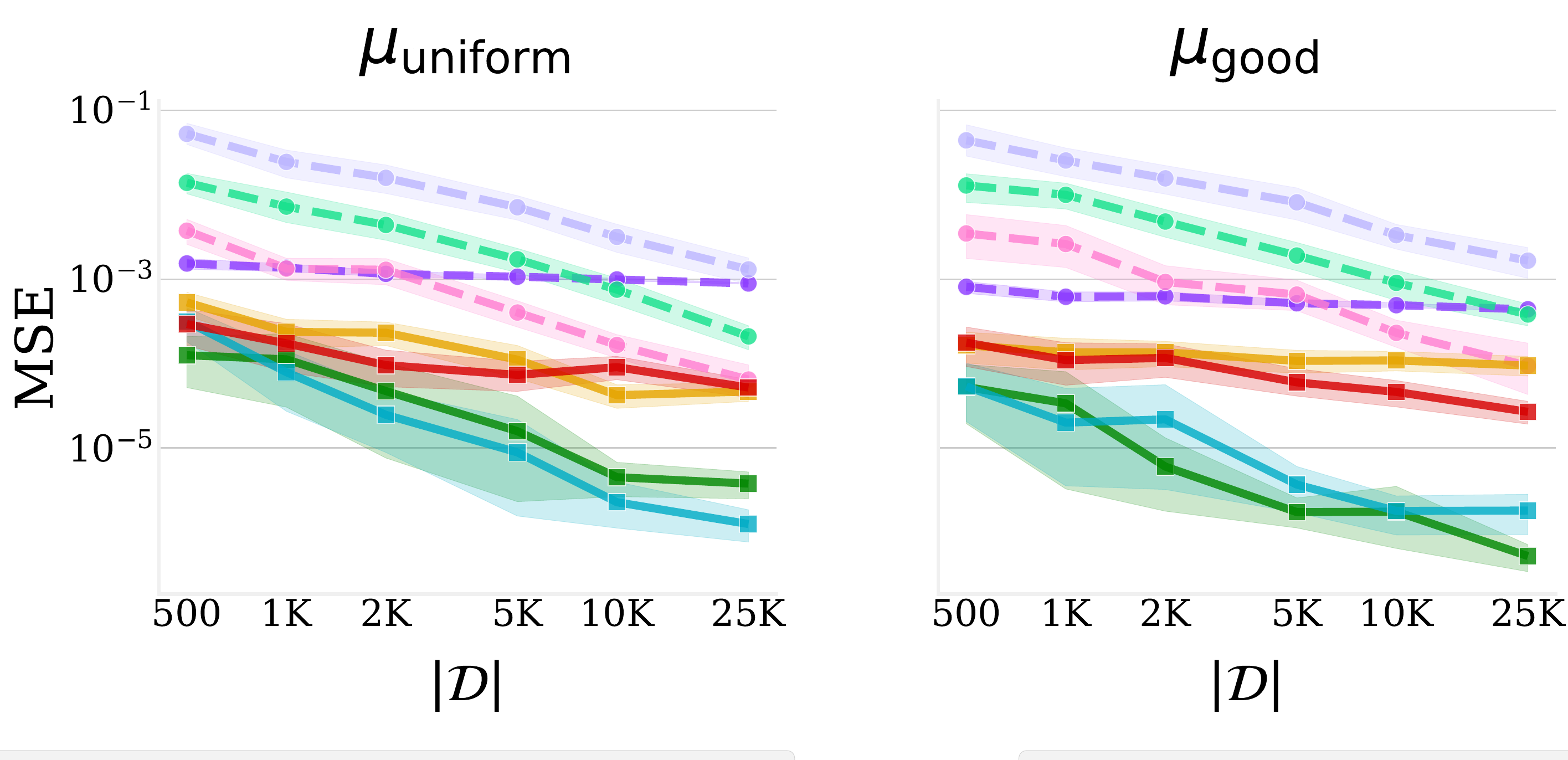}
    \end{subfigure} \hfill %
    \begin{subfigure}{0.495\textwidth}
        \caption{\oracle-\sndr}
        \vspace{0.1cm}
        \includegraphics[width=\linewidth,trim={19.5cm 4.7cm 24cm 0},clip]{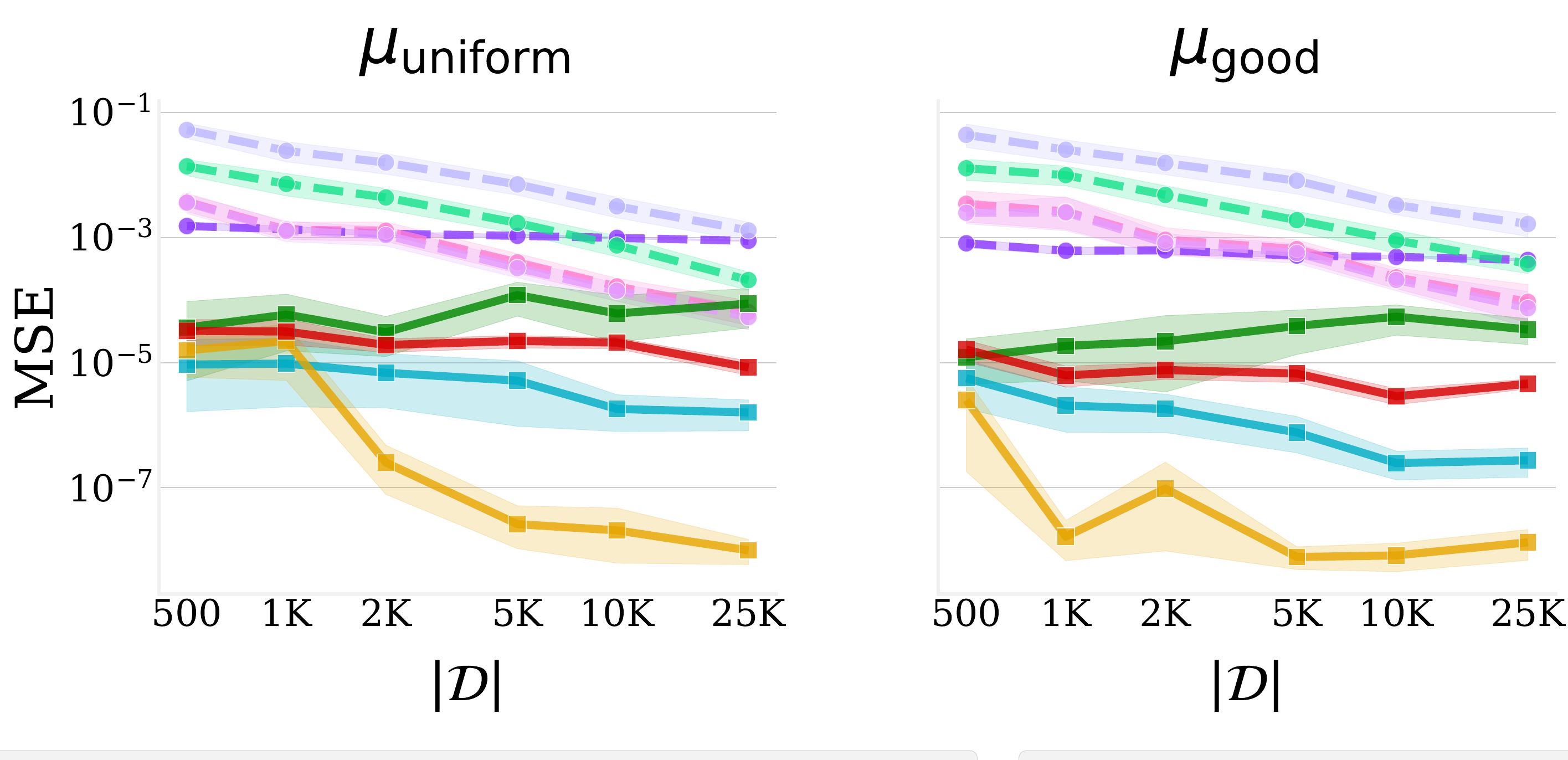}
    \end{subfigure}
    \includegraphics[width=\linewidth,trim={0 0 0 22cm},clip]{figures/custom_n_data/ml-100k/eps_0.8/SNDR.pdf}
    \vspace{0.001cm}
    \caption{Change in MSE while estimating $V(\pi_{\mathsf{bad}})$ with varying amount of bandit feedback ($\log$-$\log$ scale) for the movielens dataset.}
    \label{fig:v_data_ml_pi_bad}
\end{minipage}
\end{figure*}

\clearpage

\begin{figure*}
\begin{minipage}[c][\textheight][c]{\textwidth}
    \centering
    \begin{subfigure}{0.49\textwidth}
        \caption{Bias-Variance trade-off for \oracle-\ips}
        \includegraphics[width=\linewidth,trim={23cm 5.5cm 0 0},clip]{appendix_figures/eps_0.05/varying_support/synthetic_beta_0/bias_variance/IPS.pdf}
    \end{subfigure} \hfill %
    \begin{subfigure}{0.49\textwidth}
        \caption{Bias-Variance trade-off for \oracle-\snips}
        \includegraphics[width=\linewidth,trim={23cm 5.5cm 0 0},clip]{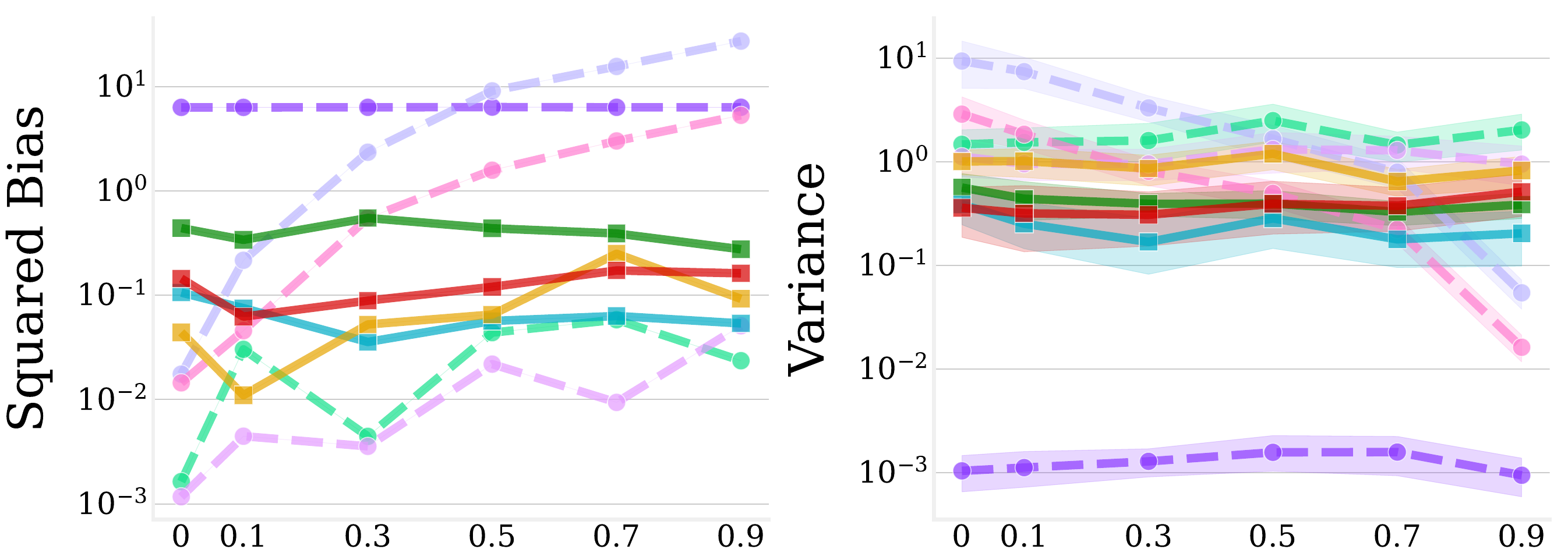}
    \end{subfigure}
    \begin{subfigure}{0.49\textwidth}
        \caption{Bias-Variance trade-off for \oracle-\dr}
        \includegraphics[width=\linewidth,trim={23cm 5.5cm 0 0},clip]{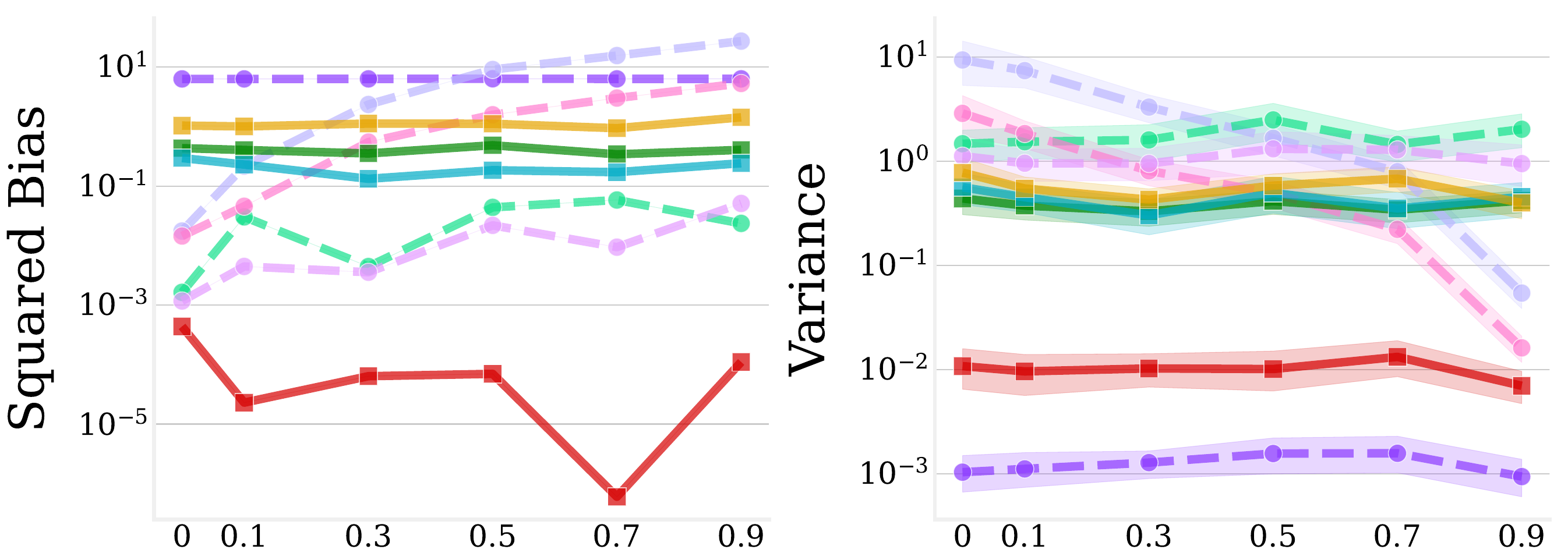}
    \end{subfigure} \hfill %
    \begin{subfigure}{0.49\textwidth}
        \caption{Bias-Variance trade-off for \oracle-\sndr}
        \includegraphics[width=\linewidth,trim={23cm 5.5cm 0 0},clip]{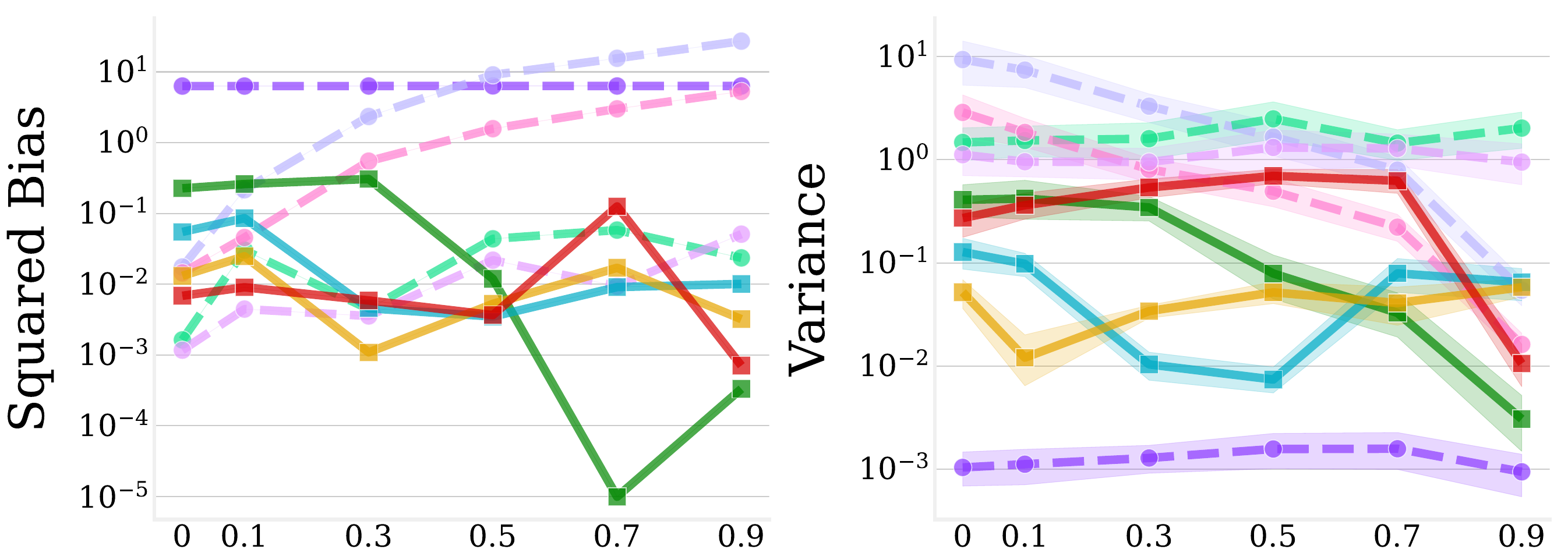}
    \end{subfigure}

    \vspace{0.3cm}
    
    \begin{subfigure}{\textwidth}
        \includegraphics[width=\linewidth,trim={0 0 0 0},clip]{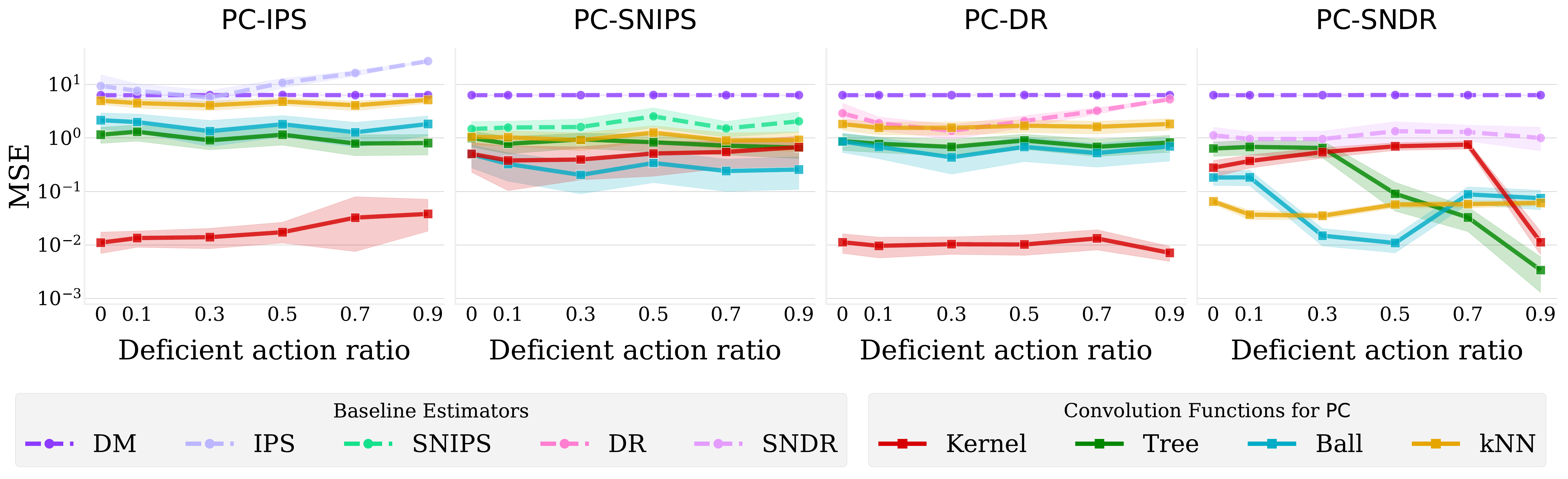}
    \end{subfigure} 
    \caption{Change in MSE, Squared Bias, and Variance while estimating $V(\pi_{\mathsf{good}})$ with varying support ($\log$-$\log$ scale) for the synthetic dataset (with $2000$ actions), using data logged by $\mu_{\mathsf{uniform}}$.}
    \label{fig:v_support_pi_good_mu_unif}
\end{minipage}
\end{figure*}

\begin{figure*}
\begin{minipage}[c][\textheight][c]{\textwidth}
    \centering
    \begin{subfigure}{0.49\textwidth}
        \caption{Bias-Variance trade-off for \oracle-\ips}
        \includegraphics[width=\linewidth,trim={23cm 5.5cm 0 0},clip]{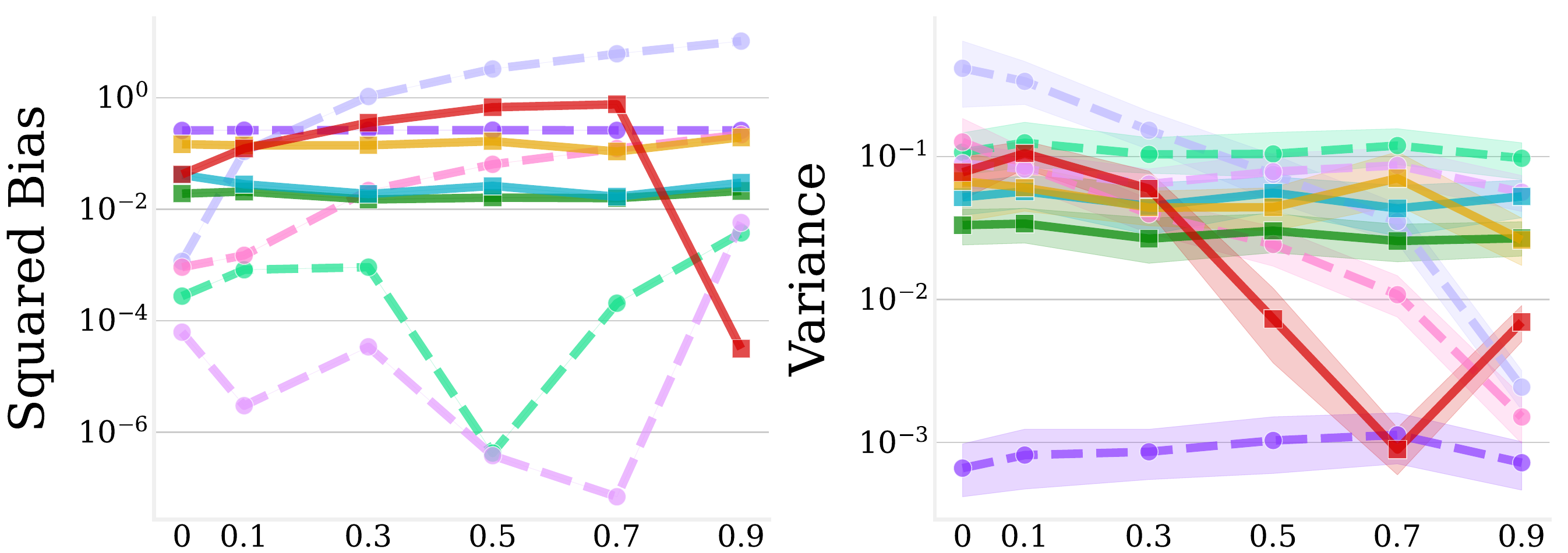}
    \end{subfigure} \hfill %
    \begin{subfigure}{0.49\textwidth}
        \caption{Bias-Variance trade-off for \oracle-\snips}
        \includegraphics[width=\linewidth,trim={23cm 5.5cm 0 0},clip]{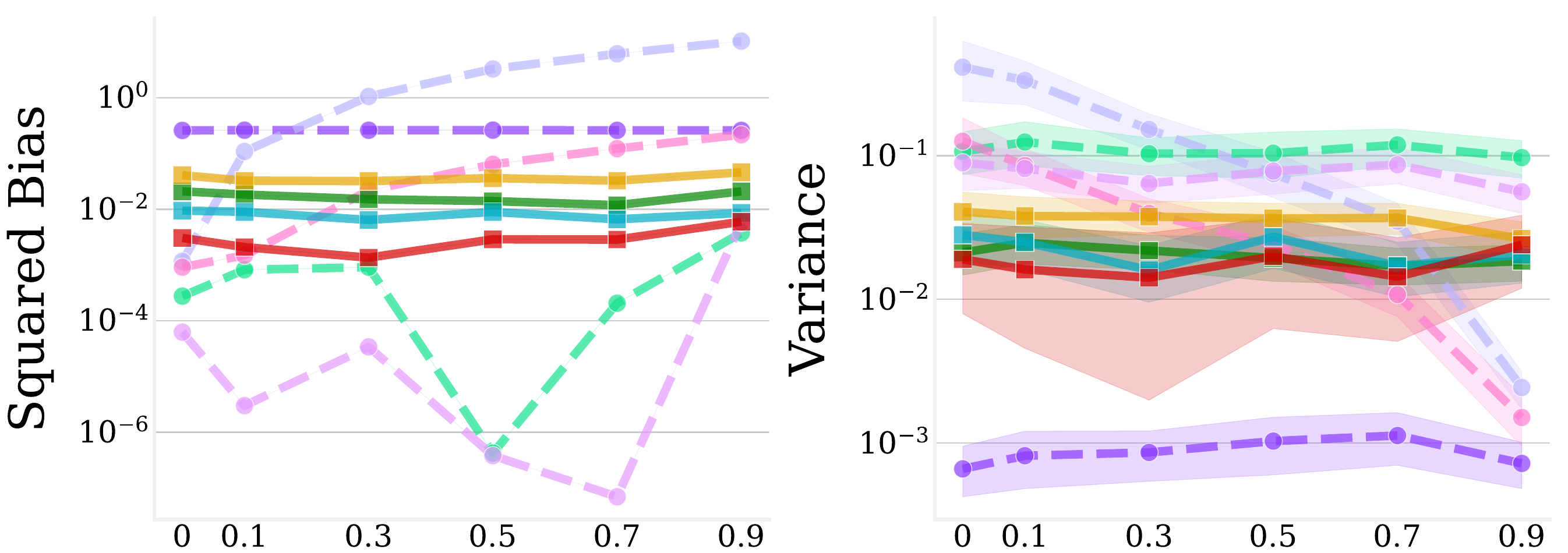}
    \end{subfigure}
    \begin{subfigure}{0.49\textwidth}
        \caption{Bias-Variance trade-off for \oracle-\dr}
        \includegraphics[width=\linewidth,trim={23cm 5.5cm 0 0},clip]{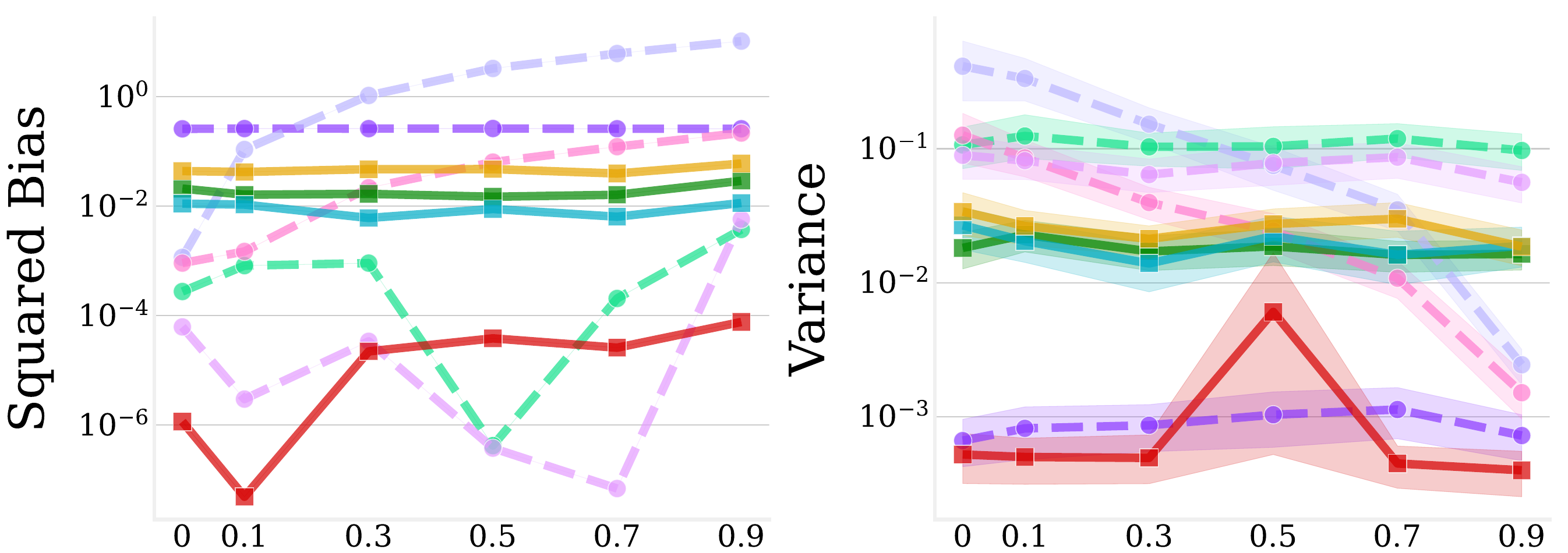}
    \end{subfigure} \hfill %
    \begin{subfigure}{0.49\textwidth}
        \caption{Bias-Variance trade-off for \oracle-\sndr}
        \includegraphics[width=\linewidth,trim={23cm 5.5cm 0 0},clip]{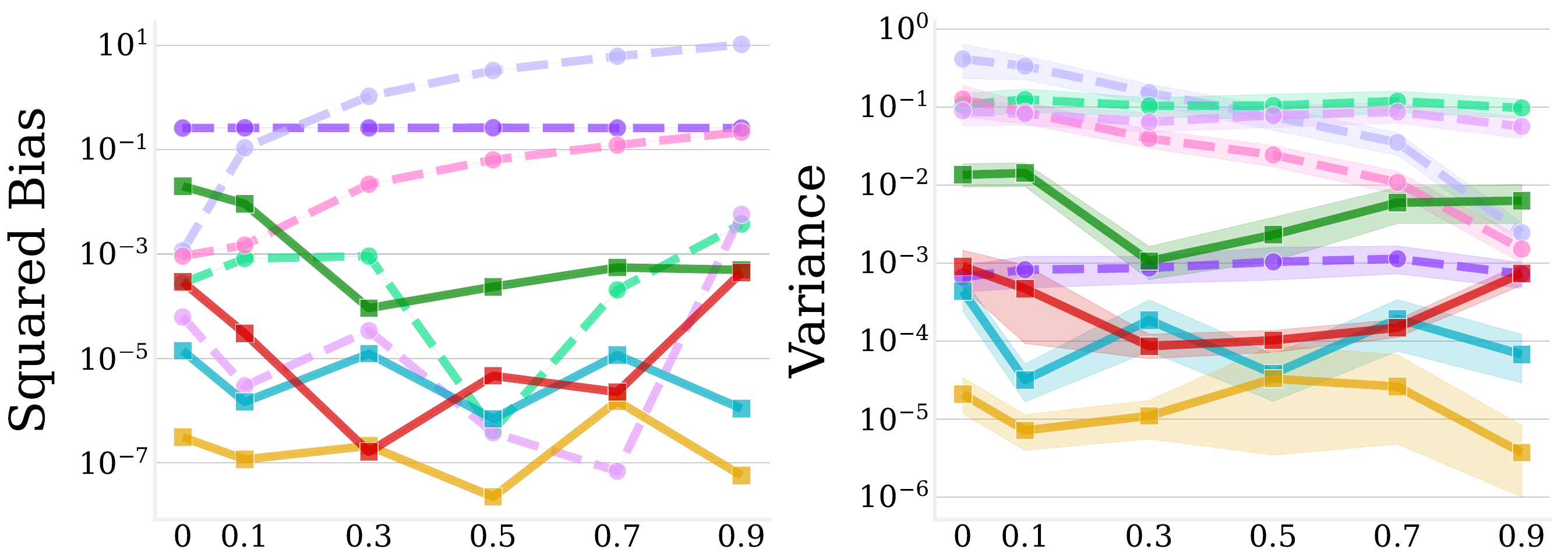}
    \end{subfigure}

    \vspace{0.3cm}
    
    \begin{subfigure}{\textwidth}
        \includegraphics[width=\linewidth,trim={0 0 0 0},clip]{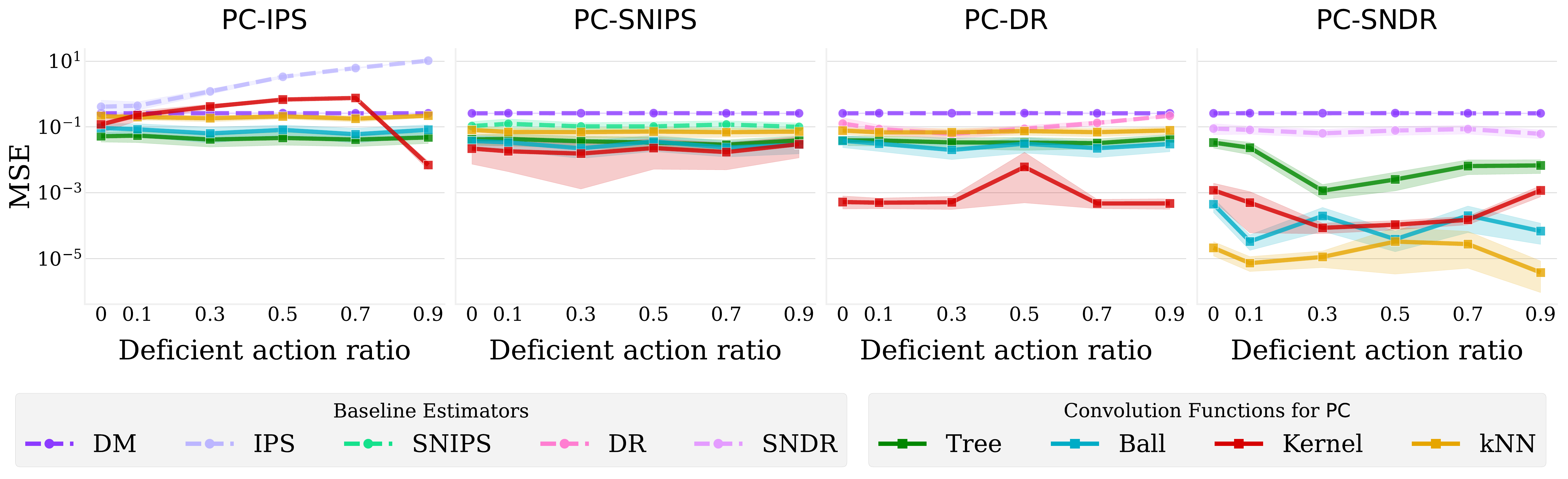}
    \end{subfigure} 
    \caption{Change in MSE, Squared Bias, and Variance while estimating $V(\pi_{\mathsf{bad}})$ with varying support ($\log$-$\log$ scale) for the synthetic dataset (with $2000$ actions), using data logged by $\mu_{\mathsf{uniform}}$.}
    \label{fig:v_support_pi_bad_mu_unif}
\end{minipage}
\end{figure*}

\clearpage

\begin{figure*}
\begin{minipage}[c][\textheight][c]{\textwidth}
    \centering
    \begin{subfigure}{0.49\textwidth}
        \caption{Bias-Variance trade-off for \oracle-\ips}
        \includegraphics[width=\linewidth,trim={23cm 5.5cm 0 0},clip]{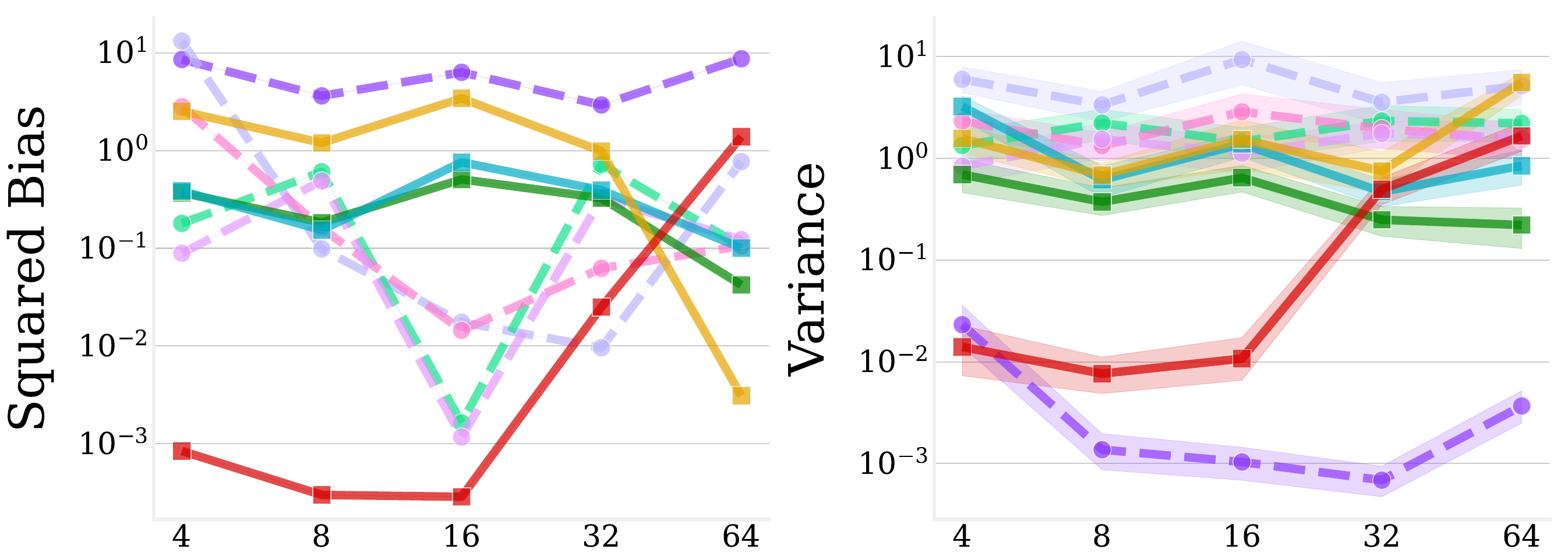}
    \end{subfigure} \hfill %
    \begin{subfigure}{0.49\textwidth}
        \caption{Bias-Variance trade-off for \oracle-\snips}
        \includegraphics[width=\linewidth,trim={23cm 5.5cm 0 0},clip]{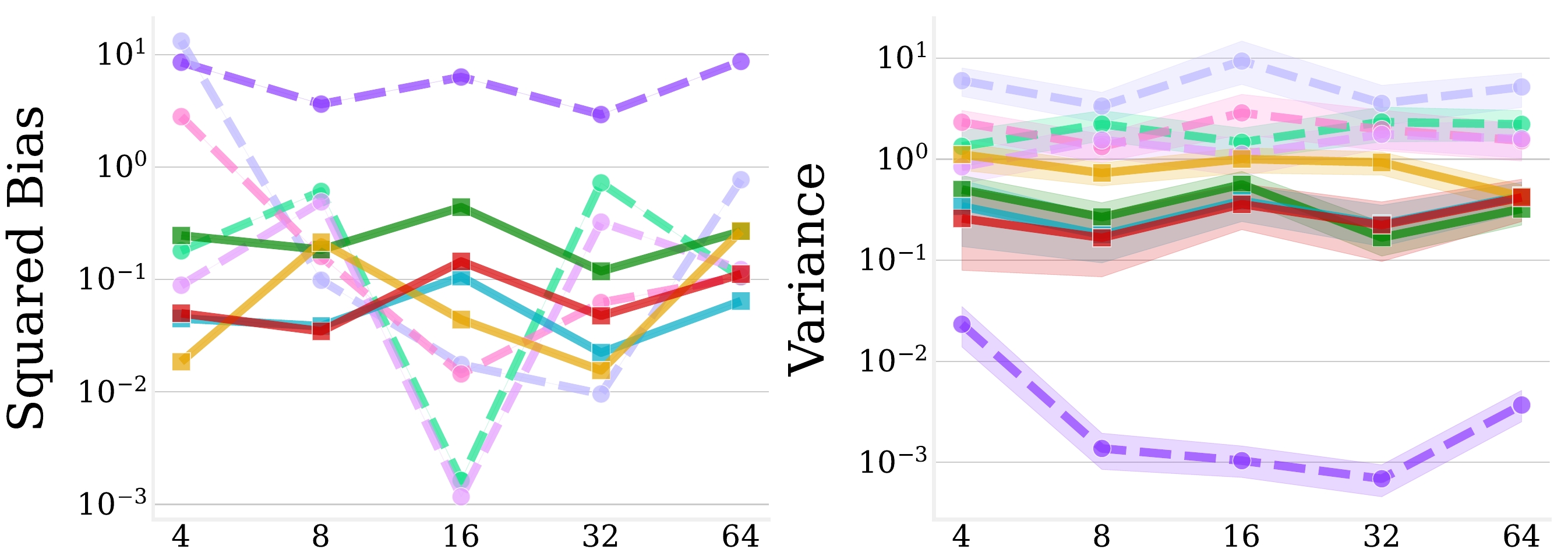}
    \end{subfigure}
    \begin{subfigure}{0.49\textwidth}
        \caption{Bias-Variance trade-off for \oracle-\dr}
        \includegraphics[width=\linewidth,trim={23cm 5.5cm 0 0},clip]{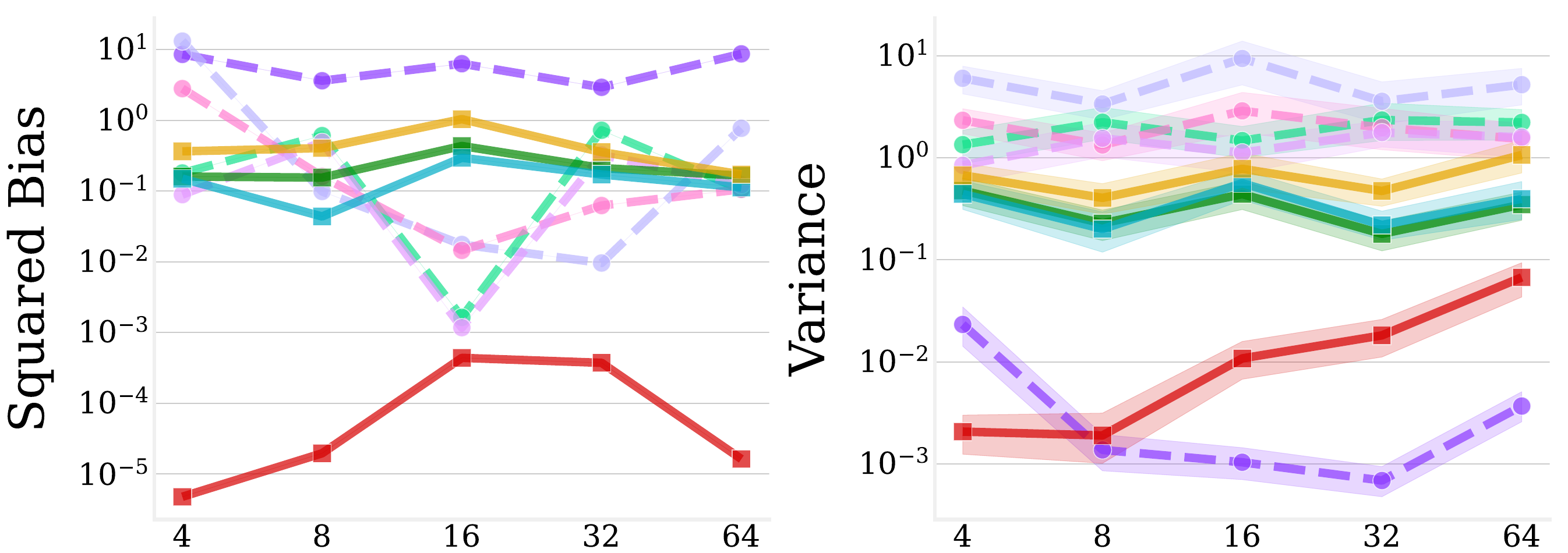}
    \end{subfigure} \hfill %
    \begin{subfigure}{0.49\textwidth}
        \caption{Bias-Variance trade-off for \oracle-\sndr}
        \includegraphics[width=\linewidth,trim={23cm 5.5cm 0 0},clip]{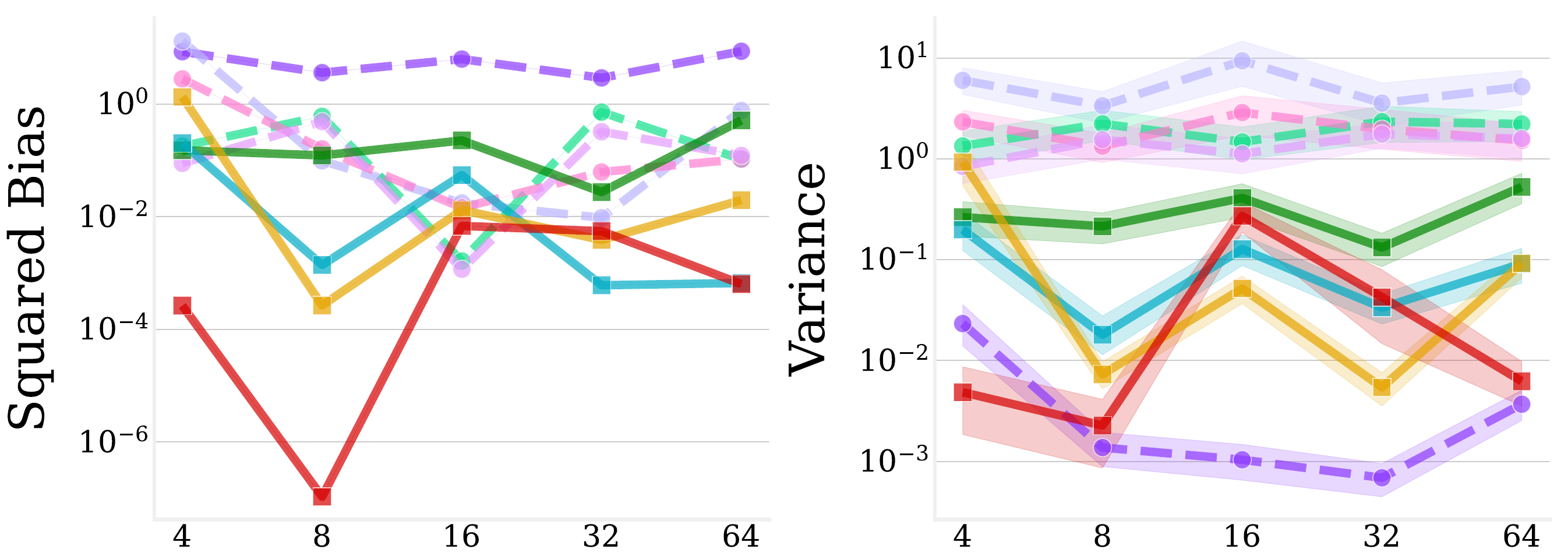}
    \end{subfigure}

    \vspace{0.3cm}
    
    \begin{subfigure}{\textwidth}
        \includegraphics[width=\linewidth,trim={0 0 0 0},clip]{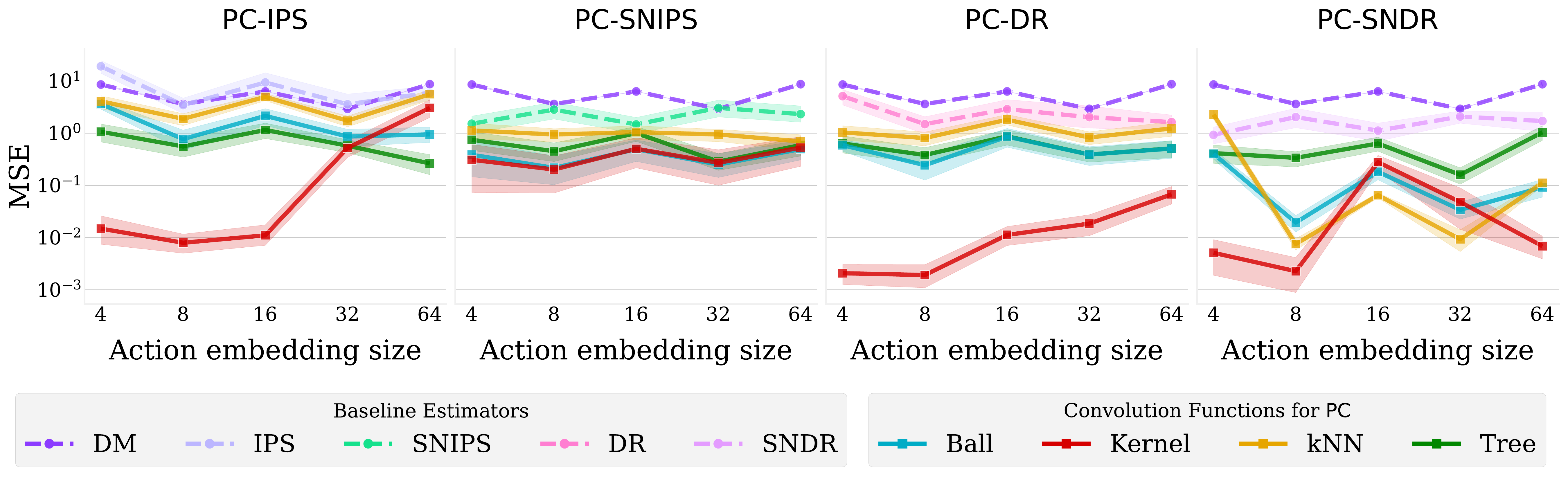}
    \end{subfigure} 
    \caption{Change in MSE, Squared Bias \& Variance while estimating $V(\pi_{\mathsf{good}})$ with varying action embedding size ($\log$-$\log$ scale) for the synthetic dataset ($2000$ actions), using data logged by $\mu_{\mathsf{uniform}}$.}
    \label{fig:v_embed_size_synthetic_pi_good_mu_unif}
\end{minipage}
\end{figure*}

\begin{figure*}
\begin{minipage}[c][\textheight][c]{\textwidth}
    \centering
    \begin{subfigure}{0.49\textwidth}
        \caption{Bias-Variance trade-off for \oracle-\ips}
        \includegraphics[width=\linewidth,trim={23cm 5.5cm 0 0},clip]{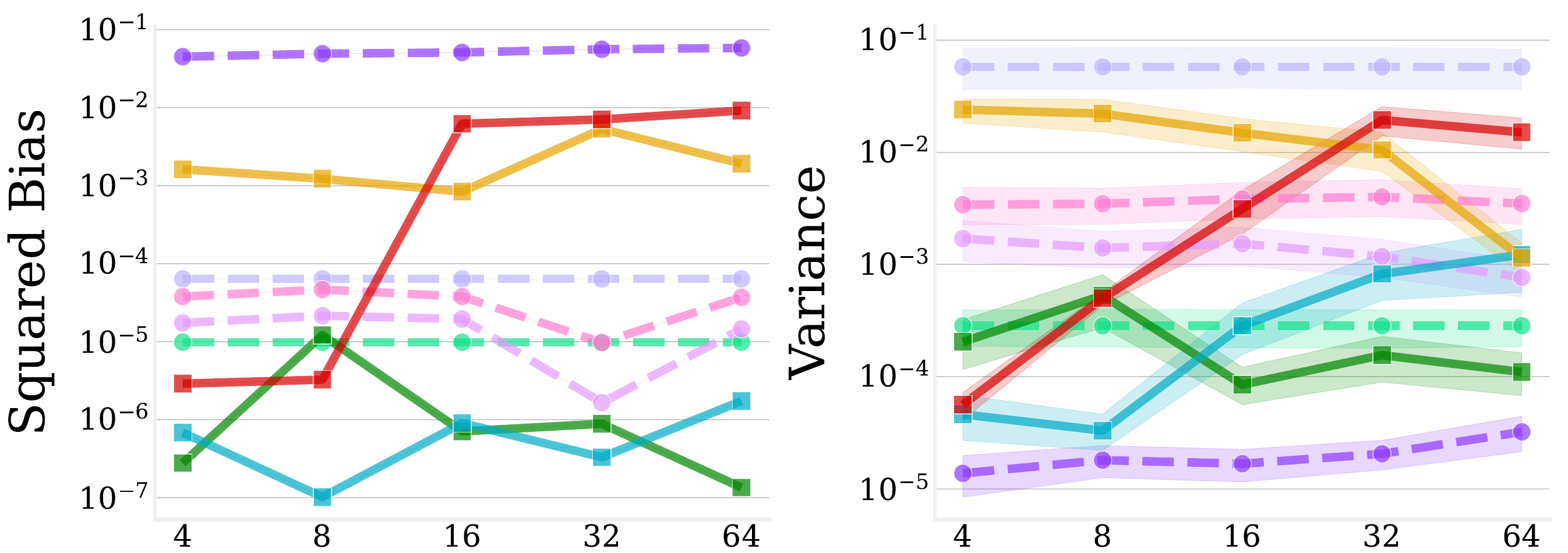}
    \end{subfigure} \hfill %
    \begin{subfigure}{0.49\textwidth}
        \caption{Bias-Variance trade-off for \oracle-\snips}
        \includegraphics[width=\linewidth,trim={23cm 5.5cm 0 0},clip]{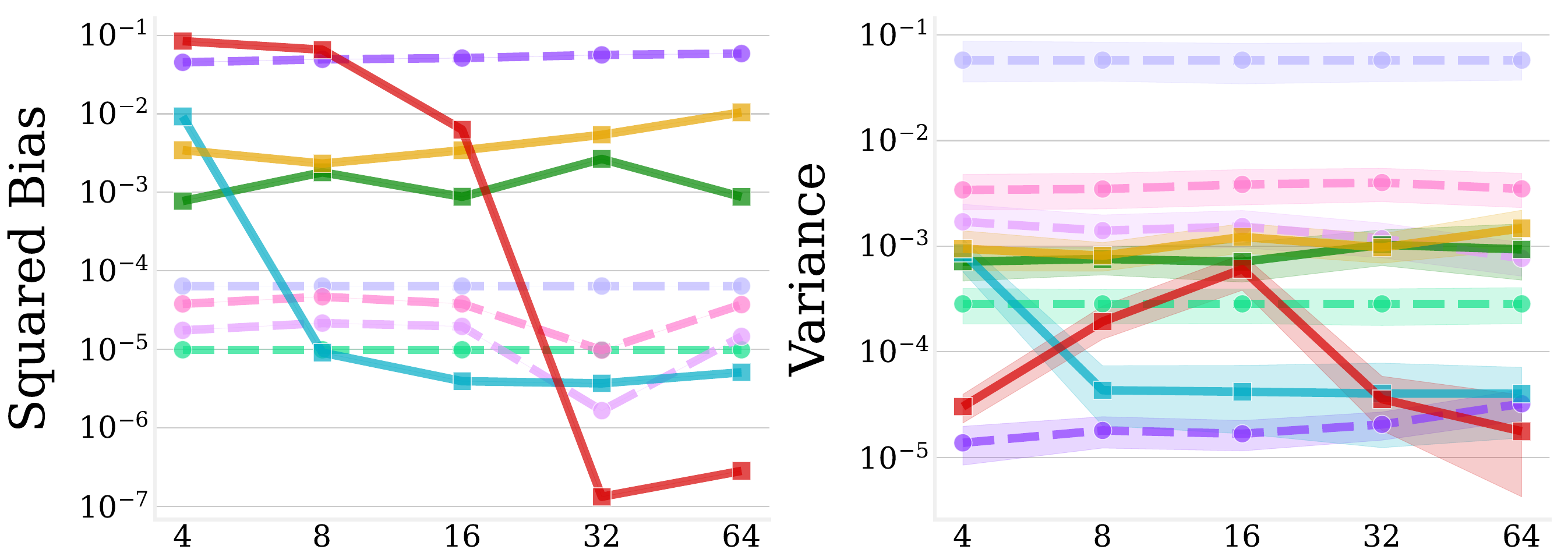}
    \end{subfigure}
    \begin{subfigure}{0.49\textwidth}
        \caption{Bias-Variance trade-off for \oracle-\dr}
        \includegraphics[width=\linewidth,trim={23cm 5.5cm 0 0},clip]{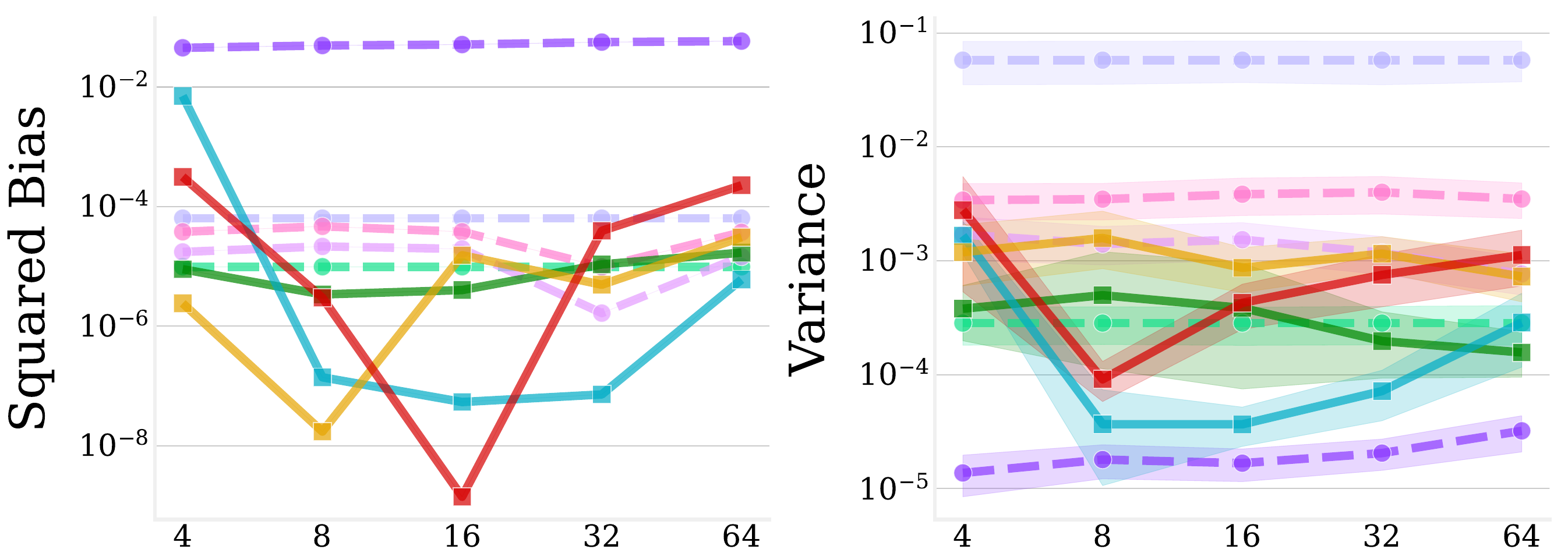}
    \end{subfigure} \hfill %
    \begin{subfigure}{0.49\textwidth}
        \caption{Bias-Variance trade-off for \oracle-\sndr}
        \includegraphics[width=\linewidth,trim={23cm 5.5cm 0 0},clip]{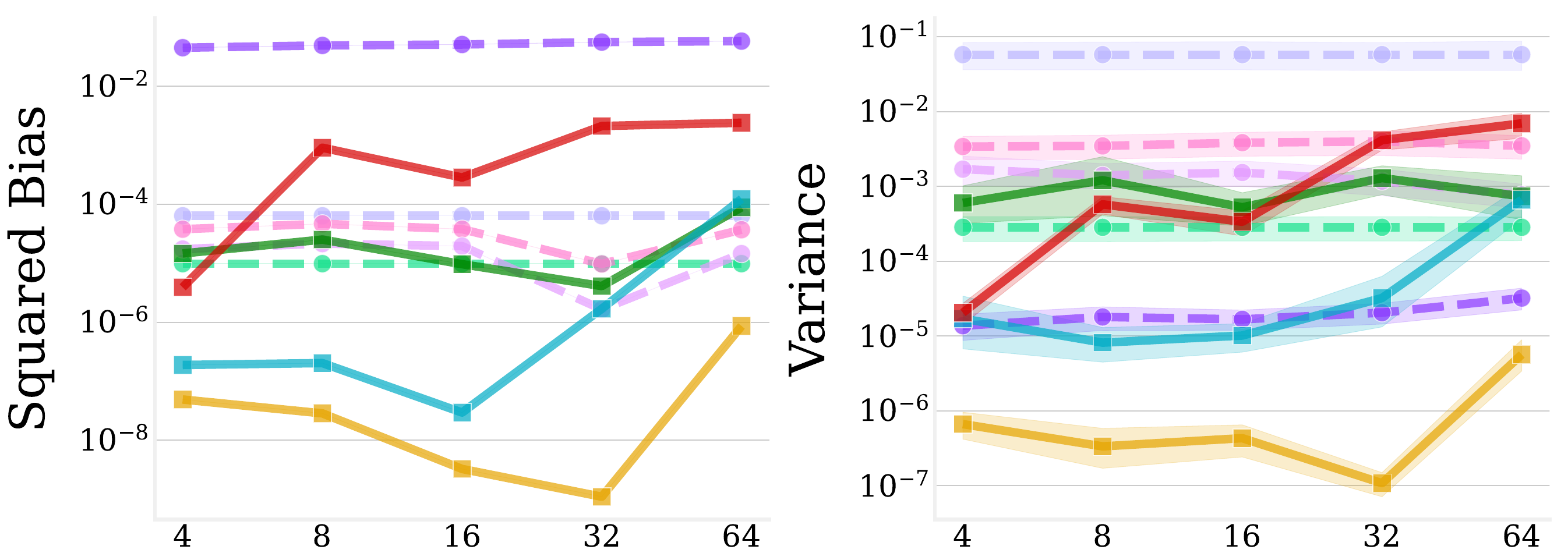}
    \end{subfigure}

    \vspace{0.3cm}
    
    \begin{subfigure}{\textwidth}
        \includegraphics[width=\linewidth,trim={0 0 0 0},clip]{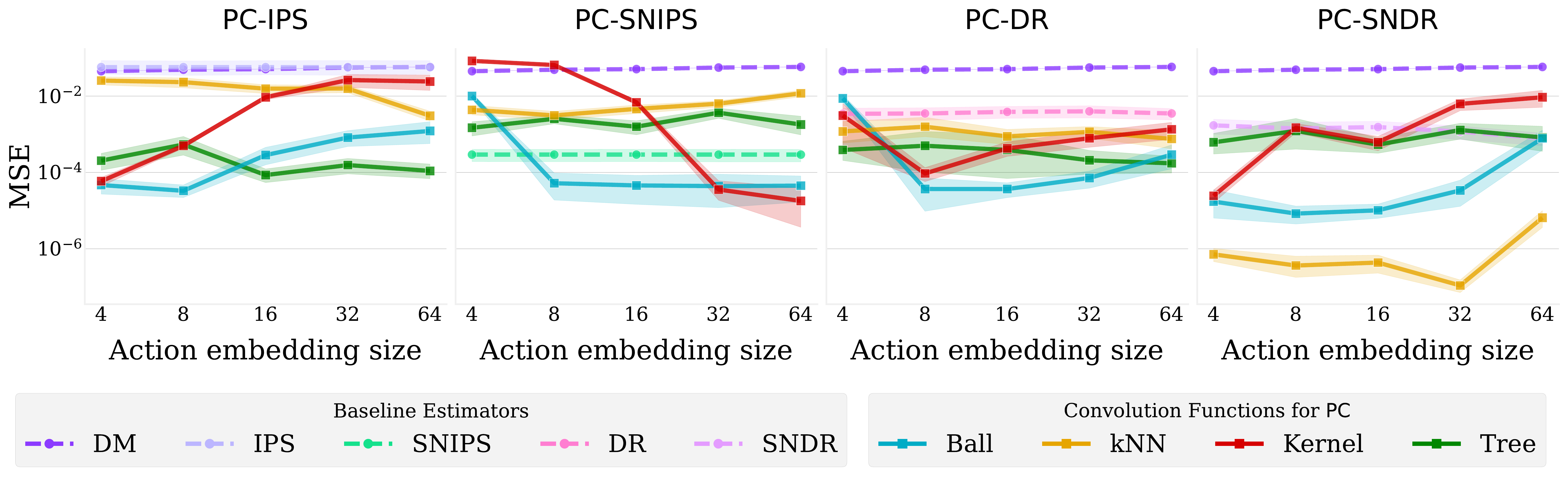}
    \end{subfigure} 
    \caption{Change in MSE, Squared Bias \& Variance while estimating $V(\pi_{\mathsf{good}})$ with varying action embedding size ($\log$-$\log$ scale) for the movielens dataset, using data logged by $\mu_{\mathsf{uniform}}$.}
    \label{fig:v_embed_size_ml_pi_good_mu_unif}
\end{minipage}
\end{figure*}

\clearpage

\begin{figure*}
\begin{minipage}[c][\textheight][c]{\textwidth}
    \centering
    \includegraphics[width=\linewidth,trim={0cm 0 0 0},clip]{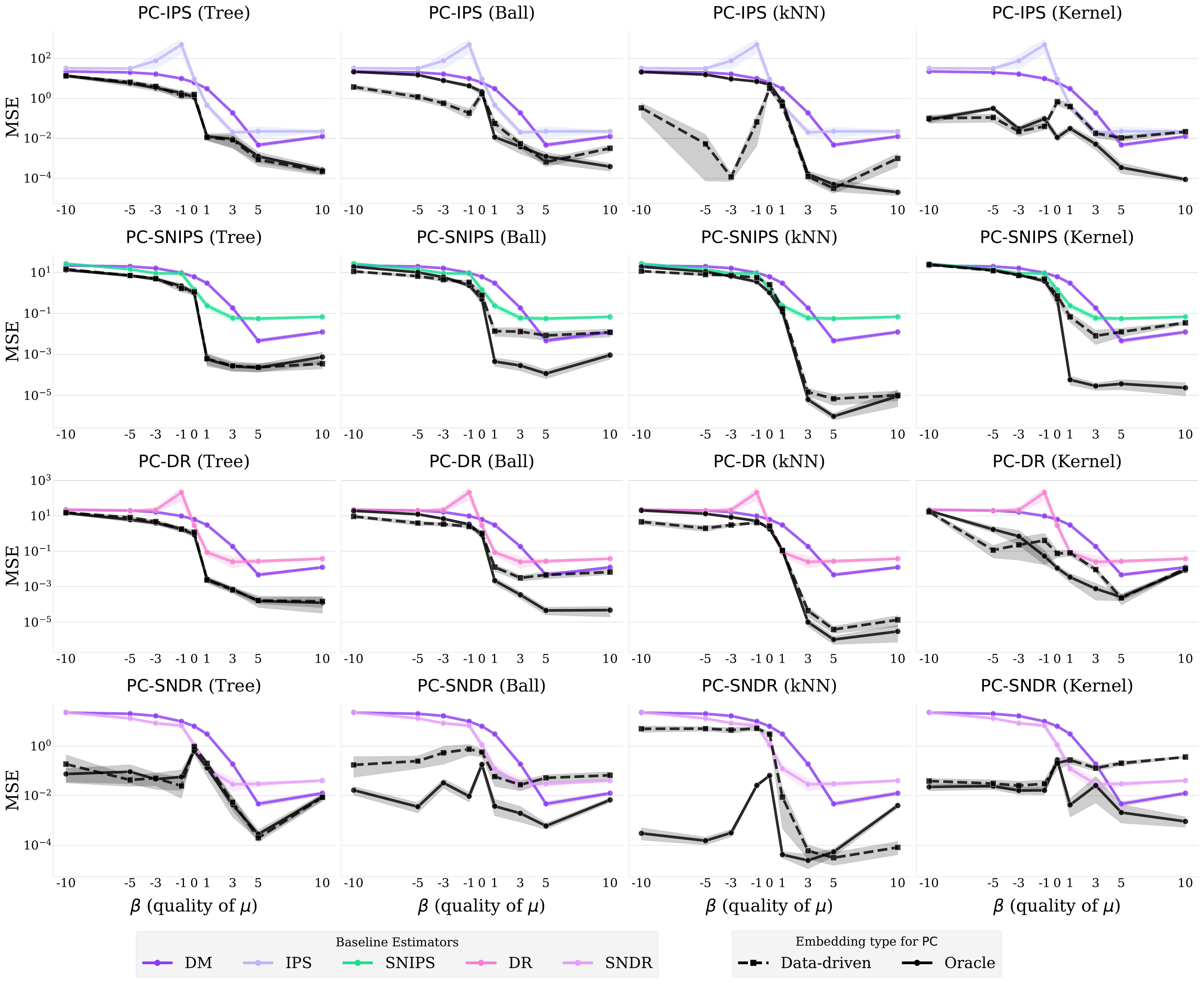}
    \caption{Change in MSE while estimating $V(\pi_{\mathsf{good}})$ with varying logging policies ($\log$-scale) and using \oracle with \texttt{Oracle} \vs \texttt{Data-driven} action embeddings for the synthetic dataset.}
    \label{fig:v_embed_type_pi_good_mu_unif}
\end{minipage}
\end{figure*}

\begin{figure*}
\begin{minipage}[c][\textheight][c]{\textwidth}
    \centering
    \includegraphics[width=\linewidth,trim={0cm 0 0 0},clip]{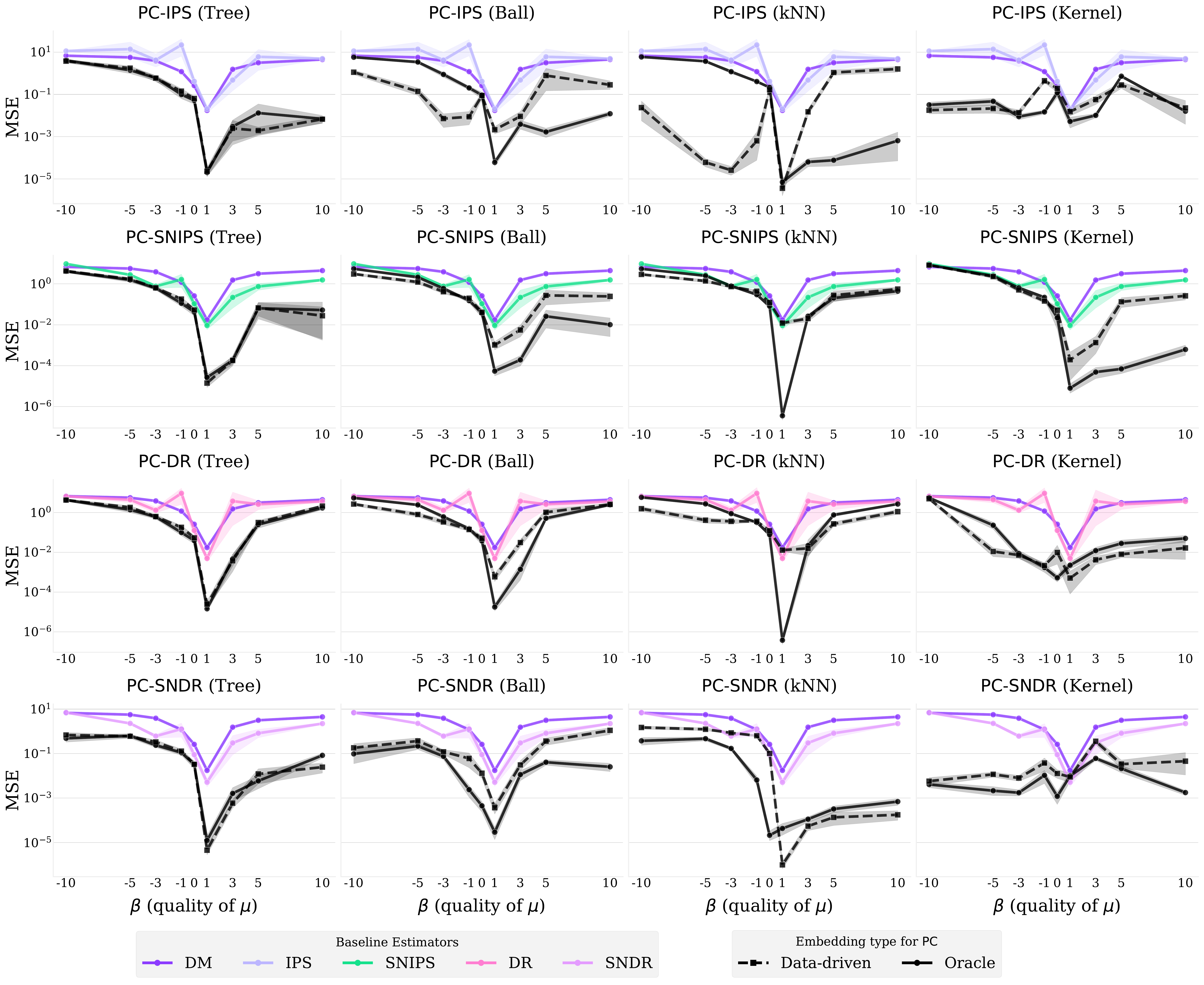}
    \caption{Change in MSE while estimating $V(\pi_{\mathsf{bad}})$ with varying logging policies ($\log$-scale) and using \oracle with \texttt{Oracle} \vs \texttt{Data-driven} action embeddings for the synthetic dataset.}
    \label{fig:v_embed_type_pi_bad_mu_unif}
\end{minipage}
\end{figure*}

\clearpage

\begin{figure*}
\begin{minipage}[c][\textheight][c]{\textwidth}
    \centering
    \begin{subfigure}{0.8\textwidth}
        \caption{Bias-Variance trade-off for \oracle-\ips}
        \vspace{0.1cm}
        \includegraphics[width=0.495\linewidth,trim={0cm 3.4cm 0 0},clip]{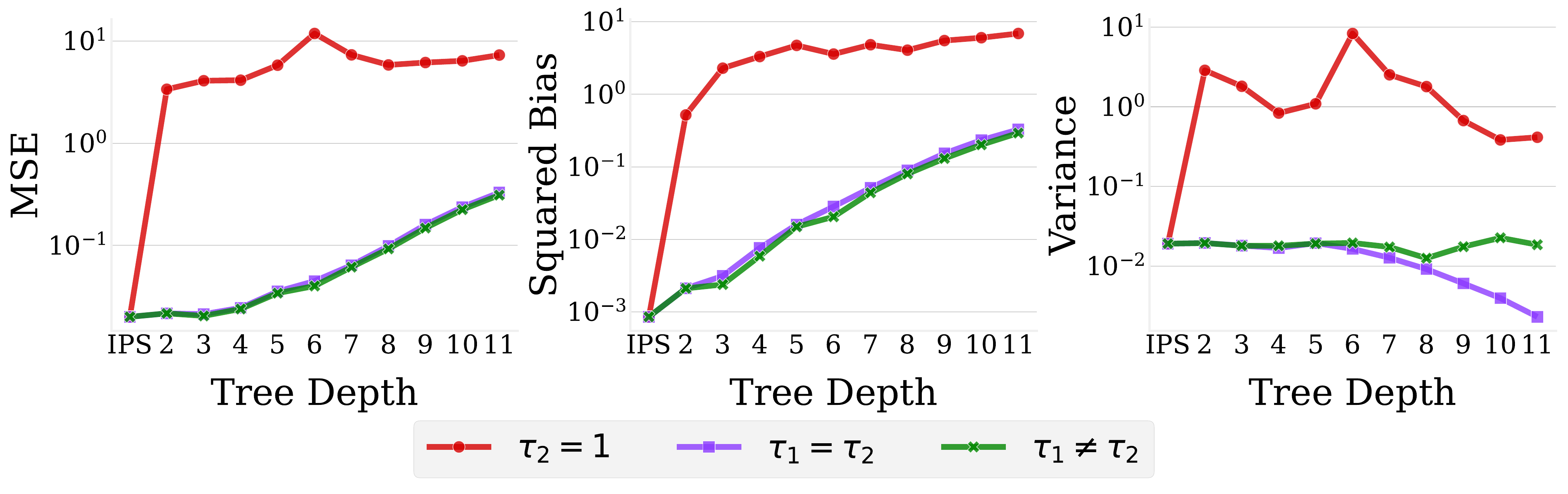} \hfill %
        \includegraphics[width=0.495\linewidth,trim={0cm 3.4cm 0 0},clip]{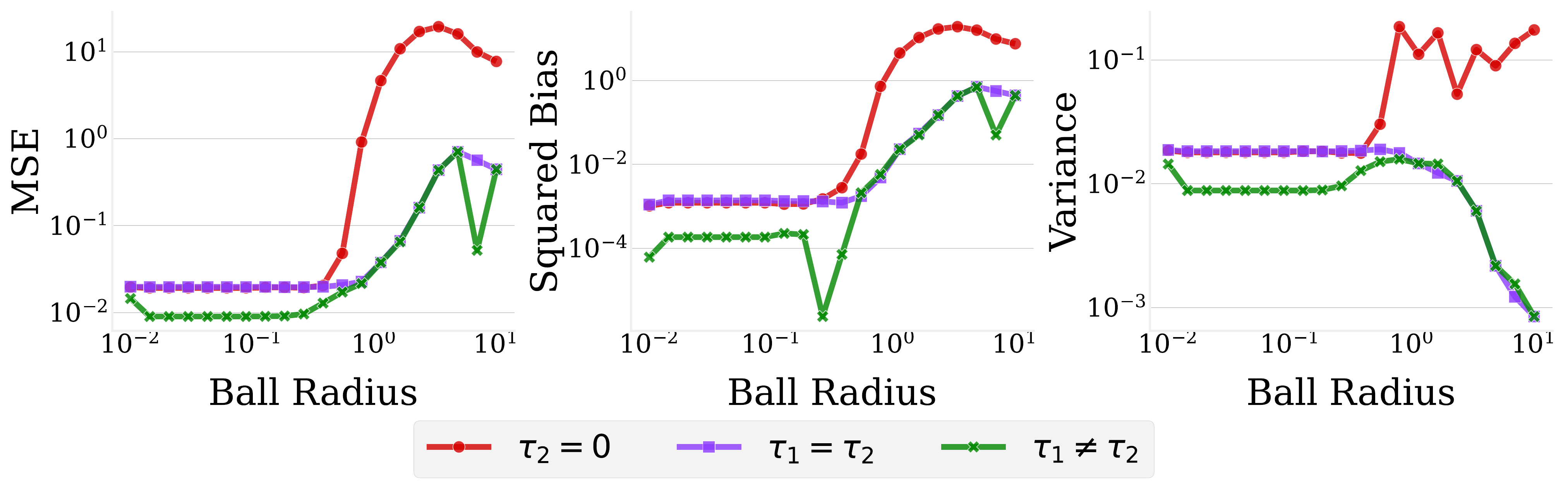}
        \includegraphics[width=0.495\linewidth,trim={0cm 3.4cm 0 0},clip]{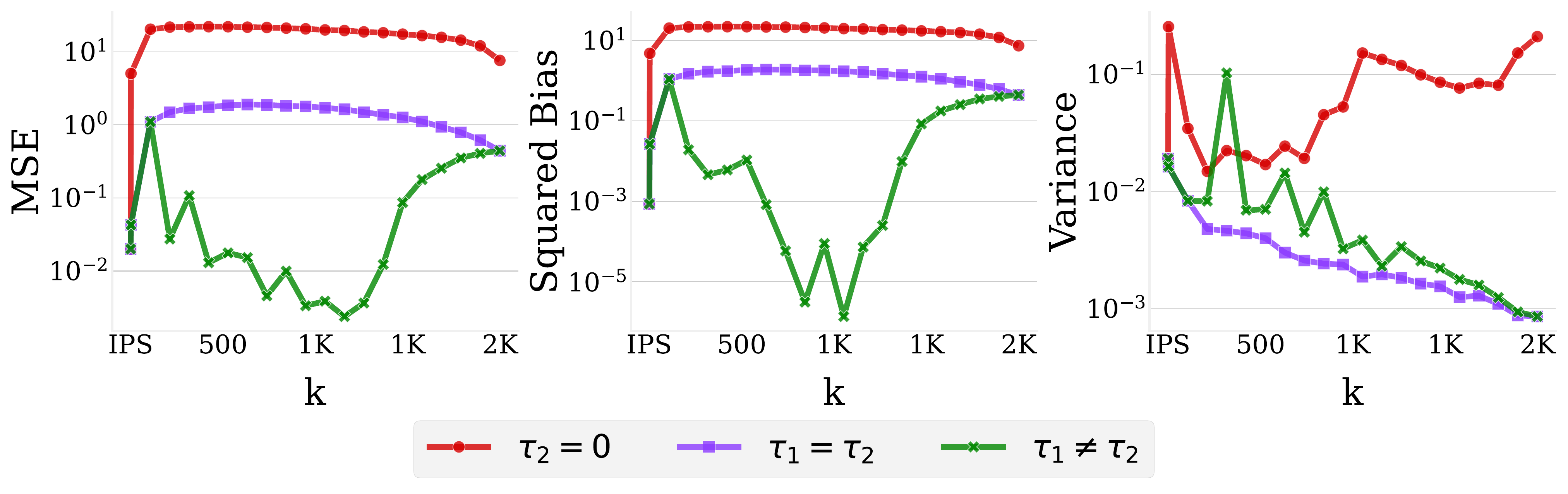} \hfill %
        \includegraphics[width=0.495\linewidth,trim={0cm 3.4cm 0 0},clip]{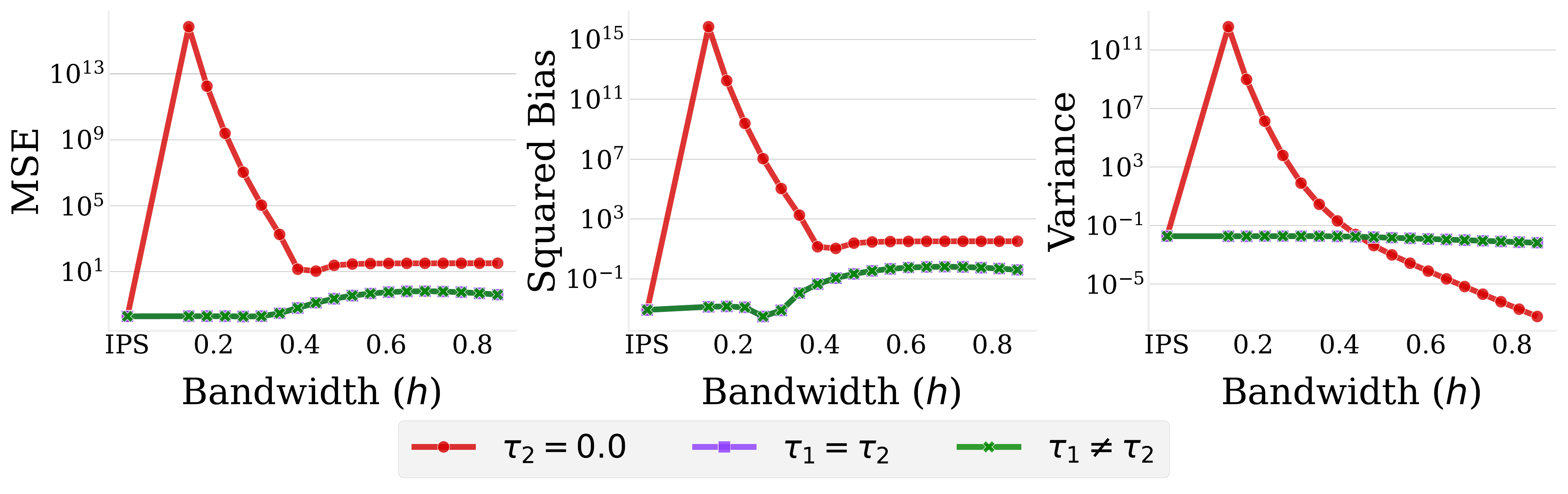}
        \includegraphics[width=\linewidth,trim={0 0 0 18cm},clip]{figures/bias_variance/synthetic/2k_actions_beta_3_eps_0.05/IPS/kNN_num.pdf}
    \end{subfigure}
    \begin{subfigure}{0.8\textwidth}
        \caption{Bias-Variance trade-off for \oracle-\snips}
        \vspace{0.1cm}
        \includegraphics[width=0.495\linewidth,trim={0cm 3.4cm 0 0},clip]{figures/bias_variance/synthetic/2k_actions_beta_3_eps_0.05/SNIPS/Tree_num.pdf} \hfill %
        \includegraphics[width=0.495\linewidth,trim={0cm 3.4cm 0 0},clip]{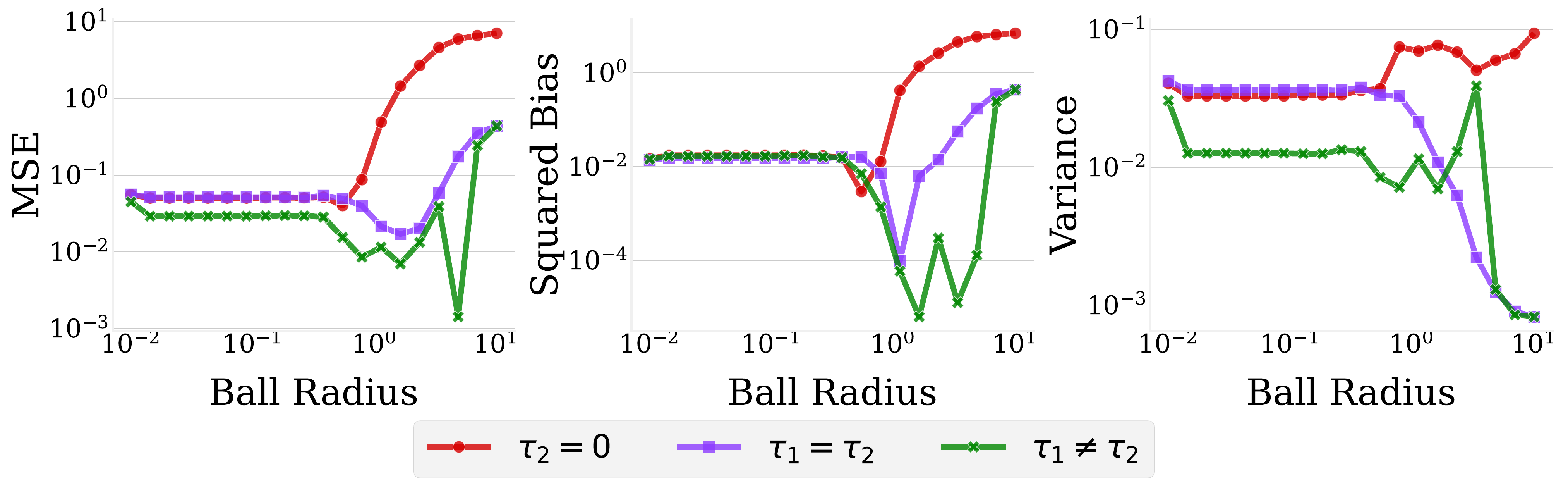}
        \includegraphics[width=0.495\linewidth,trim={0cm 3.4cm 0 0},clip]{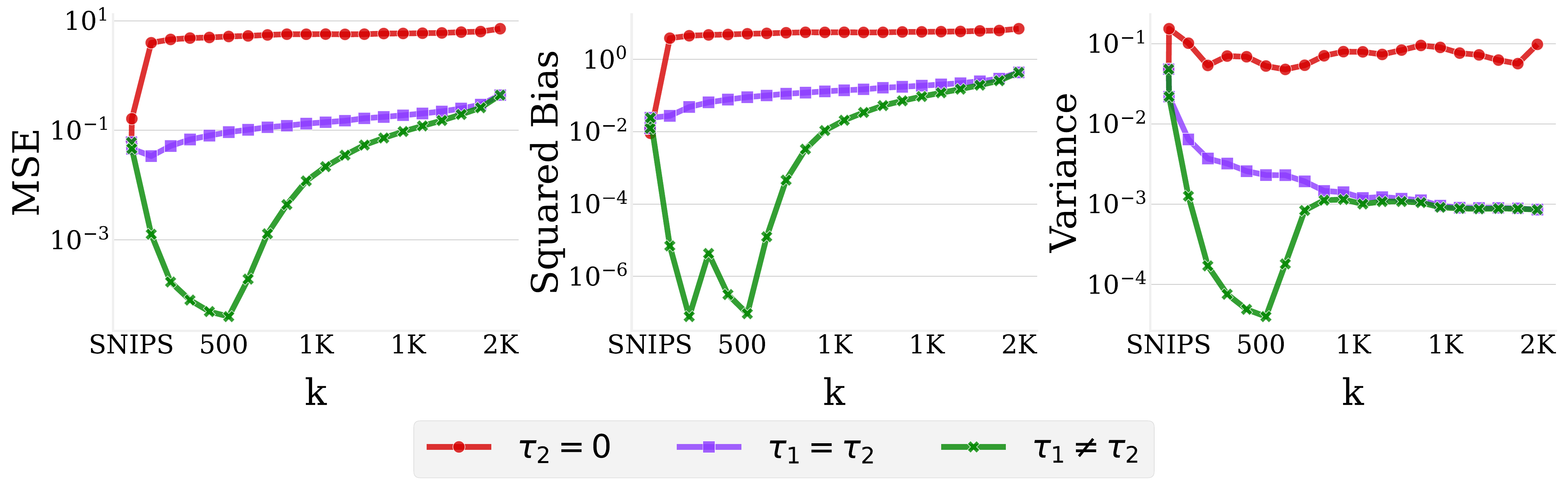} \hfill %
        \includegraphics[width=0.495\linewidth,trim={0cm 3.4cm 0 0},clip]{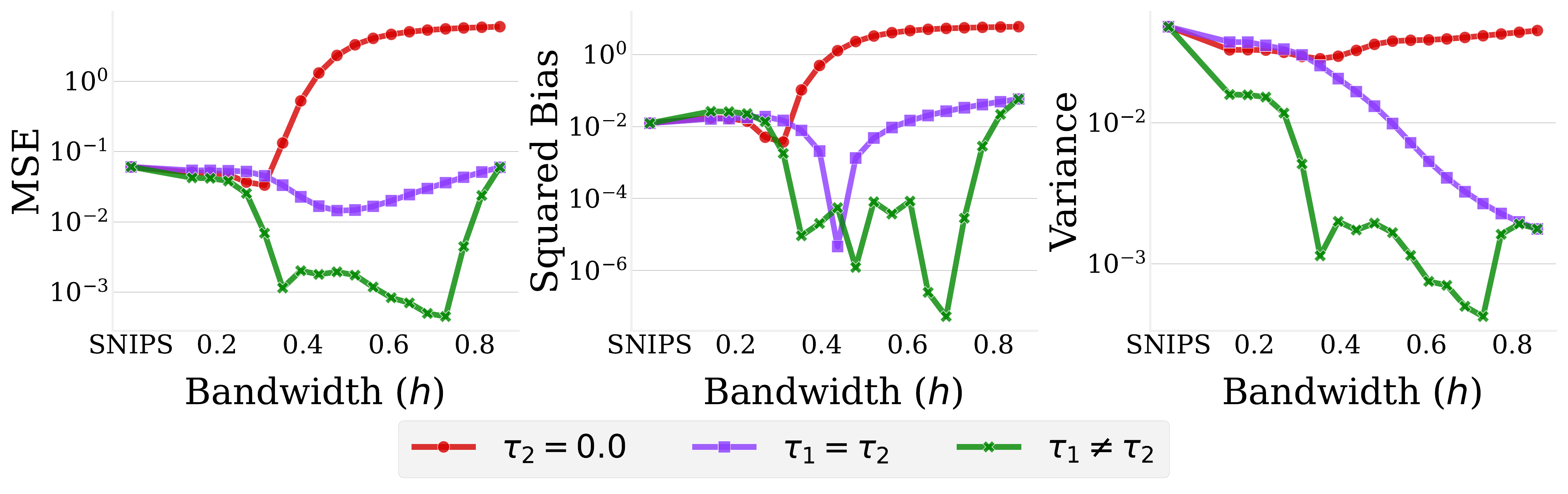}
        \includegraphics[width=\linewidth,trim={0 0 0 18cm},clip]{figures/bias_variance/synthetic/2k_actions_beta_3_eps_0.05/IPS/kNN_num.pdf}
    \end{subfigure}
    \begin{subfigure}{0.8\textwidth}
        \caption{Bias-Variance trade-off for \oracle-\dr}
        \vspace{0.1cm}
        \includegraphics[width=0.495\linewidth,trim={0cm 3.4cm 0 0},clip]{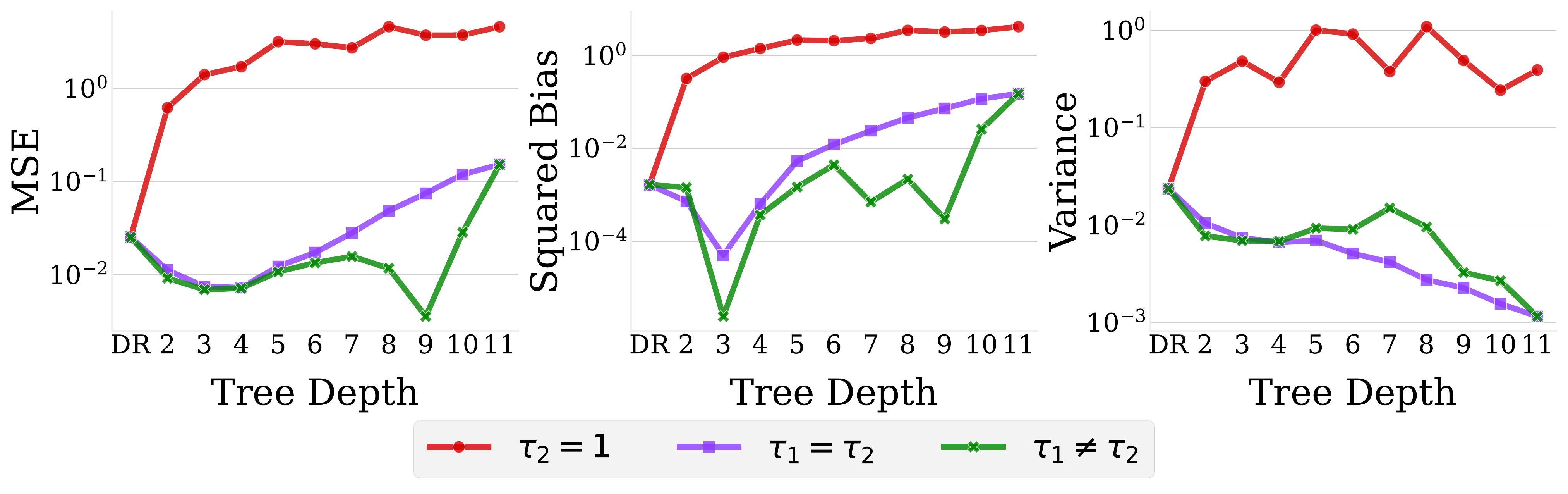} \hfill %
        \includegraphics[width=0.495\linewidth,trim={0cm 3.4cm 0 0},clip]{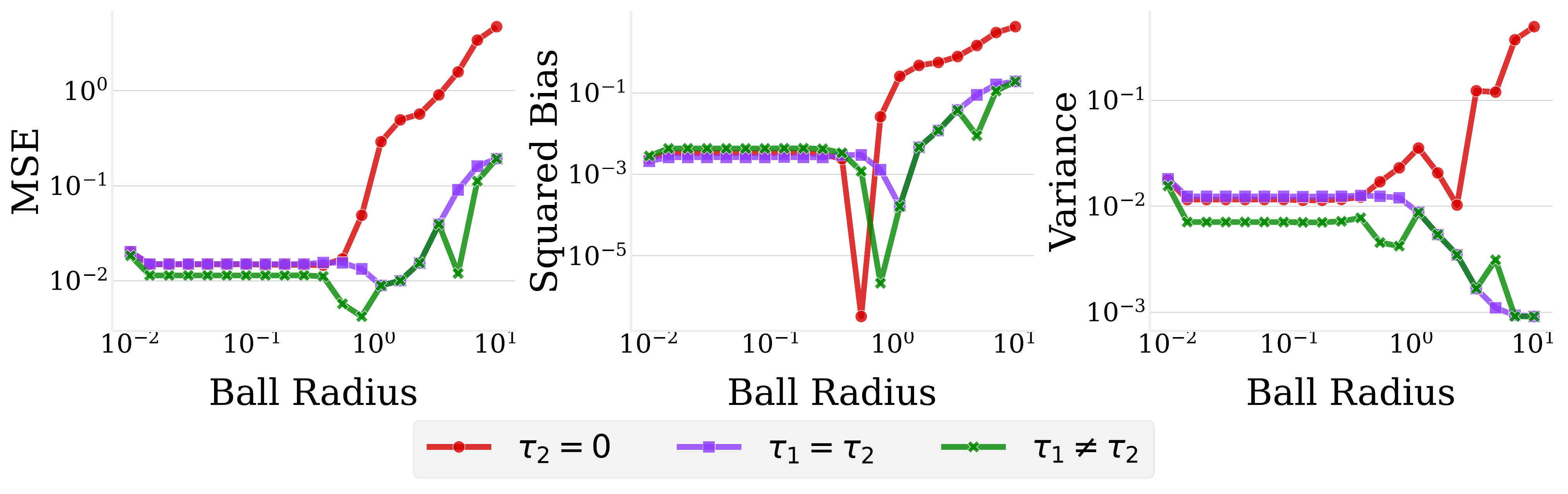}
        \includegraphics[width=0.495\linewidth,trim={0cm 3.4cm 0 0},clip]{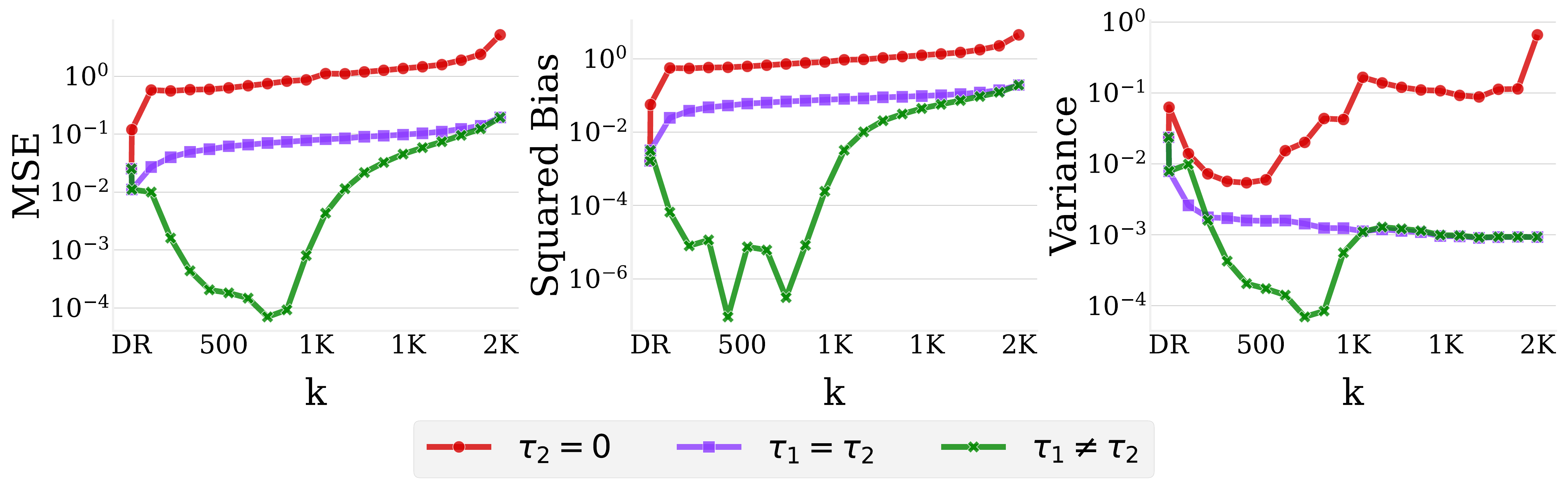} \hfill %
        \includegraphics[width=0.495\linewidth,trim={0cm 3.4cm 0 0},clip]{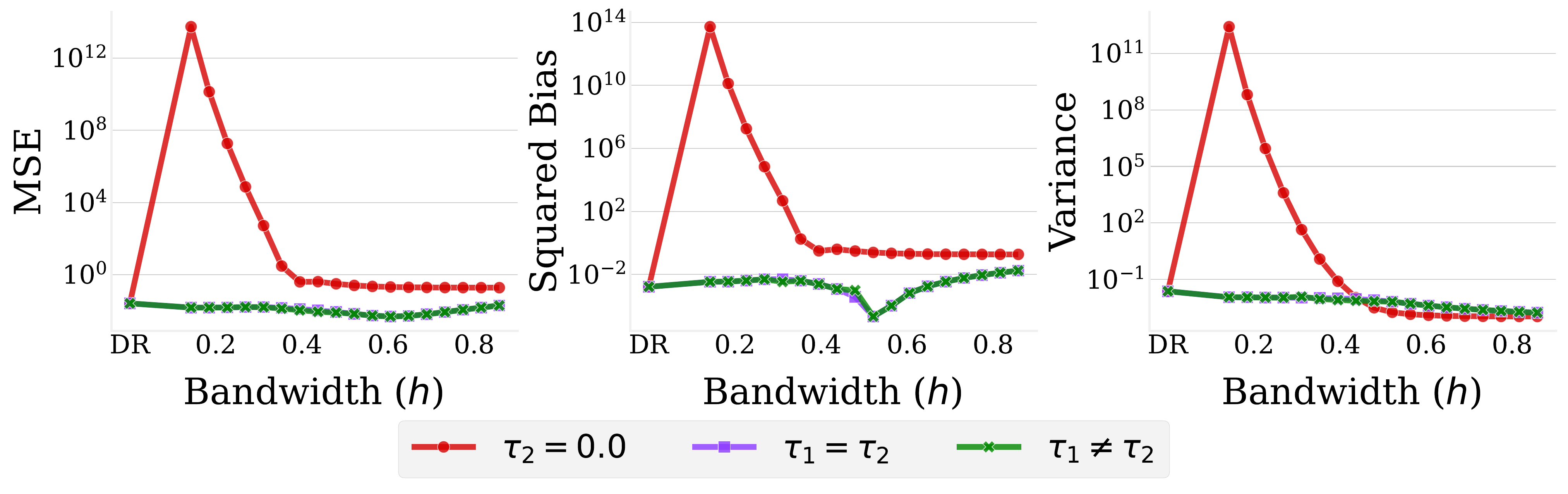}
        \includegraphics[width=\linewidth,trim={0 0 0 18cm},clip]{figures/bias_variance/synthetic/2k_actions_beta_3_eps_0.05/IPS/kNN_num.pdf}
    \end{subfigure}
    \begin{subfigure}{0.8\textwidth}
        \caption{Bias-Variance trade-off for \oracle-\sndr}
        \vspace{0.1cm}
        \includegraphics[width=0.495\linewidth,trim={0cm 3.4cm 0 0},clip]{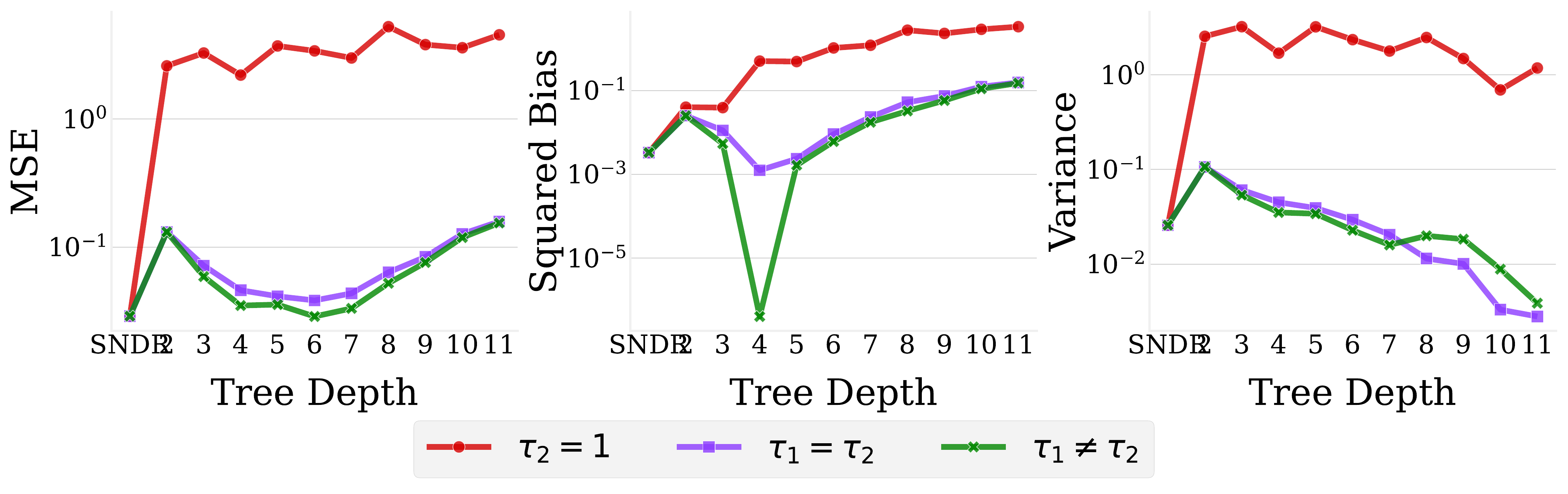} \hfill %
        \includegraphics[width=0.495\linewidth,trim={0cm 3.4cm 0 0},clip]{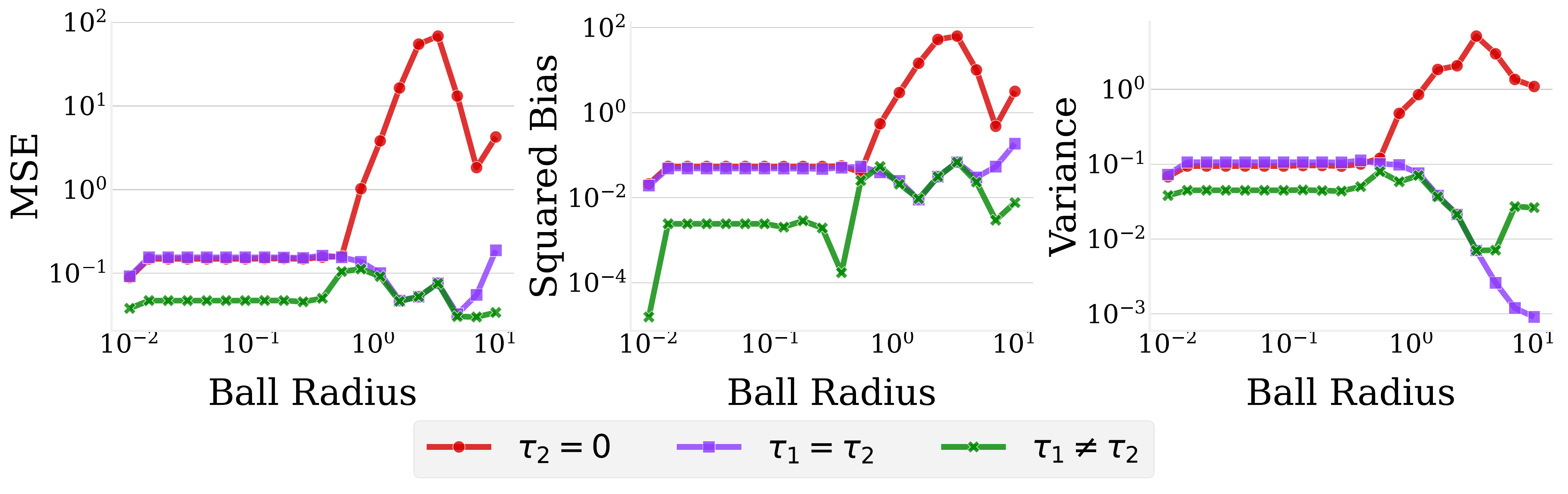}
        \includegraphics[width=0.495\linewidth,trim={0cm 3.4cm 0 0},clip]{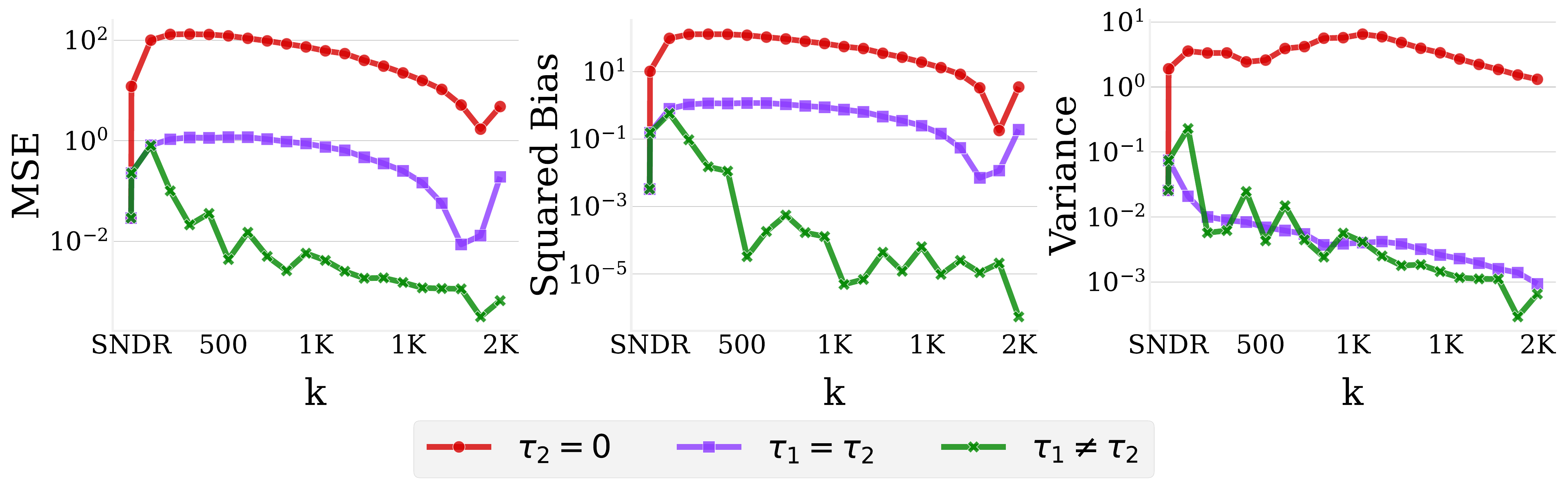} \hfill %
        \includegraphics[width=0.495\linewidth,trim={0cm 3.4cm 0 0},clip]{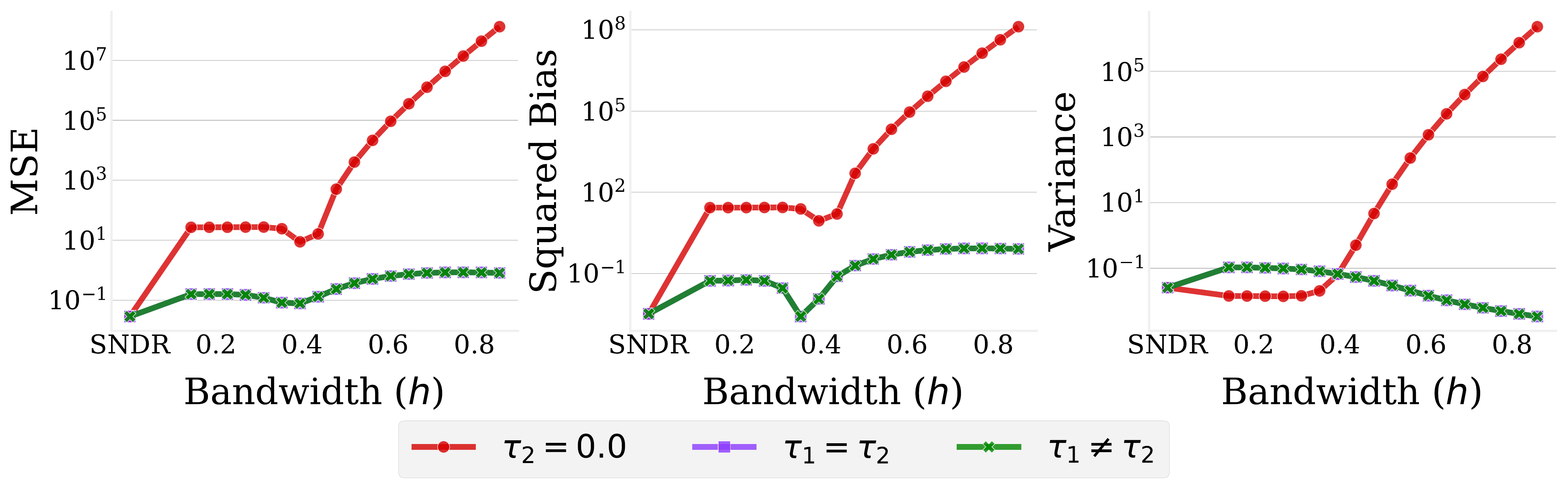}
        \includegraphics[width=\linewidth,trim={0 0 0 18cm},clip]{figures/bias_variance/synthetic/2k_actions_beta_3_eps_0.05/IPS/kNN_num.pdf}
    \end{subfigure}
    \vspace{-10pt}
    \caption{Visualizing the bias-variance trade-off for the \oracle estimator for different backbones \& pooling strategies, while estimating $V(\pi_{\mathsf{good}})$ with varying amount of pooling ($\log$-$\log$ scale) on the synthetic dataset (with $2000$ actions), using $\mu_{\mathsf{good}}$ for logging. Note that the respective na\"ive backbone estimators are the left-most point in each plot, \ie, when there's \textbf{no} pooling.}
    \label{fig:v_pooling_synthetic_pi_good_mu_good}
\end{minipage}
\end{figure*}

\begin{figure*}
\begin{minipage}[c][\textheight][c]{\textwidth}
    \centering
    \begin{subfigure}{0.8\textwidth}
        \caption{Bias-Variance trade-off for \oracle-\ips}
        \vspace{0.1cm}
        \includegraphics[width=0.495\linewidth,trim={0cm 3.4cm 0 0},clip]{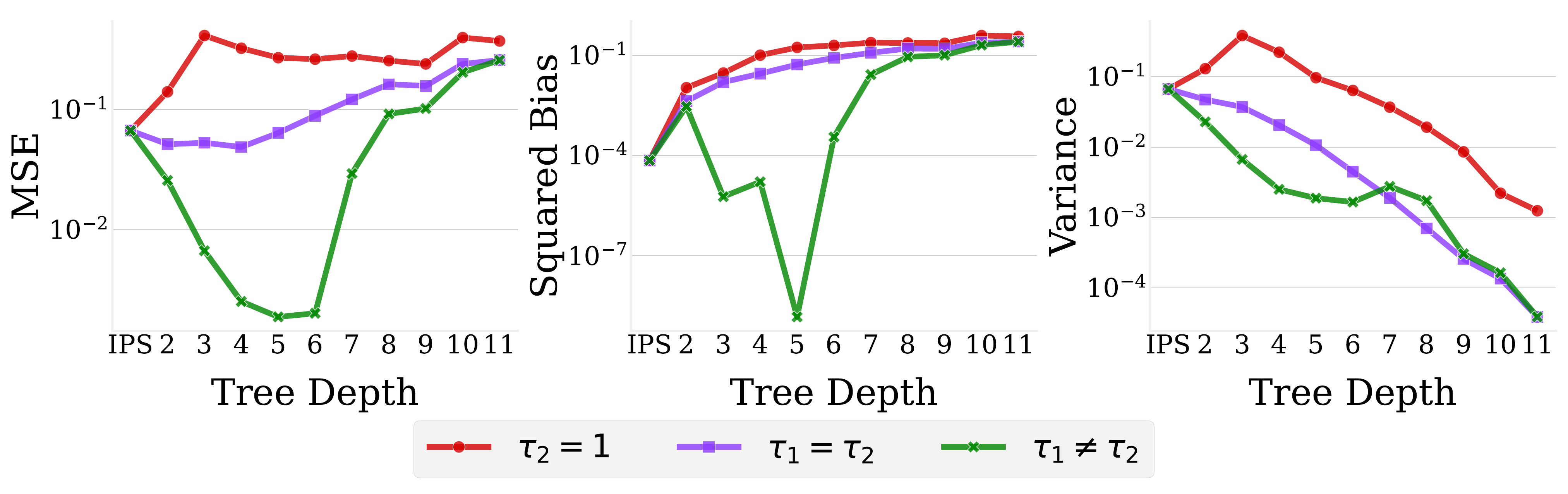} \hfill %
        \includegraphics[width=0.495\linewidth,trim={0cm 3.4cm 0 0},clip]{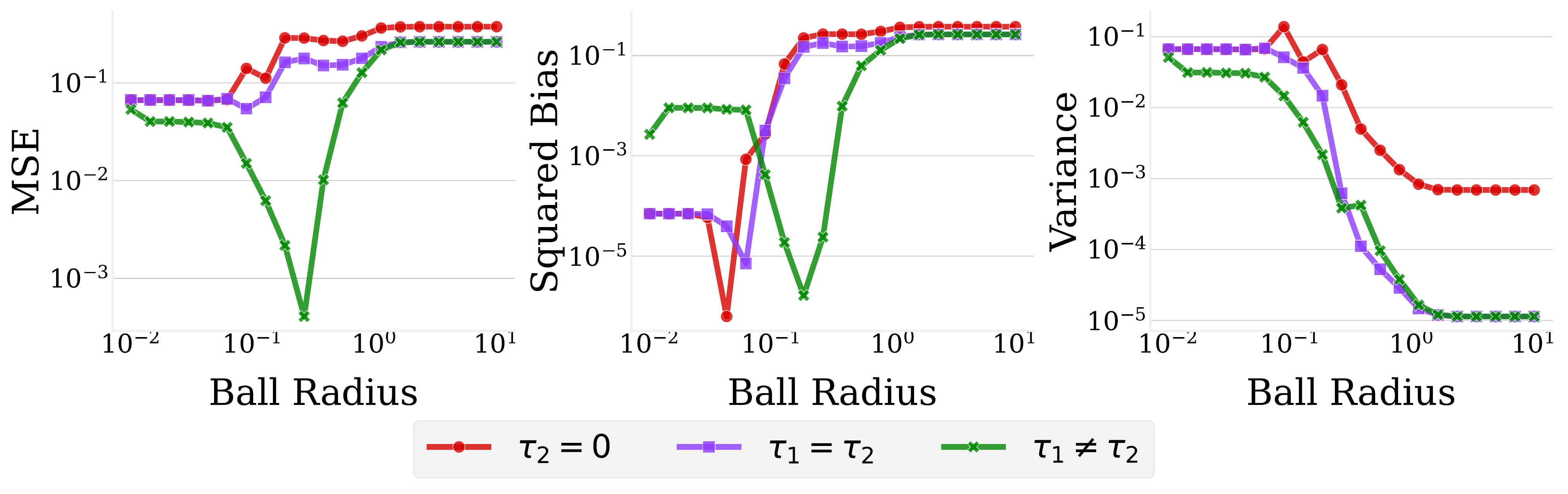}
        \includegraphics[width=0.495\linewidth,trim={0cm 3.4cm 0 0},clip]{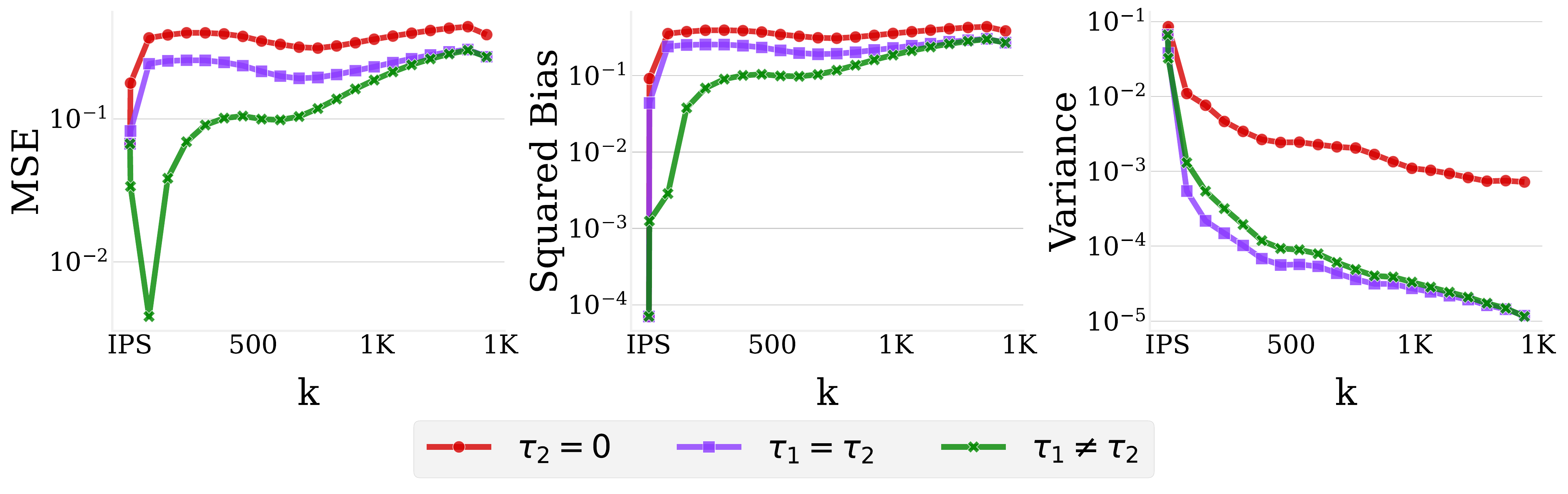} \hfill %
        \includegraphics[width=0.495\linewidth,trim={0cm 3.4cm 0 0},clip]{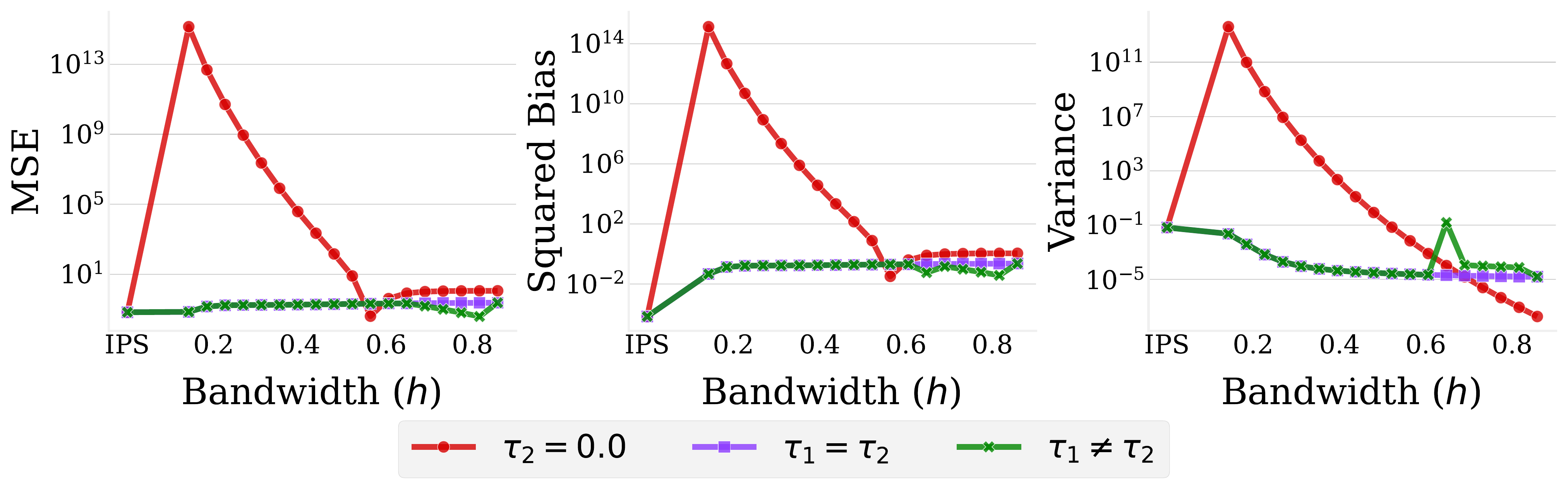}
        \includegraphics[width=\linewidth,trim={0 0 0 18cm},clip]{figures/bias_variance/synthetic/2k_actions_beta_3_eps_0.05/IPS/kNN_num.pdf}
    \end{subfigure}
    \begin{subfigure}{0.8\textwidth}
        \caption{Bias-Variance trade-off for \oracle-\snips}
        \vspace{0.1cm}
        \includegraphics[width=0.495\linewidth,trim={0cm 3.4cm 0 0},clip]{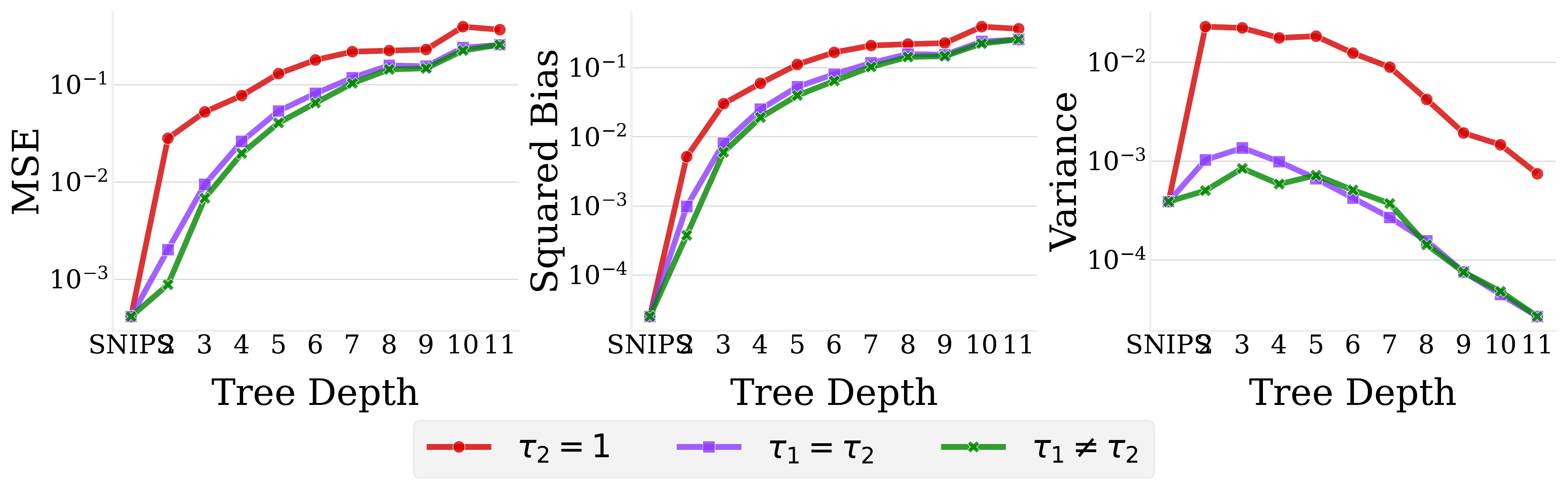} \hfill %
        \includegraphics[width=0.495\linewidth,trim={0cm 3.4cm 0 0},clip]{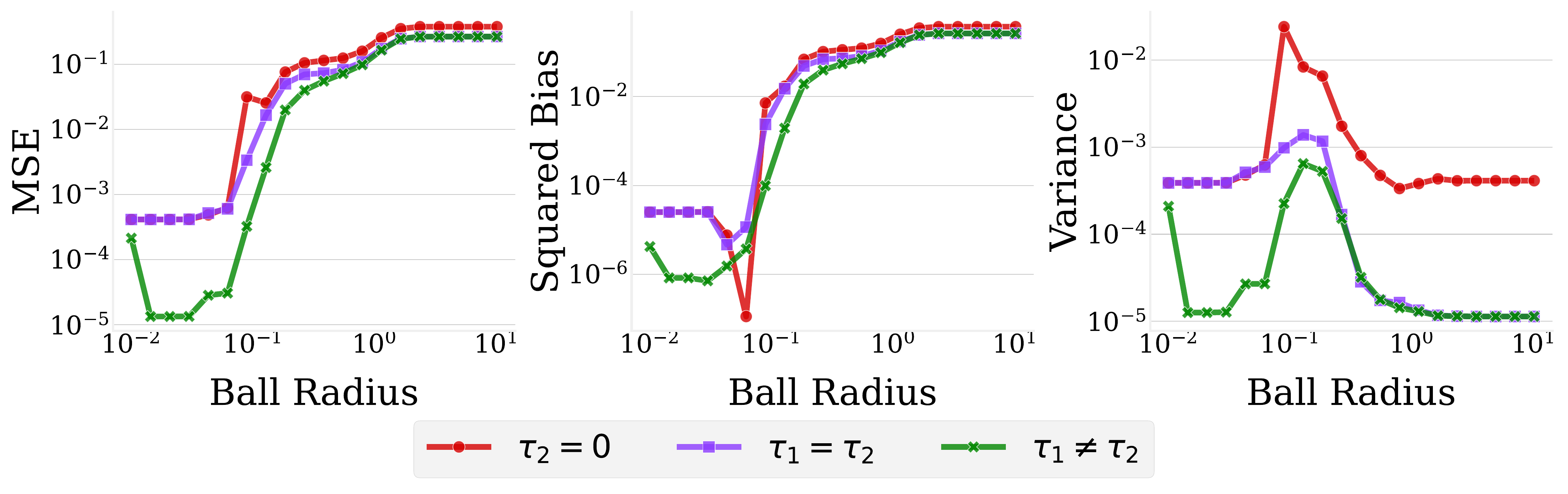}
        \includegraphics[width=0.495\linewidth,trim={0cm 3.4cm 0 0},clip]{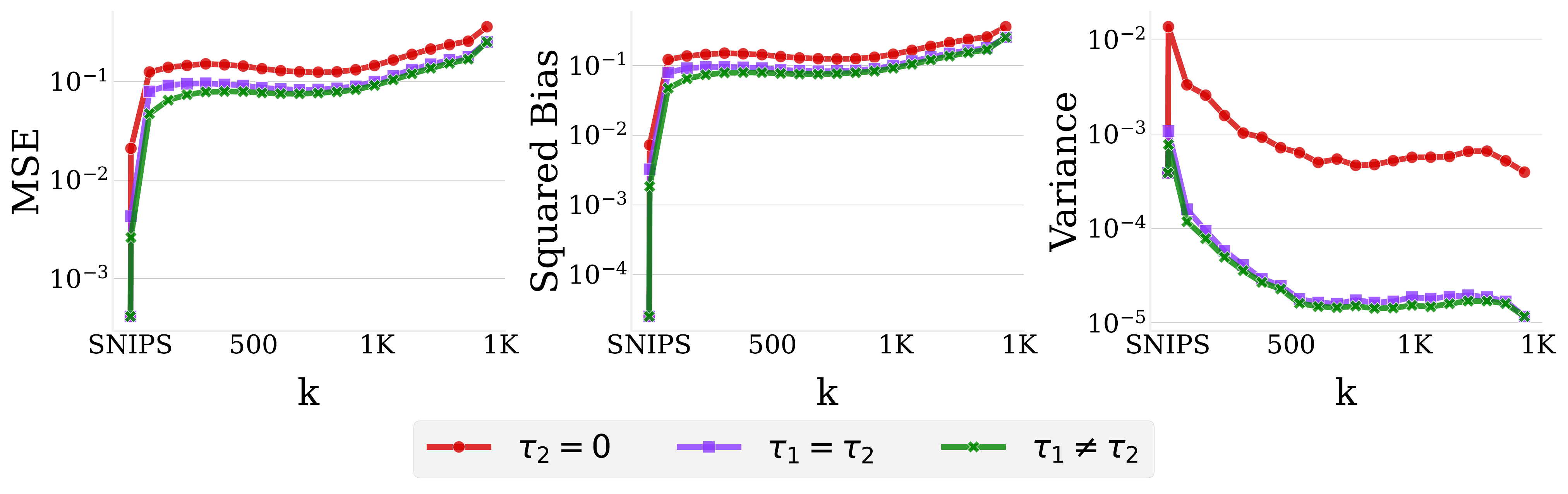} \hfill %
        \includegraphics[width=0.495\linewidth,trim={0cm 3.4cm 0 0},clip]{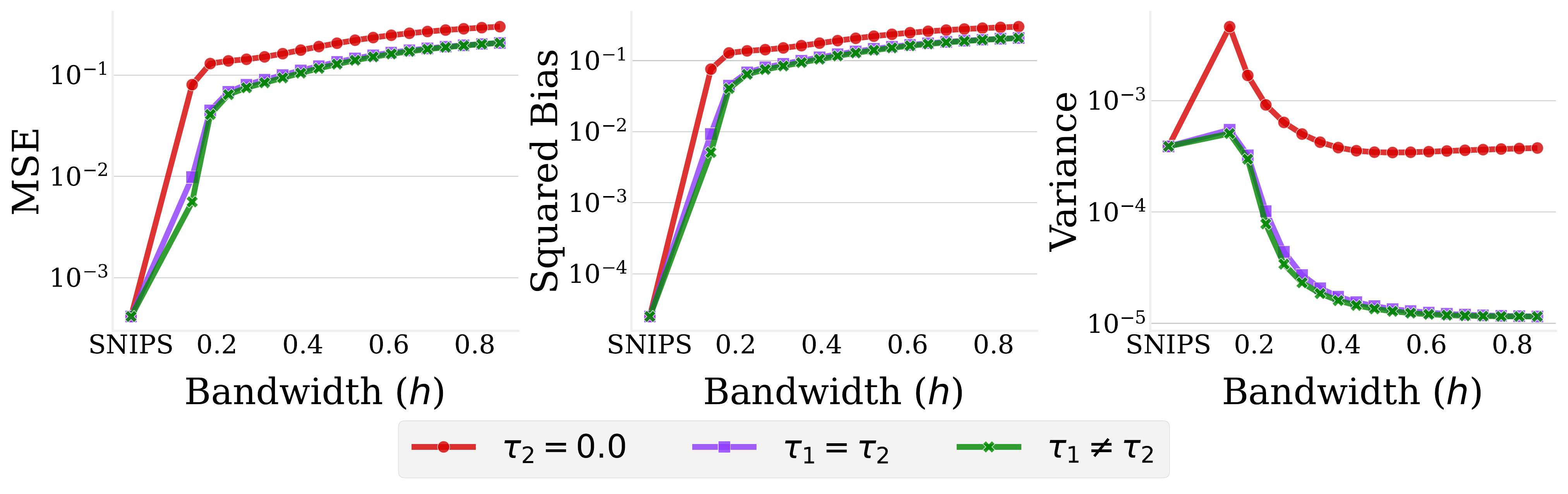}
        \includegraphics[width=\linewidth,trim={0 0 0 18cm},clip]{figures/bias_variance/synthetic/2k_actions_beta_3_eps_0.05/IPS/kNN_num.pdf}
    \end{subfigure}
    \begin{subfigure}{0.8\textwidth}
        \caption{Bias-Variance trade-off for \oracle-\dr}
        \vspace{0.1cm}
        \includegraphics[width=0.495\linewidth,trim={0cm 3.4cm 0 0},clip]{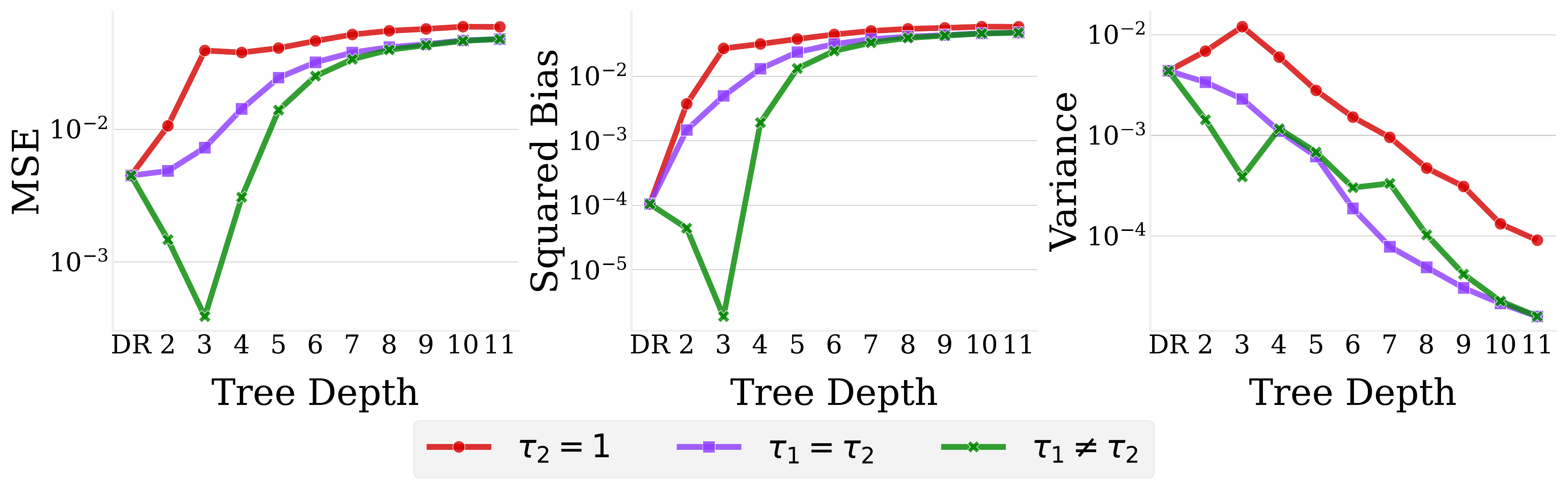} \hfill %
        \includegraphics[width=0.495\linewidth,trim={0cm 3.4cm 0 0},clip]{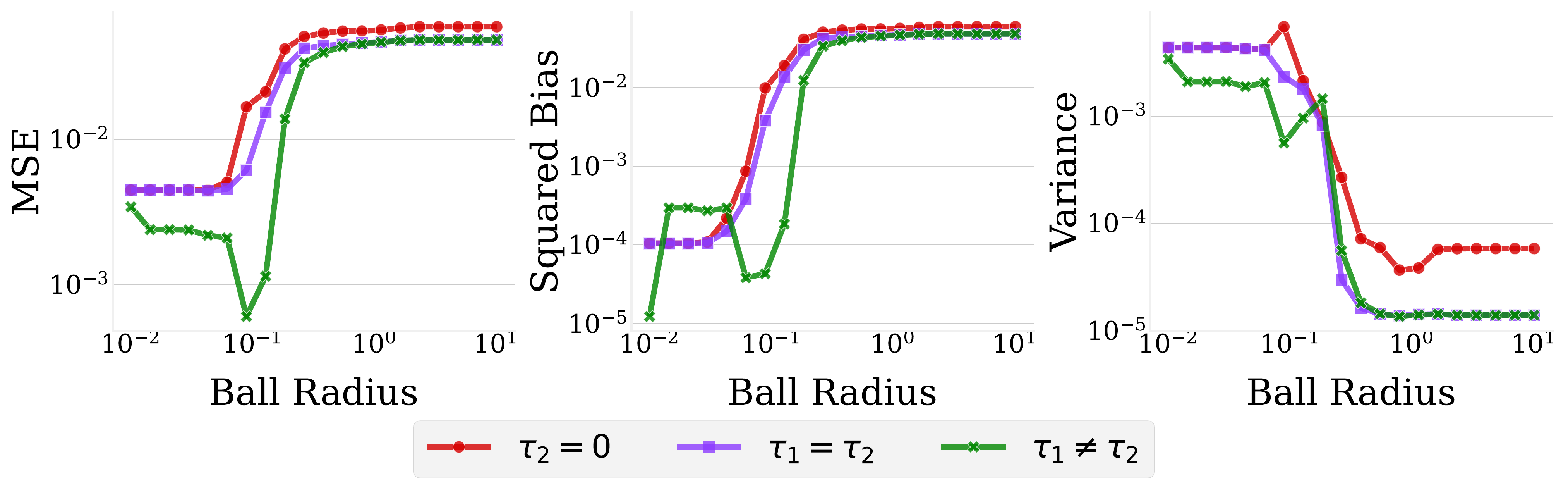}
        \includegraphics[width=0.495\linewidth,trim={0cm 3.4cm 0 0},clip]{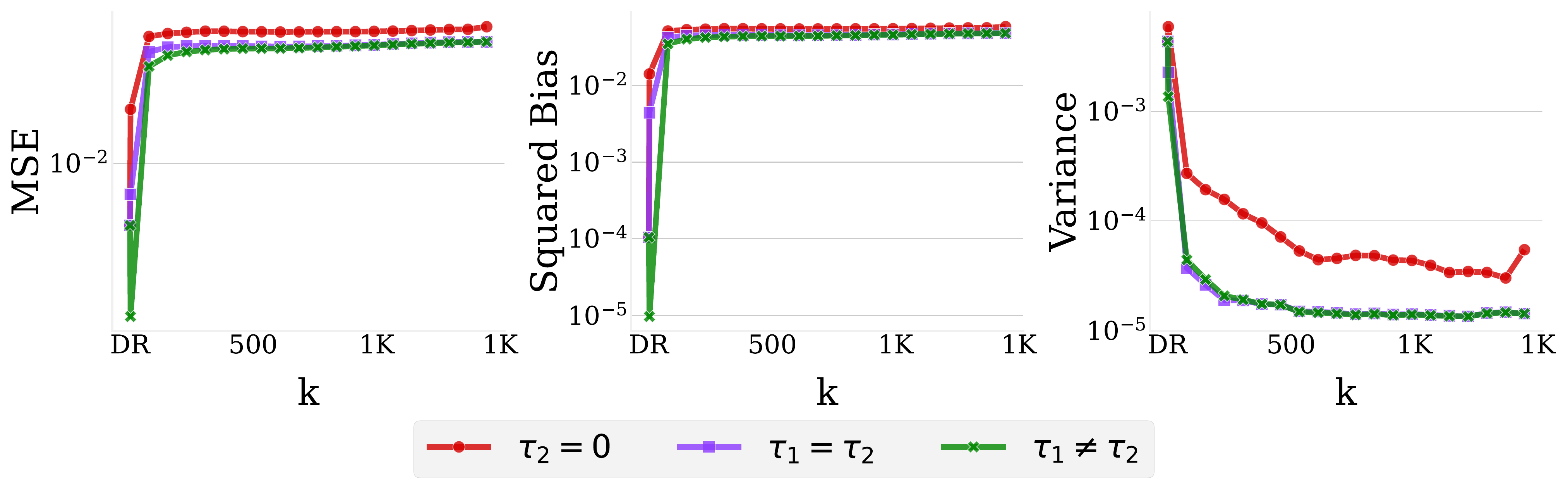} \hfill %
        \includegraphics[width=0.495\linewidth,trim={0cm 3.4cm 0 0},clip]{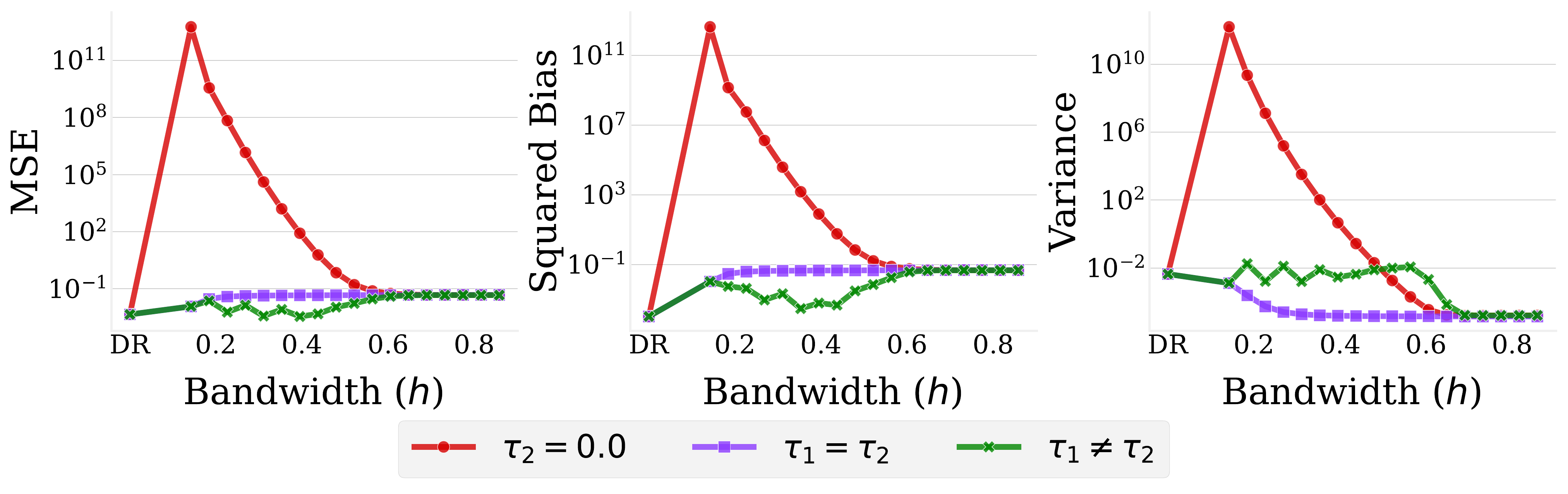}
        \includegraphics[width=\linewidth,trim={0 0 0 18cm},clip]{figures/bias_variance/synthetic/2k_actions_beta_3_eps_0.05/IPS/kNN_num.pdf}
    \end{subfigure}
    \begin{subfigure}{0.8\textwidth}
        \caption{Bias-Variance trade-off for \oracle-\sndr}
        \vspace{0.1cm}
        \includegraphics[width=0.495\linewidth,trim={0cm 3.4cm 0 0},clip]{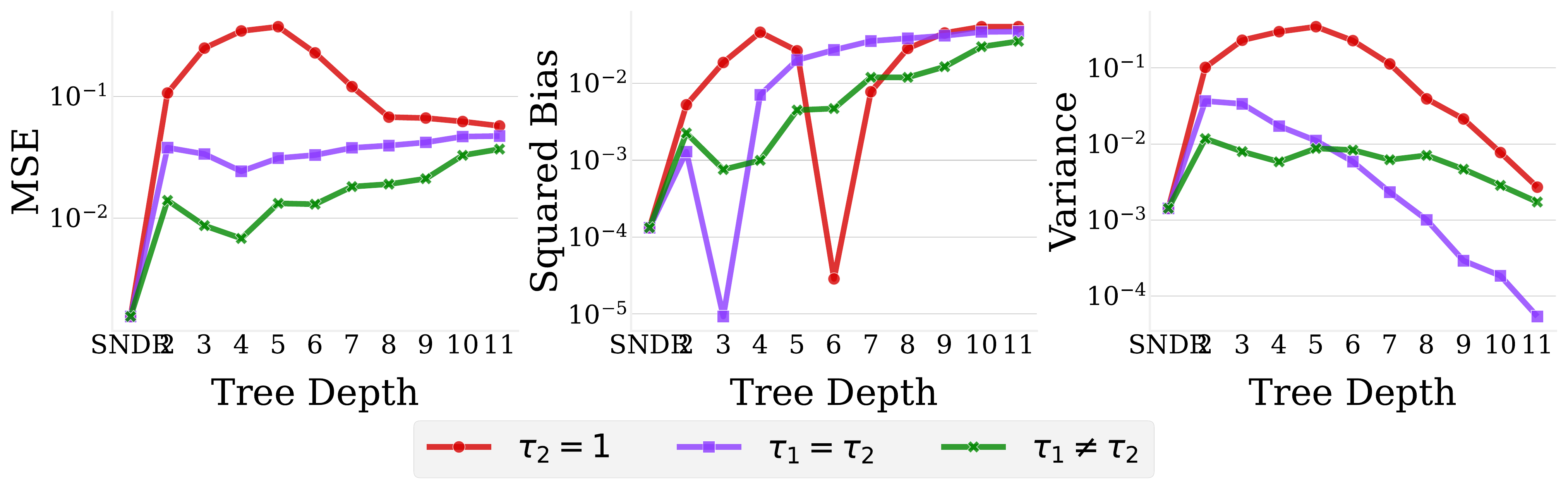} \hfill %
        \includegraphics[width=0.495\linewidth,trim={0cm 3.4cm 0 0},clip]{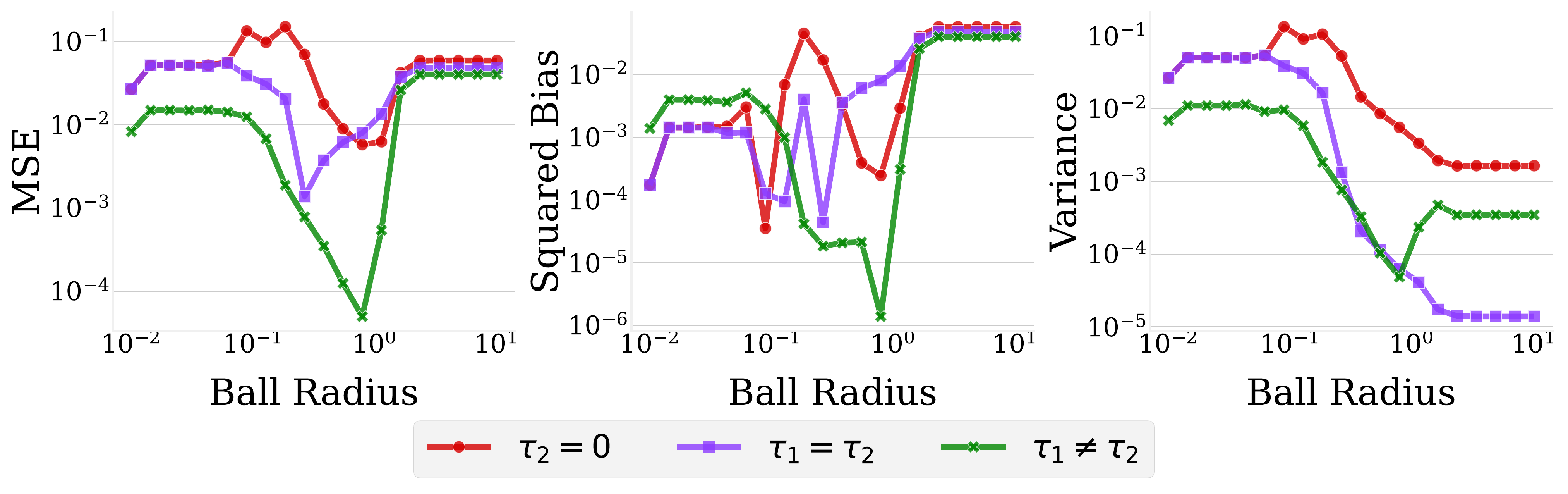}
        \includegraphics[width=0.495\linewidth,trim={0cm 3.4cm 0 0},clip]{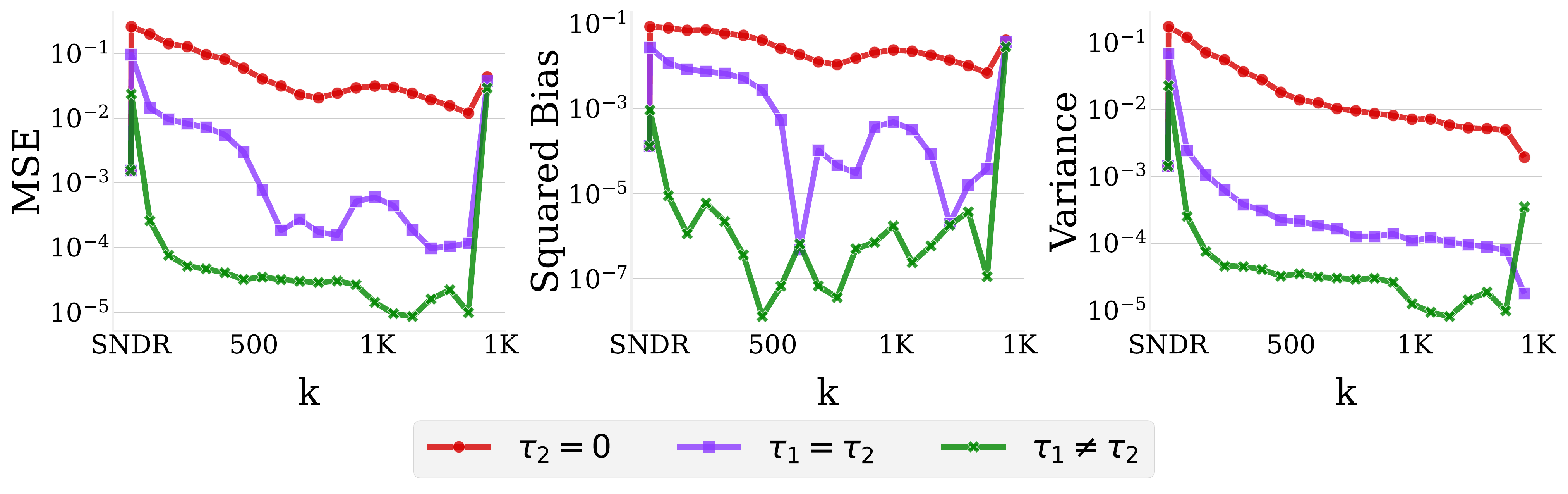} \hfill %
        \includegraphics[width=0.495\linewidth,trim={0cm 3.4cm 0 0},clip]{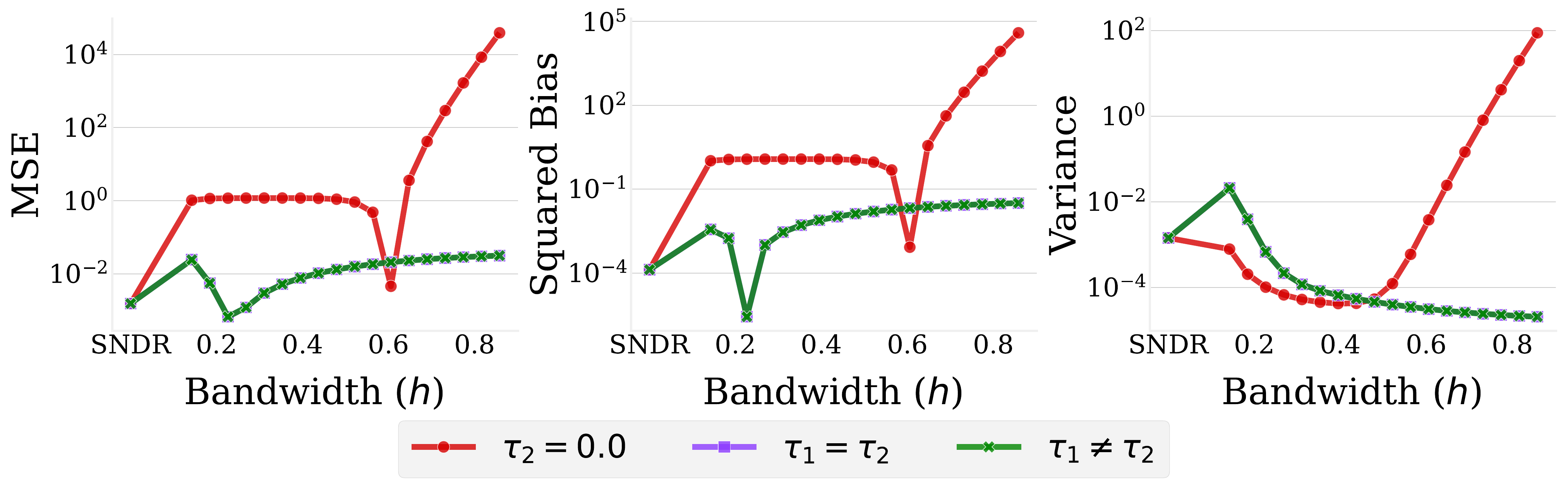}
        \includegraphics[width=\linewidth,trim={0 0 0 18cm},clip]{figures/bias_variance/synthetic/2k_actions_beta_3_eps_0.05/IPS/kNN_num.pdf}
    \end{subfigure}
    \vspace{-10pt}
    \caption{Visualizing the bias-variance trade-off for the \oracle estimator for different backbones \& pooling strategies, while estimating $V(\pi_{\mathsf{good}})$ with varying amount of pooling ($\log$-$\log$ scale) on the movielens dataset, using $\mu_{\mathsf{good}}$ for logging. Note that the respective na\"ive backbone estimators are the left-most point in each plot, \ie, when there's \textbf{no} pooling.}
    \label{fig:v_pooling_ml_pi_good_mu_good}
\end{minipage}
\end{figure*}

\clearpage

\begin{figure*}
\begin{minipage}[c][\textheight][c]{\textwidth}
    \centering
    \begin{subfigure}{\textwidth}
        \caption{Optimal amount of pooling with varying number of actions, \ie, $|\mathcal{A}|$ ($\log$ scale)}
        \includegraphics[width=\linewidth,trim={0 5.2cm 0 0},clip]{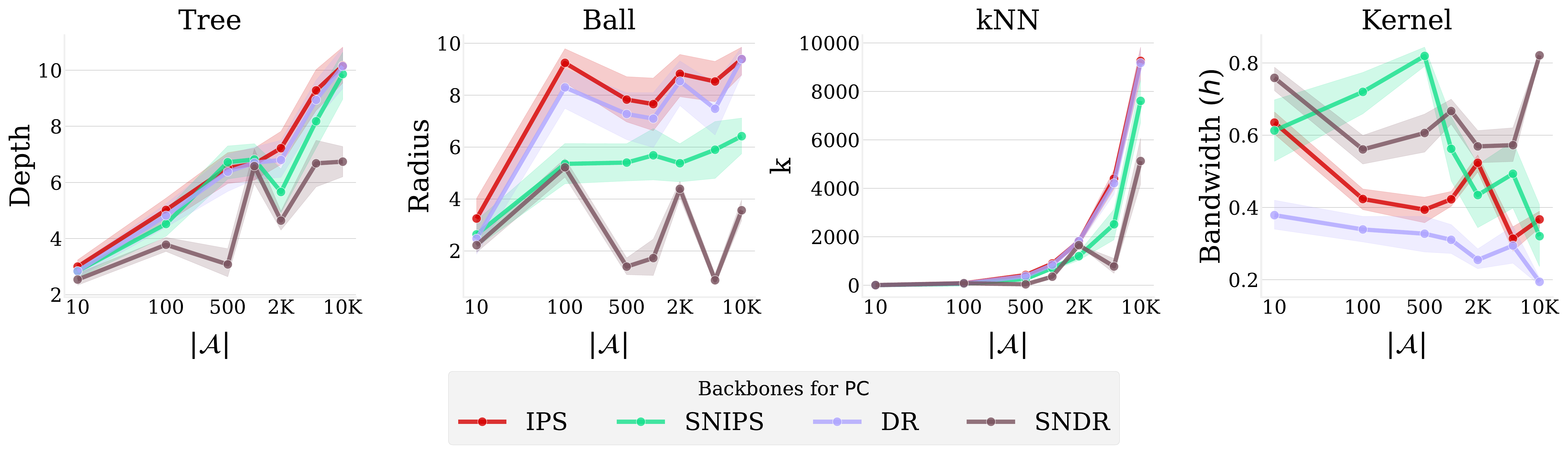}
    \end{subfigure} 
    \begin{subfigure}{\textwidth}
        \caption{Optimal amount of pooling with varying amount of bandit feedback, \ie, $|\mathcal{D}|$ ($\log$ scale)}
        \includegraphics[width=\linewidth,trim={0 0 0 0},clip]{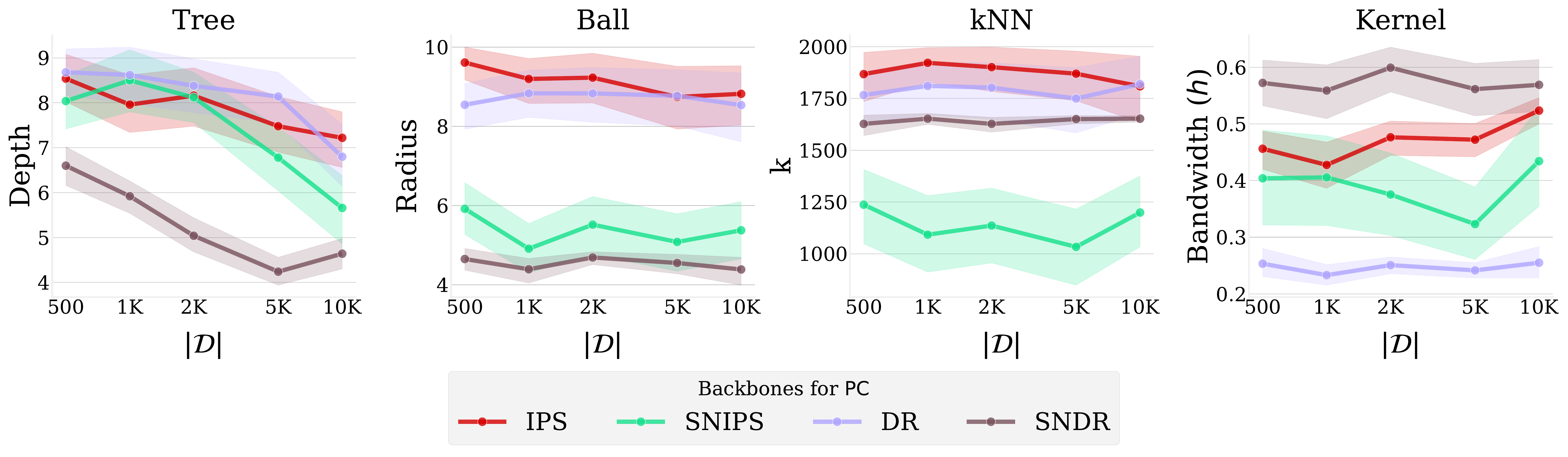}
    \end{subfigure}
    \caption{Change in the optimal amount pooling for \estimator while estimating $V(\pi_{\mathsf{good}})$ for the synthetic dataset, using data logged by $\mu_{\mathsf{bad}}$.}
    \label{fig:synthetic_pi_good_mu_bad}
\end{minipage}
\end{figure*}

\begin{figure*}
\begin{minipage}[c][\textheight][c]{\textwidth}
    \centering
    \begin{subfigure}{\textwidth}
        \caption{Optimal amount of pooling with varying number of actions, \ie, $|\mathcal{A}|$ ($\log$ scale)}
        \includegraphics[width=\linewidth,trim={0 5.2cm 0 0},clip]{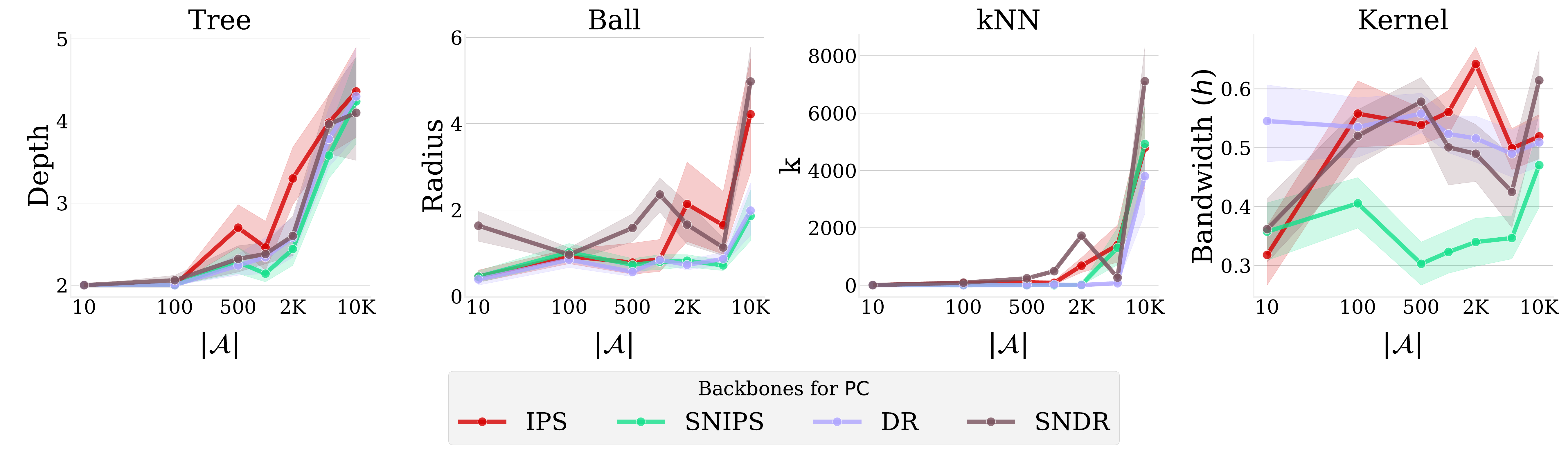}
    \end{subfigure} 
    \begin{subfigure}{\textwidth}
        \caption{Optimal amount of pooling with varying amount of bandit feedback, \ie, $|\mathcal{D}|$ ($\log$ scale)}
        \includegraphics[width=\linewidth,trim={0 0 0 0},clip]{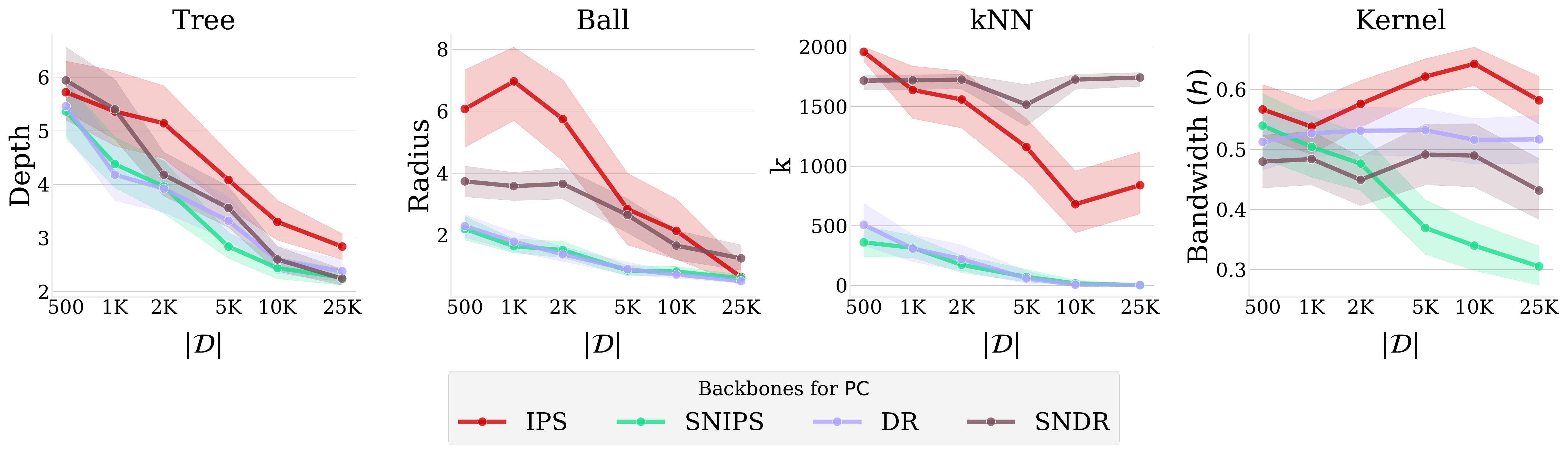}
    \end{subfigure}
    \caption{Change in the optimal amount pooling for \estimator while estimating $V(\pi_{\mathsf{good}})$ for the synthetic dataset, using data logged by $\mu_{\mathsf{uniform}}$.}
    \label{fig:synthetic_pi_good_mu_unif}
\end{minipage}
\end{figure*}

\begin{figure*}
\begin{minipage}[c][\textheight][c]{\textwidth}
    \centering
    \begin{subfigure}{\textwidth}
        \caption{Optimal amount of pooling with varying number of actions, \ie, $|\mathcal{A}|$ ($\log$ scale)}
        \includegraphics[width=\linewidth,trim={0 5.2cm 0 0},clip]{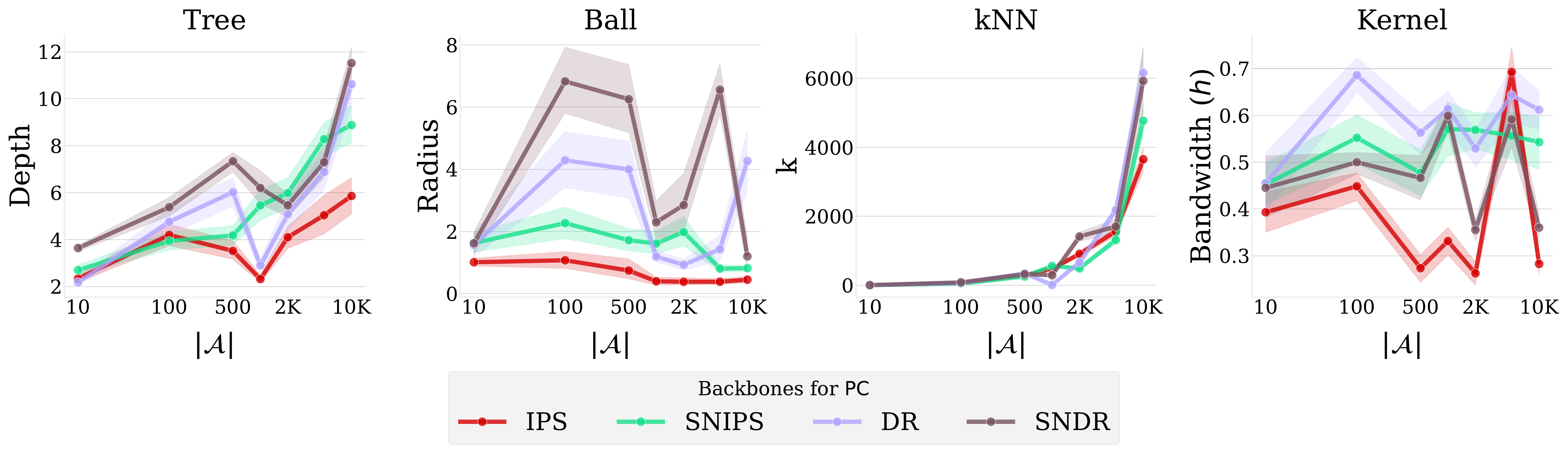}
    \end{subfigure} 
    \begin{subfigure}{\textwidth}
        \caption{Optimal amount of pooling with varying amount of bandit feedback, \ie, $|\mathcal{D}|$ ($\log$ scale)}
        \includegraphics[width=\linewidth,trim={0 0 0 0},clip]{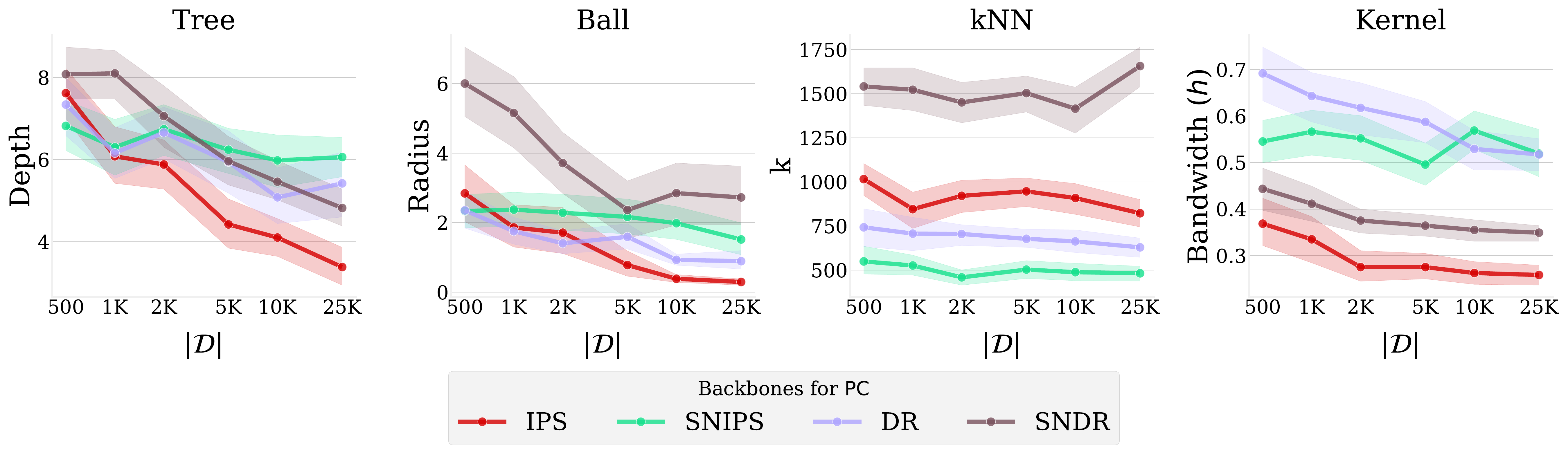}
    \end{subfigure}
    \caption{Change in the optimal amount pooling for \estimator while estimating $V(\pi_{\mathsf{good}})$ for the synthetic dataset, using data logged by $\mu_{\mathsf{good}}$.}
    \label{fig:synthetic_pi_good_mu_good}
\end{minipage}
\end{figure*}

\end{document}